\documentclass[twoside,11pt]{article}
\usepackage[preprint,abbrvbib]{jmlr2e}
\usepackage[T1]{fontenc}
\usepackage{amsmath,amssymb,booktabs,longtable,graphicx,array,microtype,xurl,placeins}
\usepackage{newtxtext,newtxmath}
\makeatletter
\g@addto@macro\normalsize{\abovedisplayskip=4pt plus 2pt minus 1pt
  \belowdisplayskip=4pt plus 2pt minus 1pt
  \abovedisplayshortskip=0pt plus 2pt
  \belowdisplayshortskip=4pt plus 2pt minus 1pt}
\long\def\@makecaption#1#2{\vskip 10pt
  {\footnotesize\settowidth{\@tempdima}{#1: }\noindent\hangindent\@tempdima\hangafter1 #1: #2\par}}
\makeatother
\renewenvironment{proof}{\par\noindent{\bfseries Proof\ }}
  {\hfill\BlackBox\par\addvspace{2mm}}
\hypersetup{hidelinks,pdftitle={Continuous Noisy-Loop Models},pdfauthor={An T. Le},pdfsubject={Statistical learning and tractable probabilistic inference}}
\ShortHeadings{Continuous Noisy-Loop Models}{Le}
\firstpageno{1}
\newcommand{\E}{\mathbb{E}}
\newcommand{\R}{\mathbb{R}}
\newcommand{\KL}{\operatorname{KL}}
\newcommand{\TV}{\operatorname{TV}}
\newcommand{\Aff}{\operatorname{Aff}}
\newcommand{\Var}{\operatorname{Var}}
\newcommand{\diag}{\operatorname{diag}}
\newcommand{\ramp}{\operatorname{ramp}}
\newcommand{\clip}{\operatorname{clip}}
\newcommand{\artanh}{\operatorname{artanh}}
\newcommand{\girth}{\operatorname{girth}}
\newcommand{\sign}{\operatorname{sign}}

\newcommand{\SubmissionValue}[1]{\csname Submission#1\endcsname}\expandafter\def\csname Submissiont0-moment-kl\endcsname{0.017}\expandafter\def\csname Submissiont0-moment-n\endcsname{10}\expandafter\def\csname Submissiont0-likelihood-kl\endcsname{0.010}\expandafter\def\csname Submissiont0-likelihood-n\endcsname{10}\expandafter\def\csname Submissiont0-em-kl\endcsname{0.003}\expandafter\def\csname Submissiont0-em-n\endcsname{10}\expandafter\def\csname Submissiont0-gpu-kl\endcsname{1.946}\expandafter\def\csname Submissiont0-gpu-n\endcsname{10}\expandafter\def\csname Submissiont0-gmm-kl\endcsname{1.913}\expandafter\def\csname Submissiont0-gmm-n\endcsname{10}\expandafter\def\csname Submissiont0-mpc-kl\endcsname{1.984}\expandafter\def\csname Submissiont0-mpc-n\endcsname{10}\expandafter\def\csname Submissiont0-osos-kl\endcsname{2.806}\expandafter\def\csname Submissiont0-osos-n\endcsname{10}\expandafter\def\csname Submissiont0-independence-kl\endcsname{1.971}\expandafter\def\csname Submissiont0-independence-n\endcsname{10}\expandafter\def\csname Submissiont0-shrinkage-kl\endcsname{0.017}\expandafter\def\csname Submissiont0-shrinkage-n\endcsname{10}\expandafter\def\csname Submissiont0-corner-kl\endcsname{0.016}\expandafter\def\csname Submissiont0-corner-n\endcsname{10}\expandafter\def\csname Submissiont1-moment-kl\endcsname{0.039}\expandafter\def\csname Submissiont1-moment-n\endcsname{10}\expandafter\def\csname Submissiont1-likelihood-kl\endcsname{0.031}\expandafter\def\csname Submissiont1-likelihood-n\endcsname{10}\expandafter\def\csname Submissiont1-em-kl\endcsname{0.022}\expandafter\def\csname Submissiont1-em-n\endcsname{10}\expandafter\def\csname Submissiont1-gpu-kl\endcsname{1.639}\expandafter\def\csname Submissiont1-gpu-n\endcsname{10}\expandafter\def\csname Submissiont1-gmm-kl\endcsname{1.736}\expandafter\def\csname Submissiont1-gmm-n\endcsname{10}\expandafter\def\csname Submissiont1-mpc-kl\endcsname{1.797}\expandafter\def\csname Submissiont1-mpc-n\endcsname{10}\expandafter\def\csname Submissiont1-osos-kl\endcsname{2.615}\expandafter\def\csname Submissiont1-osos-n\endcsname{10}\expandafter\def\csname Submissiont1-independence-kl\endcsname{1.785}\expandafter\def\csname Submissiont1-independence-n\endcsname{10}\expandafter\def\csname Submissiont1-shrinkage-kl\endcsname{0.039}\expandafter\def\csname Submissiont1-shrinkage-n\endcsname{10}\expandafter\def\csname Submissiont1-corner-kl\endcsname{0.039}\expandafter\def\csname Submissiont1-corner-n\endcsname{10}\expandafter\def\csname Submissiont2-moment-kl\endcsname{3.324}\expandafter\def\csname Submissiont2-moment-n\endcsname{10}\expandafter\def\csname Submissiont2-likelihood-kl\endcsname{3.323}\expandafter\def\csname Submissiont2-likelihood-n\endcsname{10}\expandafter\def\csname Submissiont2-em-kl\endcsname{3.324}\expandafter\def\csname Submissiont2-em-n\endcsname{10}\expandafter\def\csname Submissiont2-gpu-kl\endcsname{0.033}\expandafter\def\csname Submissiont2-gpu-n\endcsname{10}\expandafter\def\csname Submissiont2-gmm-kl\endcsname{0.004}\expandafter\def\csname Submissiont2-gmm-n\endcsname{10}\expandafter\def\csname Submissiont2-mpc-kl\endcsname{0.097}\expandafter\def\csname Submissiont2-mpc-n\endcsname{10}\expandafter\def\csname Submissiont2-osos-kl\endcsname{2.242}\expandafter\def\csname Submissiont2-osos-n\endcsname{10}\expandafter\def\csname Submissiont2-independence-kl\endcsname{3.324}\expandafter\def\csname Submissiont2-independence-n\endcsname{10}\expandafter\def\csname Submissiont2-shrinkage-kl\endcsname{3.324}\expandafter\def\csname Submissiont2-shrinkage-n\endcsname{10}\expandafter\def\csname Submissiont2-corner-kl\endcsname{3.324}\expandafter\def\csname Submissiont2-corner-n\endcsname{10}\expandafter\def\csname Submissiont3-moment-kl\endcsname{0.116}\expandafter\def\csname Submissiont3-moment-n\endcsname{10}\expandafter\def\csname Submissiont3-likelihood-kl\endcsname{0.073}\expandafter\def\csname Submissiont3-likelihood-n\endcsname{10}\expandafter\def\csname Submissiont3-em-kl\endcsname{0.014}\expandafter\def\csname Submissiont3-em-n\endcsname{10}\expandafter\def\csname Submissiont3-gpu-kl\endcsname{11.862}\expandafter\def\csname Submissiont3-gpu-n\endcsname{10}\expandafter\def\csname Submissiont3-gmm-kl\endcsname{11.918}\expandafter\def\csname Submissiont3-gmm-n\endcsname{10}\expandafter\def\csname Submissiont3-mpc-kl\endcsname{11.881}\expandafter\def\csname Submissiont3-mpc-n\endcsname{10}\expandafter\def\csname Submissiont3-osos-kl\endcsname{13.786}\expandafter\def\csname Submissiont3-osos-n\endcsname{10}\expandafter\def\csname Submissiont3-independence-kl\endcsname{11.820}\expandafter\def\csname Submissiont3-independence-n\endcsname{10}\expandafter\def\csname Submissiont3-shrinkage-kl\endcsname{0.116}\expandafter\def\csname Submissiont3-shrinkage-n\endcsname{10}\expandafter\def\csname Submissiont3-corner-kl\endcsname{0.115}\expandafter\def\csname Submissiont3-corner-n\endcsname{10}\expandafter\def\csname Submissiont4-moment-kl\endcsname{0.254}\expandafter\def\csname Submissiont4-moment-n\endcsname{10}\expandafter\def\csname Submissiont4-likelihood-kl\endcsname{0.173}\expandafter\def\csname Submissiont4-likelihood-n\endcsname{10}\expandafter\def\csname Submissiont4-em-kl\endcsname{0.124}\expandafter\def\csname Submissiont4-em-n\endcsname{10}\expandafter\def\csname Submissiont4-gpu-kl\endcsname{10.728}\expandafter\def\csname Submissiont4-gpu-n\endcsname{10}\expandafter\def\csname Submissiont4-gmm-kl\endcsname{10.768}\expandafter\def\csname Submissiont4-gmm-n\endcsname{10}\expandafter\def\csname Submissiont4-mpc-kl\endcsname{10.731}\expandafter\def\csname Submissiont4-mpc-n\endcsname{10}\expandafter\def\csname Submissiont4-osos-kl\endcsname{12.606}\expandafter\def\csname Submissiont4-osos-n\endcsname{10}\expandafter\def\csname Submissiont4-independence-kl\endcsname{10.668}\expandafter\def\csname Submissiont4-independence-n\endcsname{10}\expandafter\def\csname Submissiont4-shrinkage-kl\endcsname{0.254}\expandafter\def\csname Submissiont4-shrinkage-n\endcsname{10}\expandafter\def\csname Submissiont4-corner-kl\endcsname{0.254}\expandafter\def\csname Submissiont4-corner-n\endcsname{10}\expandafter\def\csname Submissiont5-moment-kl\endcsname{15.964}\expandafter\def\csname Submissiont5-moment-n\endcsname{10}\expandafter\def\csname Submissiont5-likelihood-kl\endcsname{15.963}\expandafter\def\csname Submissiont5-likelihood-n\endcsname{10}\expandafter\def\csname Submissiont5-em-kl\endcsname{15.963}\expandafter\def\csname Submissiont5-em-n\endcsname{10}\expandafter\def\csname Submissiont5-gpu-kl\endcsname{0.168}\expandafter\def\csname Submissiont5-gpu-n\endcsname{10}\expandafter\def\csname Submissiont5-gmm-kl\endcsname{0.101}\expandafter\def\csname Submissiont5-gmm-n\endcsname{10}\expandafter\def\csname Submissiont5-mpc-kl\endcsname{0.493}\expandafter\def\csname Submissiont5-mpc-n\endcsname{10}\expandafter\def\csname Submissiont5-osos-kl\endcsname{10.267}\expandafter\def\csname Submissiont5-osos-n\endcsname{10}\expandafter\def\csname Submissiont5-independence-kl\endcsname{15.963}\expandafter\def\csname Submissiont5-independence-n\endcsname{10}\expandafter\def\csname Submissiont5-shrinkage-kl\endcsname{15.963}\expandafter\def\csname Submissiont5-shrinkage-n\endcsname{10}\expandafter\def\csname Submissiont5-corner-kl\endcsname{15.964}\expandafter\def\csname Submissiont5-corner-n\endcsname{10}\expandafter\def\csname Submissionlikelihood-loop-cycle-starts\endcsname{32}\expandafter\def\csname Submissionlikelihood-loop-starts\endcsname{40}\expandafter\def\csname Submissionem-loop-cycle-starts\endcsname{27}\expandafter\def\csname Submissionem-loop-starts\endcsname{40}\expandafter\def\csname SubmissionL4-gaussian-forest-floor\endcsname{1.1522}\expandafter\def\csname SubmissionL8-gaussian-forest-floor\endcsname{6.9044}\expandafter\def\csname Submissiont3-moment-random25-conditional-kl\endcsname{0.0852}\expandafter\def\csname Submissiont3-likelihood-random25-conditional-kl\endcsname{0.0534}\expandafter\def\csname Submissiont3-em-random25-conditional-kl\endcsname{0.0104}\expandafter\def\csname Submissionweak-ramp-e0.28-energy-bound\endcsname{0.00263}\expandafter\def\csname Submissionweak-ramp-e0.375-energy-bound\endcsname{0.02802}\expandafter\def\csname Submissionweak-ramp-e0.28-n32768-raw-kl\endcsname{0.0661}\expandafter\def\csname Submissionweak-ramp-e0.28-n32768-shrunk-kl\endcsname{0.00134}\expandafter\def\csname Submissionweak-ramp-e0.28-independence-kl\endcsname{0.00132}\expandafter\def\csname Submissionweak-ramp-e0.28-independence-se\endcsname{0.00005}\expandafter\def\csname Submissionweak-ramp-e0.5-n8192-raw-kl\endcsname{0.1785}\expandafter\def\csname Submissionweak-ramp-e0.5-n8192-shrunk-kl\endcsname{0.0915}\expandafter\def\csname Submissionweak-ramp-e0.5-n8192-paired-independence-kl\endcsname{0.1457}\expandafter\def\csname Submissionweak-ramp-e0.5-n8192-paired-shrink-delta\endcsname{-0.0542}\expandafter\def\csname Submissionweak-snr-threshold-n8192\endcsname{0.472}\expandafter\def\csname Submissionweak-ramp-e0.5-n32768-raw-kl\endcsname{0.0549}\expandafter\def\csname Submissionweak-ramp-e0.5-n32768-shrunk-kl\endcsname{0.0538}\expandafter\def\csname Submissionweak-ramp-e0.5-n32768-paired-independence-kl\endcsname{0.1457}\expandafter\def\csname Submissionweak-ramp-e0.5-n32768-paired-shrink-delta\endcsname{-0.0919}\expandafter\def\csname Submissionweak-snr-threshold-n32768\endcsname{0.420}\expandafter\def\csname Submissionweak-ramp-e0.5-independence-kl\endcsname{0.1453}\expandafter\def\csname Submissionweak-snr-e0.28\endcsname{0.0076}\expandafter\def\csname Submissionweak-snr-e0.5\endcsname{8}\expandafter\def\csname Submissionweak-independence-bound-ratio-min\endcsname{0.47}\expandafter\def\csname Submissionweak-independence-bound-ratio-max\endcsname{0.61}\expandafter\def\csname Submissionweak-e0.5-n8192-lambda\endcsname{0.75}\expandafter\def\csname Submissionweak-e0.83125-lambda\endcsname{1}\expandafter\def\csname Submissionweak-seeds\endcsname{5}\expandafter\def\csname Submissionprimary-planned\endcsname{42}\expandafter\def\csname Submissionprimary-min-pairs\endcsname{10}\expandafter\def\csname Submissionosos-large-loop-capped\endcsname{20}\expandafter\def\csname Submissionosos-large-loop-n\endcsname{20}\expandafter\def\csname Submissionem-large-loop-capped\endcsname{7}\expandafter\def\csname Submissionem-large-loop-n\endcsname{20}\expandafter\def\csname Submissionloop-corner-gain-max\endcsname{0.0008}\expandafter\def\csname Submissionloop-moment-fit-median-min\endcsname{0.36}\expandafter\def\csname Submissionloop-moment-fit-median-max\endcsname{0.53}\expandafter\def\csname Submissionloop-likelihood-fit-median-min\endcsname{5.8}\expandafter\def\csname Submissionloop-likelihood-fit-median-max\endcsname{70.9}\expandafter\def\csname Submissionloop-em-fit-median-min\endcsname{322}\expandafter\def\csname Submissionloop-em-fit-median-max\endcsname{1127}\expandafter\def\csname Submissionconfirmation-trajectories\endcsname{540}\expandafter\def\csname Submissionconfirmation-capped\endcsname{57}\expandafter\def\csname Submissiondevelopment-trajectories\endcsname{144}\expandafter\def\csname Submissiondevelopment-capped\endcsname{44}\expandafter\def\csname Submissionloop-gpu-width\endcsname{32}\expandafter\def\csname Submissionloop-mpc-width\endcsname{8}\expandafter\def\csname Submissionloop-osos-width\endcsname{32}\expandafter\def\csname Submissiont0-gmm-components\endcsname{16}\expandafter\def\csname Submissiont1-gmm-components\endcsname{8}\expandafter\def\csname Submissiont3-gmm-components\endcsname{1}\expandafter\def\csname Submissiont4-gmm-components\endcsname{1}\expandafter\def\csname SubmissionL4-gaussian-forest-size\endcsname{12}\expandafter\def\csname SubmissionL8-gaussian-forest-size\endcsname{56}\expandafter\def\csname Submissionmoment-query-fits\endcsname{460}\expandafter\def\csname Submissionmoment-query-checks\endcsname{5{,}520}\expandafter\def\csname Submissionmoment-query-max-error\endcsname{5.7\times10^{-14}}\expandafter\def\csname Submissionquadrature-fits\endcsname{440}\expandafter\def\csname Submissionquadrature-max-error\endcsname{1.9\times10^{-15}}\expandafter\def\csname Submissionrate-t0-n524288-k8-kl\endcsname{0.0017}\expandafter\def\csname Submissionrate-t0-n524288-kgrow-kl\endcsname{0.0143}\expandafter\def\csname Submissionrate-t1-n524288-k8-kl\endcsname{0.0240}\expandafter\def\csname Submissionrate-t1-n524288-kgrow-kl\endcsname{0.0123}\expandafter\def\csname Submissionrate-t3-n524288-k8-kl\endcsname{0.0118}\expandafter\def\csname Submissionrate-t3-n524288-kgrow-kl\endcsname{0.0992}\expandafter\def\csname Submissionrate-t4-n524288-k8-kl\endcsname{0.1548}\expandafter\def\csname Submissionrate-t4-n524288-kgrow-kl\endcsname{0.0866}\expandafter\def\csname Submissionrate-fits\endcsname{320}\expandafter\def\csname Submissionrate-kl-ise-min-24\endcsname{0.60}\expandafter\def\csname Submissionrate-kl-ise-max-24\endcsname{2.03}\expandafter\def\csname Submissionrate-kl-ise-min-112\endcsname{0.81}\expandafter\def\csname Submissionrate-kl-ise-max-112\endcsname{2.62}\expandafter\def\csname Submissionrate-ramp-k8-kl-decay-min\endcsname{0.78}\expandafter\def\csname Submissionrate-ramp-k8-kl-decay-max\endcsname{0.81}\expandafter\def\csname Submissionrate-smooth-k8-kl-decay-min\endcsname{0.23}\expandafter\def\csname Submissionrate-smooth-k8-kl-decay-max\endcsname{0.25}\expandafter\def\csname Submissionrate-kgrow-kl-decay-min\endcsname{0.48}\expandafter\def\csname Submissionrate-kgrow-kl-decay-max\endcsname{0.50}\expandafter\def\csname Submissionrate-ramp-k8-ise-decay-min\endcsname{0.88}\expandafter\def\csname Submissionrate-ramp-k8-ise-decay-max\endcsname{0.90}\expandafter\def\csname Submissionrate-kgrow-ise-decay-min\endcsname{0.61}\expandafter\def\csname Submissionrate-kgrow-ise-decay-max\endcsname{0.62}\expandafter\def\csname Submissionrate-seeds\endcsname{10}\expandafter\def\csname Submissionrate-test-vectors\endcsname{65{,}536}\expandafter\def\csname Submissionnonramp-fits\endcsname{120}\expandafter\def\csname Submissionnonramp-fits-above-independence\endcsname{120}\expandafter\def\csname Submissionnonramp-L8-n32768-kl\endcsname{0.224}\expandafter\def\csname Submissionnonramp-L8-n32768-independence-kl\endcsname{0.044}\expandafter\def\csname Submissionnonramp-L8-n32768-interior-rise\endcsname{1.129}\expandafter\def\csname Submissionnonramp-L8-n32768-boundary-rise\endcsname{1.129}\expandafter\def\csname Submissionnonramp-L8-n32768-corner-rise\endcsname{3.255}\expandafter\def\csname Submissionnonramp-n32768-regions\endcsname{9}\expandafter\def\csname Submissionnonramp-n32768-regions-above-one\endcsname{9}\expandafter\def\csname Submissionnonramp-n32768-clip-min\endcsname{20.4}\expandafter\def\csname Submissionnonramp-n32768-clip-max\endcsname{27.5}\expandafter\def\csname Submissionnonramp-moment-min\endcsname{3.38\times10^{-6}}\expandafter\def\csname Submissionnonramp-moment-max\endcsname{1.68\times10^{-5}}\providecommand{\SubmissionSupport}[1]{\csname SubmissionSupport#1\endcsname}\expandafter\def\csname SubmissionSupportnumerical-verified\endcsname{72}\expandafter\def\csname SubmissionSupportnumerical-scheduled\endcsname{72}\expandafter\def\csname SubmissionSupportnumerical-queries\endcsname{144}\expandafter\def\csname SubmissionSupportnumerical-rejected\endcsname{2}\expandafter\def\csname SubmissionSupportnumerical-accepted\endcsname{142}\expandafter\def\csname SubmissionSupportnumerical-rejected-dim-1\endcsname{1{,}912}\expandafter\def\csname SubmissionSupportnumerical-rejected-bond-error-1\endcsname{1.9\times10^{-3}}\expandafter\def\csname SubmissionSupportnumerical-rejected-dim-2\endcsname{14{,}332}\expandafter\def\csname SubmissionSupportnumerical-rejected-bond-error-2\endcsname{3.1\times10^{-7}}\expandafter\def\csname SubmissionSupportnumerical-accepted-log-error\endcsname{2.7\times10^{-8}}\expandafter\def\csname SubmissionSupportnumerical-accepted-bond-error\endcsname{5.5\times10^{-9}}

\newcommand{\TilingSixTV}{0.7303}

\newcommand{\TilingDraws}{10{,}000}
\newcommand{\TilingKLMin}{0.0213}
\newcommand{\TilingKLMax}{0.0363}
\newcommand{\TilingTVBoundMin}{0.330}
\newcommand{\TilingTVBoundMax}{0.915}
\newcommand{\TilingCheckDraws}{8}
\newcommand{\TilingTransferMaxSide}{24}
\newcommand{\TilingTransferMaxErr}{1.2\times10^{-13}}
\newcommand{\TilingDenseMaxErr}{1.4\times10^{-12}}
\newcommand{\TilingCondMax}{4.3\times10^{6}}
\providecommand{\DiagnosticValue}[1]{\csname Diagnostic#1\endcsname}\expandafter\def\csname Diagnosticnonramp-current-d24-n512-legacy-kl\endcsname{0.044}\expandafter\def\csname Diagnosticnonramp-current-d24-n512-short-kl\endcsname{0.037}\expandafter\def\csname Diagnosticnonramp-current-d24-n512-oracle-kl\endcsname{0.093}\expandafter\def\csname Diagnosticnonramp-current-d24-n2048-legacy-kl\endcsname{0.061}\expandafter\def\csname Diagnosticnonramp-current-d24-n2048-short-kl\endcsname{0.040}\expandafter\def\csname Diagnosticnonramp-current-d24-n2048-oracle-kl\endcsname{0.119}\expandafter\def\csname Diagnosticnonramp-current-d24-n8192-legacy-kl\endcsname{0.056}\expandafter\def\csname Diagnosticnonramp-current-d24-n8192-short-kl\endcsname{0.034}\expandafter\def\csname Diagnosticnonramp-current-d24-n8192-oracle-kl\endcsname{0.063}\expandafter\def\csname Diagnosticnonramp-current-d24-n32768-legacy-kl\endcsname{0.042}\expandafter\def\csname Diagnosticnonramp-current-d24-n32768-short-kl\endcsname{0.016}\expandafter\def\csname Diagnosticnonramp-current-d24-n32768-oracle-kl\endcsname{0.025}\expandafter\def\csname Diagnosticnonramp-current-d112-n512-legacy-kl\endcsname{0.236}\expandafter\def\csname Diagnosticnonramp-current-d112-n512-short-kl\endcsname{0.154}\expandafter\def\csname Diagnosticnonramp-current-d112-n512-oracle-kl\endcsname{0.446}\expandafter\def\csname Diagnosticnonramp-current-d112-n2048-legacy-kl\endcsname{0.312}\expandafter\def\csname Diagnosticnonramp-current-d112-n2048-short-kl\endcsname{0.206}\expandafter\def\csname Diagnosticnonramp-current-d112-n2048-oracle-kl\endcsname{0.511}\expandafter\def\csname Diagnosticnonramp-current-d112-n8192-legacy-kl\endcsname{0.281}\expandafter\def\csname Diagnosticnonramp-current-d112-n8192-short-kl\endcsname{0.176}\expandafter\def\csname Diagnosticnonramp-current-d112-n8192-oracle-kl\endcsname{0.338}\expandafter\def\csname Diagnosticnonramp-current-d112-n32768-legacy-kl\endcsname{0.224}\expandafter\def\csname Diagnosticnonramp-current-d112-n32768-short-kl\endcsname{0.125}\expandafter\def\csname Diagnosticnonramp-current-d112-n32768-oracle-kl\endcsname{0.153}\expandafter\def\csname Diagnosticnonramp-current-d480-n512-legacy-kl\endcsname{1.165}\expandafter\def\csname Diagnosticnonramp-current-d480-n512-short-kl\endcsname{0.899}\expandafter\def\csname Diagnosticnonramp-current-d480-n512-oracle-kl\endcsname{2.206}\expandafter\def\csname Diagnosticnonramp-current-d480-n2048-legacy-kl\endcsname{1.517}\expandafter\def\csname Diagnosticnonramp-current-d480-n2048-short-kl\endcsname{1.041}\expandafter\def\csname Diagnosticnonramp-current-d480-n2048-oracle-kl\endcsname{2.236}\expandafter\def\csname Diagnosticnonramp-current-d480-n8192-legacy-kl\endcsname{1.382}\expandafter\def\csname Diagnosticnonramp-current-d480-n8192-short-kl\endcsname{0.876}\expandafter\def\csname Diagnosticnonramp-current-d480-n8192-oracle-kl\endcsname{1.465}\expandafter\def\csname Diagnosticnonramp-current-d480-n32768-legacy-kl\endcsname{1.056}\expandafter\def\csname Diagnosticnonramp-current-d480-n32768-short-kl\endcsname{0.685}\expandafter\def\csname Diagnosticnonramp-current-d480-n32768-oracle-kl\endcsname{0.667}\expandafter\def\csname Diagnosticnonramp-current-d24-independence-kl\endcsname{0.006}\expandafter\def\csname Diagnosticnonramp-current-d112-independence-kl\endcsname{0.044}\expandafter\def\csname Diagnosticnonramp-current-d480-independence-kl\endcsname{0.186}\expandafter\def\csname Diagnosticnonramp-current-n32768-scalar-ise-min\endcsname{0.66}\expandafter\def\csname Diagnosticnonramp-current-n32768-scalar-ise-max\endcsname{0.79}\expandafter\def\csname Diagnosticnonramp-current-n32768-corner-ise-min\endcsname{1.45}\expandafter\def\csname Diagnosticnonramp-current-n32768-corner-ise-max\endcsname{1.74}\expandafter\def\csname Diagnosticnonramp-current-n32768-clip-min\endcsname{13.9}\expandafter\def\csname Diagnosticnonramp-current-n32768-clip-max\endcsname{14.6}\expandafter\def\csname Diagnosticnonramp-current-fits\endcsname{120}\expandafter\def\csname Diagnosticnonramp-current-short-above-independence\endcsname{120}\expandafter\def\csname Diagnosticnonramp-current-oracle-above-independence\endcsname{120}\expandafter\def\csname Diagnosticgeometry-kl-min\endcsname{0.0132}\expandafter\def\csname Diagnosticgeometry-kl-max\endcsname{0.0174}\expandafter\def\csname Diagnosticgeometry-aff-min\endcsname{0.00336}\expandafter\def\csname Diagnosticgeometry-aff-max\endcsname{0.00424}\expandafter\def\csname Diagnosticgeometry-max-side\endcsname{32}\expandafter\def\csname Diagnosticgeometry-max-d\endcsname{1{,}984}\expandafter\def\csname Diagnosticgeometry-samples\endcsname{4{,}000}\expandafter\def\csname Diagnosticgeometry-kappa\endcsname{0.45}\expandafter\def\csname Diagnosticgeometry-amplitude\endcsname{0.6}\expandafter\def\csname Diagnosticgeometry-c\endcsname{2.0\times10^{-7}}\expandafter\def\csname Diagnosticgeometry-upper\endcsname{0.390625}\expandafter\def\csname Diagnosticgeometry-corner-err\endcsname{8.9\times10^{-16}}\expandafter\def\csname Diagnosticgeometry-corner-naive\endcsname{0.0192}\expandafter\def\csname Diagnosticsharp-amplitude\endcsname{0.5}\expandafter\def\csname Diagnosticsharp-kappa-min\endcsname{0.02}\expandafter\def\csname Diagnosticsharp-kappa-max\endcsname{0.12}\expandafter\def\csname Diagnosticsharp-slope\endcsname{11.97}\expandafter\def\csname Diagnosticsharp-ratio-min\endcsname{0.47}\expandafter\def\csname Diagnosticsharp-ratio-max\endcsname{0.50}\expandafter\def\csname Diagnosticsharp-gap\endcsname{220}\expandafter\def\csname Diagnosticsharp-single-slope\endcsname{6.00}\expandafter\def\csname Diagnosticsharp-single-ratio\endcsname{0.25}\expandafter\def\csname Diagnosticquery-nd-mult-slope\endcsname{1.56}\expandafter\def\csname Diagnosticquery-nd-fill-slope\endcsname{1.17}\expandafter\def\csname Diagnosticquery-colamd-mult-slope\endcsname{1.73}\expandafter\def\csname Diagnosticquery-colamd-fill-slope\endcsname{1.25}\expandafter\def\csname Diagnosticquery-max-side\endcsname{256}\expandafter\def\csname Diagnosticquery-max-d\endcsname{130{,}560}\expandafter\def\csname Diagnosticquery-logdet-err\endcsname{7.3\times10^{-14}}\expandafter\def\csname Diagnosticrate-coupling-ratio-min\endcsname{0.45}\expandafter\def\csname Diagnosticrate-coupling-ratio-max\endcsname{0.66}\expandafter\def\csname Diagnosticrate-coupling-slope-min\endcsname{0.45}\expandafter\def\csname Diagnosticrate-coupling-slope-max\endcsname{0.46}\expandafter\def\csname Diagnosticrate-raw-ise-slope-min\endcsname{0.67}\expandafter\def\csname Diagnosticrate-raw-ise-slope-max\endcsname{0.67}\expandafter\def\csname Diagnosticrate-default-ise-slope-min\endcsname{0.62}\expandafter\def\csname Diagnosticrate-default-ise-slope-max\endcsname{0.62}\expandafter\def\csname Diagnosticrate-bound-slope\endcsname{0.69}\expandafter\def\csname Diagnosticrate-coupling-fisher-first-24\endcsname{0.51}\expandafter\def\csname Diagnosticrate-coupling-fisher-last-24\endcsname{0.97}\expandafter\def\csname Diagnosticrate-coupling-fisher-first-112\endcsname{0.49}\expandafter\def\csname Diagnosticrate-coupling-fisher-last-112\endcsname{0.96}\expandafter\def\csname Diagnosticrate-oracle-slope-change\endcsname{0.009}\expandafter\def\csname Diagnosticrate-diagnostic-seeds\endcsname{10}
\begin{document}
\title{Continuous Noisy-Loop Models}
\author{\name An T. Le \email an@robot-learning.de\\
\addr Center for AI Research, VinUniversity\\
Vinhomes Ocean Park, Gia Lam, Hanoi, Vietnam\\
Intelligent Autonomous Systems, TU Darmstadt\\
Hochschulstrasse 10, 64289 Darmstadt, Germany}
\maketitle
\begin{abstract}Efficient learning and inference need not imply a compact probabilistic
circuit. We quantify this distinction for continuous noisy readings of
agreement between adjacent latent spins on a known square grid with $d$ edges.
With known margins and bounded Lipschitz channels, minimax squared
Hellinger risk is $\Theta(\min\{1,dN^{-2/3}\})$ from $N$ independent
vectors, without a reliability lower bound. A spline estimator reaches this rate
in polynomial arithmetic time given a fixed positive reliability bound;
an exponential-size mixture removes that requirement.
At fixed positive reliability, nonnegative decomposable circuits need
$\exp(\Theta(\sqrt d))$ nodes for fixed total-variation error below one,
even when each leaf uses at most $\alpha d$ coordinates, $0<\alpha<1$.
Optimal Kullback--Leibler (KL) approximation error is
$\Theta(d/\max\{\log d,\sqrt\ell\})$ in either direction with at most
$d^r$ nodes and leaf scopes $\ell\leq\alpha d$, for fixed $r>1$ and $0<\alpha<2/3$.
Node counts exclude leaf computation. For three fixed masks, optimal joint
conditional KL has order $d/\log d$ when $\ell\leq(\log d)^2$.
Local cycle identities relate joint divergence
to identifiable channel products and separate weak-channel recovery
from density estimation. Factorized evidence preserves a planar
zero-field Ising model: given local integrals, exact queries cost
$O(d^{3/2})$ arithmetic operations and one root extraction.
\end{abstract}
\section{Introduction}
Four sensors around a square report noisy readings of whether adjacent
latent spins agree. Any three readings are independent, yet together
they predict the fourth. A model fitted to pairwise marginals sees
independence; a model that uses the cycle can learn their joint
law (Section~\ref{sec:loop-expansion}). We use this family to quantify
how learning and query costs differ from circuit approximation costs.

A \emph{noisy-loop law} places these sensors on a graph with $d$ edges.
With a known square grid, known margins, bounded Lipschitz channels
and a fixed positive reliability bound $\kappa$, learning and queries cost
polynomially many arithmetic operations. Yet decomposable nonnegative
circuits whose leaves each use at most $\alpha d$ coordinates, fixed $0<\alpha<1$, need
$\exp(\Theta(\sqrt d))$ nodes at fixed total-variation (TV) error below one
(Theorem~\ref{thm:separation}).

For polynomial-size circuits, we ask how much error remains if we allow
larger leaves or predict only missing coordinates. We give sharp KL
approximation bounds as leaf scope grows and sharp joint-prediction bounds
for three fixed masks (Theorems~\ref{thm:kl-size} and~\ref{thm:block-prediction}).
Corollary~\ref{cor:conditional-risk} compares these costs with prediction
from the learned spline density. Node counts exclude computation inside leaves.

Crossing cycles give the circuit bounds. Intersecting cycles identify
channel products whose errors control joint KL (Theorem~\ref{thm:witness-geometry}).
As reliability decreases, recovering these products becomes harder, with
sharp $\kappa^{12}$ dependence in the lower distance bound
(Proposition~\ref{prop:geometry-sharp}, Corollary~\ref{cor:recovery-lower});
density estimation has a uniform Hellinger rate (Theorem~\ref{thm:anchor-free}).

\noindent\emph{Related work.}
Our model uses the zero-field Ising expansion and describes Bayes-matched
disorder on the Nishimori line \citep{nishimori1981,iba1999} and spin
synchronization \citep{abbe2018grids}. Noise estimation uses ratios of
syndrome moments in \citet{wagner}, triplet moments in \citet{fu2020}, and
cycle consistency in \citet{lerman2022cemp}. We adapt cycle-moment ratios
to recover continuous channel products. The learning rate is $d$
times the one-dimensional Lipschitz rate \citep{tsybakov2009}; our query
algorithm applies planar Ising inference \citep{schraudolph2008,likhosherstov2019}.

Tractable marginals need not imply compact positive circuits:
\citet[Theorems~36--37]{martens} prove this for spanning-tree laws,
including limits of fixed-size circuits. \citet[Theorems~3--4]{dpp} use
the same law to separate determinantal point processes from nonnegative
decomposable circuits.
Approximate lower bounds at small TV tolerance hold for
deterministic decomposable circuits \citep[Theorem~4.1]{leland}.
\citet{zhao2026expressivity} analyze separation rank under a fixed vtree
and derive a separation when two functions need different vtrees for
compact representation. Our approximation bounds also apply without a
fixed vtree. Our circuit lower bounds adapt balanced decompositions for
nonnegative circuits \citep[Theorem~38]{martens} and Boolean circuits
\citep[Theorem~6]{bova2016}.
Within structured-decomposable circuits, monotone and squared
representations are incomparable \citep{wang2025squared}.
Tensor networks also have unbounded size gaps between nonnegative and
squared factorizations \citep{glasser2019}.
Mixtures of separable flows use correlated Gaussian bases \citep{yalavarthi2026},
and squared circuits allow negative or complex parameters
\citep{loconte2026orthogonal}. Our circuit bounds do not cover these classes.

\citet{kumar2026unlearnable} study local-query hardness and neural
learning failures for noisy code distributions. Our graph and spin law
fix the global constraints; we learn channels, not unknown logical sectors
(Section~\ref{sec:loop-expansion}).
\citet{cai2026masked} study identification from masked-block log loss on
finite product spaces. They show that positive probability of masking
all coordinates characterizes uniform control of joint KL by excess risk.
We instead bound circuit approximation error for three fixed masks in
our continuous family.
\section{Noisy-Loop Laws and Their Three Costs}\label{sec:model}
\noindent\emph{Notation.}
For densities relative to a common reference,
$\KL(p\Vert q)=\int p\log(p/q)$, $\TV(p,q)=\tfrac12\int|p-q|$,
$\Aff(p,q)=\int\sqrt{pq}$ and $H^2(p,q)=1-\Aff(p,q)$.
Forward KL measures excess expected log loss; TV measures the largest
error in an event probability. Affinity measures overlap between densities;
it equals one for identical laws, whereas $H^2$ then equals zero.
Logarithms are natural. We use $\mu$ for the product-uniform law on the
coordinates in question and identify laws with their densities.
Functional identities hold almost everywhere. For $F\subseteq E$,
$u_F=(u_e)_{e\in F}$ and $p_F$ is the marginal density on $F$.
Function norms $\|\cdot\|_2$ and $\|\cdot\|_\infty$ are taken in
$L^2(\mu)$ and $L^\infty(\mu)$; $\otimes$ denotes products on disjoint
coordinates. Constants may depend on fixed amplitude, anchor and smoothness parameters, but
not on $d,N$. We use $f\asymp g$ and $f=\Theta(g)$ for bounds
$cg\leq f\leq Cg$ with constants $c,C>0$ in the stated limit. Unless
stated otherwise, expectations in risk bounds are over the training sample.

\begin{figure}[htbp]
\centering
\includegraphics[width=\linewidth]{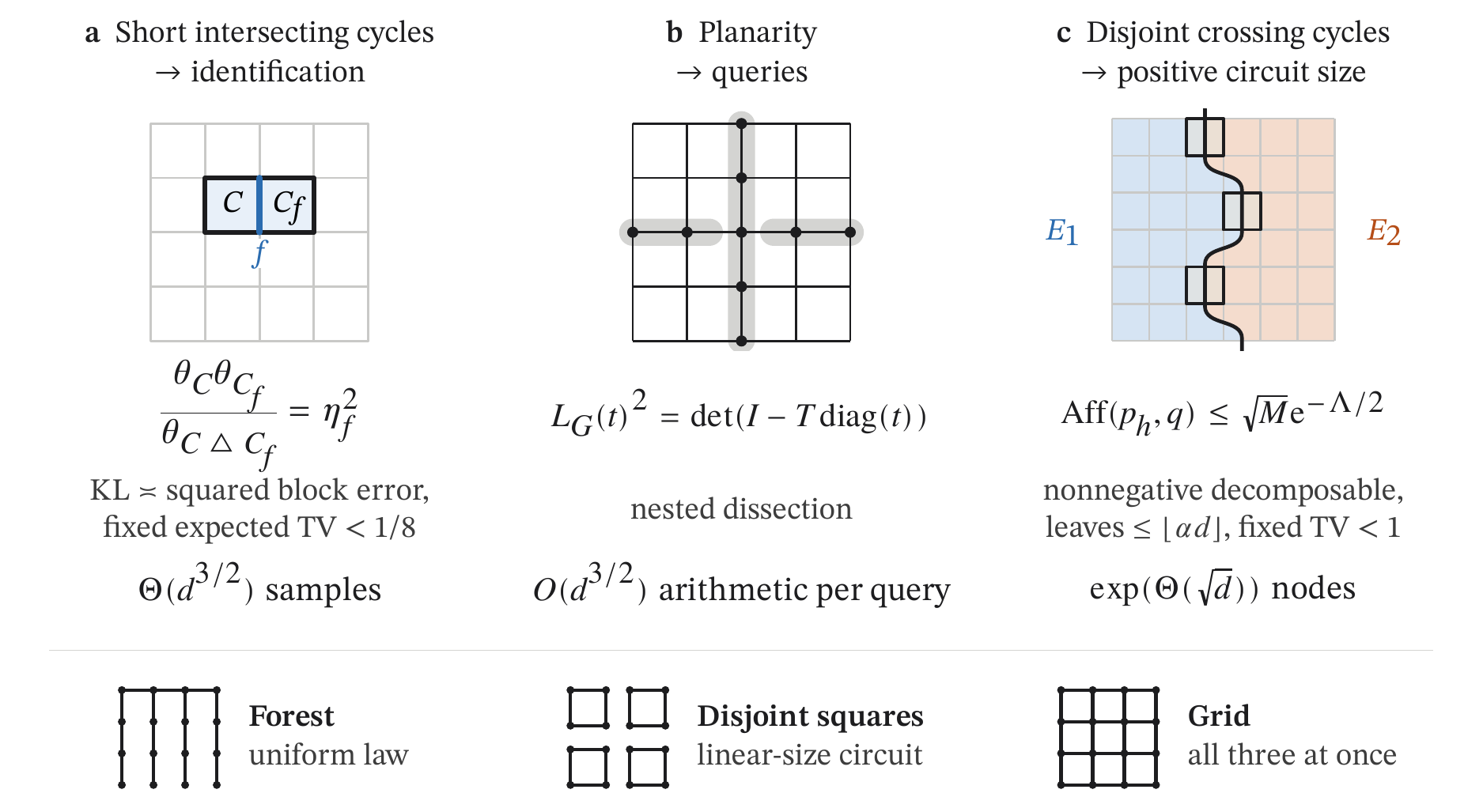}
\caption{Three costs on the anchored grid class (Theorem~\ref{thm:separation}).
(a) Adjacent squares and their six-cycle calibrate the shared-edge reliability.
KL is comparable to squared error in the identifiable blocks: scalar channels
and corner products (Corollary~\ref{cor:grid-geometry}).
(b) Given local evidence integrals, Kac--Ward evaluation uses nested dissection along $O(\sqrt d)$
separators (grey; Proposition~\ref{prop:queries}). (c) Every balanced
edge partition crosses $\Omega(\sqrt d)$ vertex-disjoint faces
(outlined), forcing large decomposable nonnegative circuits
(Corollary~\ref{cor:grid-circuit}). The displayed rates hold at fixed amplitude,
smoothness and reliability. Bottom: forests erase dependence,
disjoint squares have small positive circuits, and grids support all
three conclusions.}
\label{fig:costs}
\end{figure}

\noindent\emph{The model.}
Let $G=(V,E)$ be a finite simple graph with independent uniform spins
$S_v\in\{-1,1\}$. For $e=\{v,w\}$ the bond $B_e=S_vS_w$ records whether
the endpoints agree. Each edge carries a measurable channel
$h_e:[0,1]\to[-a_e,a_e]$ with $a_e<1$ and $\int_0^1h_e=0$. Given the
spins, the readings $U_e\in[0,1]$ are independent with densities
$1+B_eh_e(u_e)$, so the observed density is
\begin{equation}\label{eq:density}
 p_h(u)=\E\prod_{e\in E}\{1+B_eh_e(u_e)\}.
\end{equation}
Each factor is a positive probability density, so $p_h$ is positive and
normalized, and $\E B_e=0$ makes every coordinate uniform. The channel
is the posterior bond mean, $h_e(u)=\E[B_e\mid U_e=u]$. The dimension
$d=|E|$ counts observed coordinates; an $L\times L$ vertex grid has $d=2L(L-1)$ edges. Define
\[
 \eta_e=\int_0^1\phi(u)h_e(u)\,du,\qquad
 \xi_e=\int_0^1h_e(u)^2\,du,\qquad
 \phi(u)=\sign(2u-1).
\]
For an edge set $F$, put $\phi_F(u_F)=\prod_{f\in F}\phi(u_f)$ and
$\theta_F=\E\phi_F(U_F)$; we omit the argument when evaluating at the readings.
The \emph{reliability} $\eta_e=\E[B_e\phi(U_e)]$ is the correlation
between a bond and its thresholded reading. The \emph{energy} $\xi_e$
is the average of the squared posterior bond mean.
Our running example is the clipped
ramp
\[
 h(u)=a\ramp_\tau(u),\qquad
 \ramp_\tau(u)=\clip\bigl((2u-1)/\tau,-1,1\bigr),\qquad
 0<a<1,\ 0<\tau\leq1,
\]
where $\clip(x,\ell,r)=\max\{\ell,\min\{x,r\}\}$. This channel is $(2a/\tau)$-Lipschitz with
reliability $\eta=a(1-\tau/2)$. If each reading is the
probability-integral transform of a variable with a known continuous
margin, $p_h$ is the copula of a model with arbitrary margins.

\subsection{Three Costs on One Class}\label{sec:three-costs}
We compare learning, evidence queries and circuit approximation on one
class of laws. Known grids and margins isolate the cost of learning dependence.

The amplitude cap keeps emission densities away from zero; the Lipschitz
bound controls spline approximation error. A positive reliability bound,
called an \emph{anchor}, keeps the short-cycle calibration moments away
from zero (Lemma~\ref{lem:witness-identities}). Section~\ref{sec:minimax}
later separates this calibration requirement from density estimation.

\begin{samepage}
\begin{definition}[Anchored class]\label{def:anchored}
On a known $L\times L$ vertex grid, spins are independent and uniform,
margins are known (readings are uniform on $[0,1]$), and channels are
centered with $|h_e|\leq a<1$. Such a law is \emph{anchored} if in
addition $\eta_e\geq\kappa>0$ on every edge. Let
$\mathcal P_L(a,R,\kappa)$ be the anchored laws with $R$-Lipschitz
channels, and $\mathcal P^0_L(a,R)$ the laws with $R$-Lipschitz channels
without an anchor requirement. The constants $a,R,\kappa$ are fixed as $L$ grows.
\end{definition}
\end{samepage}

Write $\eta_*(a,R)$ for the largest feasible reliability, that of
$\clip(R(u-\tfrac12),-a,a)$. Its formula is in
Appendix~\ref{app:model-details}; the anchored class is nonempty exactly
when $\kappa\leq\eta_*(a,R)$.

\begin{definition}[Admissible circuits]\label{def:circuit}
Fix $0<\alpha<1$ independently of $d$. A nonnegative circuit over $E$ is
a finite directed acyclic graph whose leaves are nonnegative measurable functions of their declared scopes and whose internal nodes are
products or nonnegative weighted sums; its root is a probability density
relative to $\mu$ \citep{darwiche2003,poon2011,choi2020pc}.
Each node has a declared \emph{scope} $\subseteq E$,
including variables on which its function is constant: leaf scopes have
at most $\lfloor\alpha d\rfloor$ coordinates, internal scopes are unions
of child scopes, and the root scope is $E$. The circuit is
\emph{decomposable} if product children have disjoint scopes and
\emph{smooth} if sum children share the sum's scope. Its size $M$ counts
nodes after products are made binary; it excludes computation inside leaves.
Sum children may overlap in support, and no common partition tree
(vtree) is required.
Put $\zeta_\alpha=\min\{1/3,1-\alpha\}$; a
partition $E_1\mid E_2$ of $E$ is \emph{balanced} if $\min\{|E_1|,|E_2|\}\geq\zeta_\alpha d$.
\end{definition}

The leaf restriction prevents one leaf from storing the full density,
while allowing scopes that grow linearly with $d$. In the representation
comparison, we let the circuit depend on the target law; no learning
algorithm is imposed.

\begin{theorem}[Three costs on the grid]\label{thm:separation}
Fix $0<a<1$, $R>0$, $0<\kappa<\eta_*(a,R)$ and $0<\alpha<1$.
For $L\geq4$, the class $\mathcal P_L(a,R,\kappa)$ has the following
costs, with $d=2L(L-1)$ and constants depending only on $a,R,\kappa,\alpha$.
\begin{enumerate}
\setlength{\itemsep}{0pt}
\item \emph{Learning.} For $N$ vectors, minimax squared Hellinger risk
is $\Theta(\min\{1,dN^{-2/3}\})$. Each directed KL
minimax risk is $\Theta(dN^{-2/3})$ for $N\geq d^{3/2}$. Expected TV
error $\delta<1/8$ takes between $cd^{3/2}$ and $C\delta^{-3}d^{3/2}$
vectors. Given the anchor $\kappa$, a spline estimator meets the upper
bounds at polynomial arithmetic cost
(Theorems~\ref{thm:learning} and~\ref{thm:minimax}).
\item \emph{Queries.} Given the local evidence integrals, which are
explicit for spline channels, every factorized point, interval and
missing-coordinate evidence query costs $O(d^{3/2})$ exact arithmetic
operations and one root extraction (Proposition~\ref{prop:queries}).
\item \emph{Representation.} For every law in the class, the minimum
size of a decomposable circuit of Definition~\ref{def:circuit} within
fixed TV distance $\delta\in(0,1)$ is $\exp(\Theta(\sqrt d))$ as
$L\to\infty$. The dimension threshold may depend on $\delta$
(Corollary~\ref{cor:grid-circuit}).
\end{enumerate}
\end{theorem}

The estimator stores spline coefficients and answers queries through a
planar determinant (Section~\ref{sec:queries}). This computation need not
be a nonnegative decomposable circuit, so the representation bound does
not restrict its learning or query cost. Appendix~\ref{app:model-details}
proves the theorem; graph discovery, unknown margins and bit complexity
are excluded. Figure~\ref{fig:costs} shows the graph structures behind these costs.

\subsection{Loop Expansion and Low-Order Blindness}\label{sec:loop-expansion}
\begin{figure}[htbp]
\centering
\includegraphics[width=\linewidth]{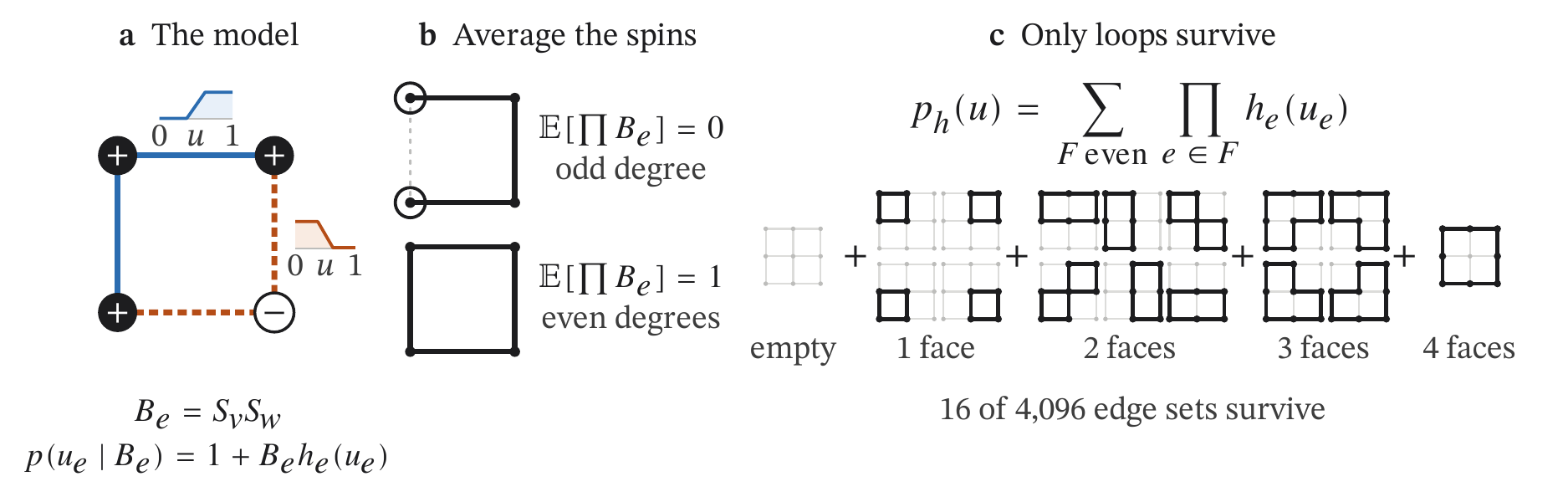}
\caption{Even edge sets carry the dependence. (a) Independent spins
generate bonds and noisy readings; blue solid bonds agree and orange
dashed bonds disagree. Emissions use the ramp with $a=0.95$, $\tau=1/4$.
(b) A bond product has mean zero if a vertex has odd degree (circled),
and one otherwise. (c) The four faces of the $3\times3$ grid generate
16 even sets among 4,096 edge subsets. Marginalization deletes edges;
forests retain only the constant term.}
\label{fig:loops}
\end{figure}

We now identify which interactions survive averaging over the latent spins.
For $F\subseteq E$ let $\partial F$ be its set of odd-degree vertices.
The \emph{loop polynomial}, the high-temperature expansion of the
zero-field Ising model \citep{vanderwaerden1941} and the loop series at the
paramagnetic belief-propagation fixed point \citep{chertkov2006loop}, sums
over even edge sets,
\[
 L_G(w)=\sum_{F\subseteq E:\,\partial F=\varnothing}\ \prod_{e\in F}w_e .
\]
A spin monomial has zero mean unless every vertex occurs an even number
of times, so expanding~\eqref{eq:density} gives $p_h(u)=L_G(h(u))$ with
$h(u)=(h_e(u_e))_e$. Since $\int_0^1h_e=0$, integrating out $u_e$ removes
every term with $e\in F$, so the marginal is $L_{G\setminus e}$ at the
remaining channels. A forest has no nonempty even subset, so every forest
marginal is uniform: any $\girth(G)-1$ coordinates are
independent, and the square grid is three-wise independent.
On one square
(Figure~\ref{fig:loops}) $p_h=1+h_1h_2h_3h_4$, and three independent readings predict the fourth:
\[
 p(u_4\mid u_1,u_2,u_3)=1+\{h_1(u_1)h_2(u_2)h_3(u_3)\}\,h_4(u_4).
\]
The cycle space $\mathcal Z(G)$, the set of even edge sets, indexes the
channel products that determine the law.

Uniform forest marginals imply a lower bound on approximation error.
If $\log q$ is a
sum of forest-supported terms, then
$\KL(p_h\Vert q)=\KL(p_h\Vert\mu)+\KL(\mu\Vert q)$
(Proposition~\ref{prop:forest-blindness}). Thus independence is optimal
within any such class that contains it. This covers visible tree
models and a Gaussian in probit coordinates, but not hidden trees or
mixtures.

Products indexed by distinct even sets are orthogonal in $L^2(\mu)$,
because an edge in their symmetric difference contributes one centered
factor. Hence
\begin{equation}\label{eq:loop-energy}
 \int p_h^2\,d\mu=L_G(\xi),\qquad
 \chi^2(p_h\Vert\mu)=L_G(\xi)-1,\qquad
 \KL(p_h\Vert\mu)\leq\log L_G(\xi),
\end{equation}
where $\chi^2(p\Vert q)=\int(p/q-1)^2\,dq$ and the last bound is Jensen's
inequality. Dependence can be weak: on a square,
$\chi^2(p_h\Vert\mu)=\prod_{i=1}^4\xi_i\leq a^8$.

\noindent\emph{Binary readings and code models.}
Integrating over half-intervals gives
$\Pr(\phi(U_e)=x\mid B_e=b)=(1+bx\eta_e)/2$.
Thus the signs are a uniform graph cut with independent flips of
probability $(1-\eta_e)/2$. Planar duality maps cuts to even subgraphs,
giving the noisy-cycle construction in surface-code models
\citep{dennis2002,kumar2026unlearnable}. Exact identification requires
matching boundaries and logical sectors; our graph and spin law fix both.

\subsection{Evidence Queries Reduce to a Planar Ising Model}\label{sec:queries}
Factorized evidence changes only edge interactions; it introduces no
single-spin field and preserves the planar graph. We can therefore use
a planar Ising determinant to evaluate the likelihood.
An observation of coordinate $e$ replaces its channel value by a signed
weight, contributing a factor
$\bar\ell_e(1+t_eB_e)$ with $\bar\ell_e>0$ and $|t_e|\leq a_e$.
Missing, point, and interval evidence $U_e\in[u_-,u_+]$, with
$0\leq u_-<u_+\leq1$, use, respectively,
\[
 (\bar\ell_e,t_e)=(1,0),\quad (1,h_e(u_e)),\quad
 \Bigl(u_+-u_-,\frac{1}{u_+-u_-}\int_{u_-}^{u_+}h_e(u)\,du\Bigr).
\]
Let $p_h(\mathcal E)$ be the marginal likelihood of an evidence
specification $\mathcal E$: a density in point-observed coordinates,
integrated over its intervals. Summing over spins
(Appendix~\ref{app:model-details}) gives
\begin{equation}\label{eq:evidence-zero}
 p_h(\mathcal E)=\Bigl(\prod_e\bar\ell_e\Bigr)L_G(t).
\end{equation}

On a straight-line planar embedding, the Kac--Ward
identity \citep{kacward1952,cimasoni,lis2016} states
\begin{equation}\label{eq:kw}
 L_G(w)^2=\det\bigl(I-T\diag(w)\bigr).
\end{equation}
Here $T$ indexes directed edges: $T_{e,f}=\exp\{\mathrm i\angle(e,f)/2\}$
if $e,f$ concatenate without reversal, and zero otherwise, with signed
turn $\angle(e,f)\in(-\pi,\pi)$. The diagonal repeats each edge weight
on both orientations. Arithmetic acts on real and imaginary parts,
with the embedding's turn phases as input constants.
For real $|t_e|<1$, $L_G(t)>0$ selects the positive square root,
including for negative couplings. The positive-definite Gram matrix
of $I-T\diag(t)$ has determinant $L_G(t)^4$. Generalized nested
dissection evaluates it in $O(d^{3/2})$ exact arithmetic operations;
one fourth root gives $L_G(t)$ (Proposition~\ref{prop:queries}).
Triangular solves give conditional bond means, and unconditional sampling
takes $O(d)$ expected channel evaluations (Appendix~\ref{app:model-details}).
The zero field avoids general-field hardness \citep{barahona1982}.
Matchgate methods also give exact surface-code decoding with independent
bit and phase errors and noiseless syndrome measurements \citep{bravyi2014}.

These are continuous virtual-evidence queries. The completeness result
of \citet[Proposition~5.4]{broadrick2025} concerns functions on Boolean
domains and permits general arithmetic circuits, but its real-RAM model excludes root
extraction. We count the root explicitly and
bound only nonnegative decomposable circuits.
\section{The Cost of Positive Representation}\label{sec:representation}
\begin{figure}[htbp]\centering
\includegraphics[width=\linewidth]{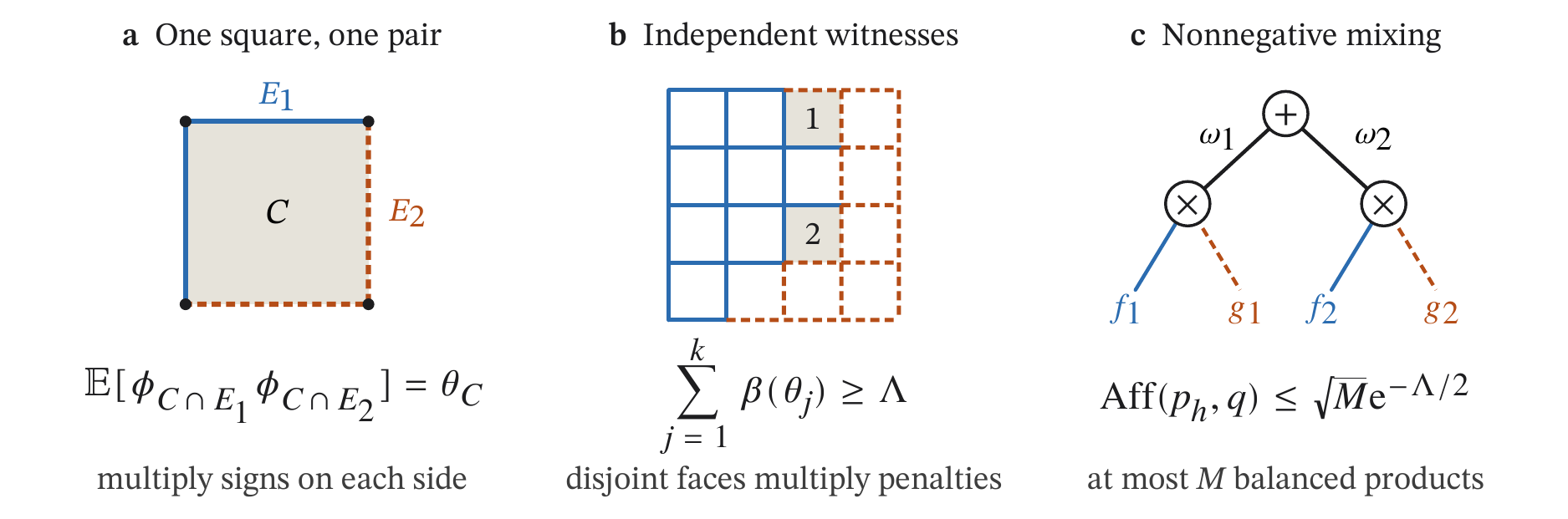}
\caption{The barrier for independent spins. (a) Blue solid and orange
dashed edges split the observed coordinates. On a mixed face $C$,
the two half-sign products are fair with correlation
$\theta_C=\prod_{e\in C}\eta_e$. (b) Vertex-disjoint faces give
independent pairs, whose affinity exponents add. Every balanced cut
has at least $k_{L,\alpha}$ such faces (Lemma~\ref{lem:grid}).
(c) A smooth decomposable nonnegative circuit of size $M$, with leaves
of at most $\lfloor\alpha d\rfloor$ coordinates, is a mixture of at most
$M$ balanced products. Each may use a different cut; mixing increases
the common affinity bound by at most $\sqrt M$
(Proposition~\ref{prop:balanced-affinity}).}
\label{fig:representation-witness}
\end{figure}

The law has a compact nonnegative factor graph with one binary latent
spin per vertex. We ask what decomposability costs.
A square's readings split into two nonempty uniform groups, but their
sign products have correlation $\theta_C$. A product across that split
cannot capture this dependence. We count such crossing cycles to bound
circuit size (Theorem~\ref{thm:witness-barrier}).
We use Definition~\ref{def:circuit} with $d\geq2$ and $\alpha d\geq1$
(Figure~\ref{fig:representation-witness}).

\subsection{A Balanced Product-Affinity Criterion}
We compare the target with product laws across balanced partitions.
Low overlap with every such product gives the following circuit-size bound.

\begin{samepage}
For $0<\zeta\leq1/2$ and a density $p$ relative to a product reference
probability, define
\[
 \mathcal I_\zeta(p)=
 \inf_{\zeta d\leq|E_1|\leq(1-\zeta)d}
 \Bigl[-2\log\sup_{f_{E_1},g_{E_2}}\Aff(p,f_{E_1}\otimes g_{E_2})\Bigr],
\]
\end{samepage}
where $E_1\subseteq E$, $E_2=E\setminus E_1$ and $f,g$ are normalized
densities. This measures dependence across the easiest balanced cut,
allowing arbitrary dependence within each side. A large value therefore
forces small affinity with every balanced product law.
At a fixed cut it is
$\inf_{f_{E_1},g_{E_2}}D_{1/2}(p\Vert f_{E_1}\otimes g_{E_2})$, with
R\'enyi divergence $D_{1/2}=-2\log\Aff$, a continuous
counterpart, in nats, of the doubly minimized R\'enyi dependence of
\citet[Eq.~(1)]{lapidoth2019}. For $\theta\in[-1,1]$, define
\[
 \beta(\theta)=-\log\frac{1+\sqrt{1-\theta^2}}2,
\]
which is increasing in $|\theta|$, with
$\beta(\theta)\geq\theta^2/4$ since $-\log x\geq1-x$ and
$\sqrt{1-\theta^2}\leq1-\theta^2/2$.
The function $\beta$ measures the contribution of a correlated binary pair
to the affinity bound below.

\begin{proposition}[Balanced affinity]\label{prop:balanced-affinity}
Let $q$ be a smooth decomposable circuit of Definition~\ref{def:circuit}
with $M$ nodes. If $\mathcal I_{\zeta_\alpha}(p)\geq\Lambda$, then
\begin{equation}\label{eq:affinity}
 \Aff(p,q)\leq\min\{1,\sqrt M e^{-\Lambda/2}\}.
\end{equation}
Consequently $\KL(p\Vert q),\KL(q\Vert p)\geq\Lambda-\log M$ and
$H^2(p,q),\TV(p,q)\geq1-\min\{1,\sqrt Me^{-\Lambda/2}\}$.
\end{proposition}

Lemma~\ref{lem:balance} writes $q$ as a mixture of at most $M$ balanced product terms, adapting
\citet[Theorem~38]{martens} and \citet[Theorem~6]{bova2016}; square-root subadditivity
limits the affinity gain from mixing to a factor $\sqrt M$, even when
the terms use different cuts (Appendix~\ref{app:affinity-proof}).
\citet[Lemma~3.3]{raz2011} give a related monotone multilinear decomposition.
We summarize readings on each side by signs, using only data from that
side. The next lemma combines these pairs when their covariance stays
bounded away from zero after earlier pairs are observed. The pairs need
not be independent. The binary-pair calculation is in
Appendix~\ref{app:dependent-witnesses}.

\begin{lemma}[Conditional crossing pairs]\label{lem:conditional-affinity}
For each balanced cut, suppose separate Markov kernels on its sides,
using independent auxiliary randomness, produce pairs
$Z_j=(X_j,Y_j)\in\{\pm1\}^2$, $j=1,\ldots,k$. Write
$Z_{<j}=(Z_1,\ldots,Z_{j-1})$ and assume
\[
 |\operatorname{Cov}(X_j,Y_j\mid Z_{<j}=z_{<j})|\geq c_j\in[0,1]
 \quad\text{for every history of positive probability}.
\]
At a fixed cut, the affinity with any product density is at most
$\exp\{-\tfrac12\sum_j\beta(c_j)\}$. If
$\sum_j\beta(c_j)\geq\Lambda$ for every cut, then
$\mathcal I_{\zeta_\alpha}(p)\geq\Lambda$.
\end{lemma}

Under a product law, successive pairs remain conditionally product.
Summing out pairs in reverse order multiplies the one-pair affinity bounds
(Appendix~\ref{app:dependent-witnesses}).

\subsection{Short Cycles Across Every Balanced Cut}\label{sec:witness-barrier}
A \emph{crossing cycle} has edges on both sides of a partition.
Let $k_{\alpha,s}(G)$ be the minimum, over balanced edge partitions, of
the largest number of vertex-disjoint crossing simple cycles of length
at most $s$. This count depends only on the graph. To treat independent
and dependent spins together, let $\pi$ be any spin law and put
\[
 p_{\pi,h}(u)=\E_\pi\prod_{e=\{v,w\}\in E}(1+S_vS_wh_e(u_e)),\qquad
 \epsilon_\pi=\max_{v\in V}\operatorname*{ess\,sup}
 |\E_\pi[S_v\mid S_{V\setminus\{v\}}]|.
\]
The \emph{finite-energy condition} $\epsilon_\pi<1$ bounds single-spin
conditionals away from zero and one. Independent uniform spins have
$\epsilon_\pi=0$ and $p_{\pi,h}=p_h$.

\begin{theorem}[Witness barrier]\label{thm:witness-barrier}\label{thm:dependent-barrier}
Let $G$ be a finite simple graph with $d\geq2$ edges, centered channels
$|h_e|\leq a<1$ and $\eta_e\geq\kappa>0$, and a spin prior with
$\epsilon_\pi<1$. Fix $0<\alpha<1$ with $\alpha d\geq1$ and an integer
$s\geq3$. Then
\[
 \mathcal I_{\zeta_\alpha}(p_{\pi,h})\geq
 \Lambda_\pi:=k_{\alpha,s}(G)\beta((1-\epsilon_\pi^2)\kappa^s)
 \geq\tfrac14 k_{\alpha,s}(G)(1-\epsilon_\pi^2)^2\kappa^{2s}.
\]
Every smooth decomposable circuit of Definition~\ref{def:circuit} satisfies
\eqref{eq:affinity} with $\Lambda=\Lambda_\pi$.
\end{theorem}

Combining the theorem with Proposition~\ref{prop:balanced-affinity}
gives size bounds. For either KL direction
at most $\varepsilon$, or TV at most $\delta<1$, respectively,
\begin{equation}\label{eq:size}
 \log M\geq\Lambda-\varepsilon,\qquad
 \log M\geq\Lambda+2\log(1-\delta).
\end{equation}
Here $\Lambda=\Lambda_\pi$; for independent uniform spins,
$\Lambda=k_{\alpha,s}(G)\beta(\kappa^s)$.
Smoothing uses at most $4dM^2$ nodes and gives the nonsmooth bound
(Appendix~\ref{app:smoothing}).

The sign products on the two parts of a crossing cycle share a latent
sign whose conditional mean is bounded by $\epsilon_\pi$.
Independent emission noises then give the
covariance bound (Appendix~\ref{app:dependent-witnesses}).

A partition of disjoint squares can keep each square whole and avoid
crossing cycles. Grid faces overlap, which forces order $\sqrt d$
vertex-disjoint crossing faces at every balanced cut (Lemma~\ref{lem:grid}).

\begin{corollary}[Grid: decomposable positive circuits]\label{cor:grid-circuit}
On the $L\times L$ grid, $L\geq2$, with independent
uniform spins, channels as in Theorem~\ref{thm:witness-barrier} and
$0<\alpha<1$ with $\alpha d\geq1$, put
\begin{equation}\label{eq:count}
 k_{L,\alpha}=\min\{k\in\{0,1,\dots\}:4k^2+3k\geq(L-1)(\zeta_\alpha L-1)\},
 \qquad \Lambda_L=k_{L,\alpha}\beta(\kappa^4).
\end{equation}
Then $k_{\alpha,4}(G)\geq k_{L,\alpha}$, so~\eqref{eq:affinity},
\eqref{eq:size} and the bound without smoothness hold with
$\Lambda=\Lambda_L$, and $\Lambda_L\geq\kappa^8\sqrt{\zeta_\alpha}\,L/12$
for $L\geq4/\zeta_\alpha$. At fixed $a,\kappa,\alpha$ and TV tolerance
$\delta\in(0,1)$,
the minimum circuit size is $\exp(\Theta(L))=\exp(\Theta(\sqrt d))$, and
every polynomial-size circuit sequence has TV tending to one and both KL
divergences diverging.
\end{corollary}

A row-frontier circuit with $2^L$ spin states and singleton leaves gives
the matching upper bound (Appendix~\ref{app:representation-proofs}).

For a zero-field Ising prior
$\pi(s)\propto\exp\{\sum_{e=\{v,w\}\in E}J_e^{(0)}s_vs_w\}$,
$\epsilon_\pi\leq\tanh(\max_v\sum_{e\ni v}|J_e^{(0)}|)$.
The learning and uniform-forest guarantees need not survive such a prior.

\subsection{Average versus Global Error}\label{sec:average-global}
The size bound concerns global error. A small circuit can still have
low average log loss, as the following construction shows.
We delete edges between vertex tiles of side at most $b$ and give them
independent uniform readings. The tiled law $q_b$ has a smooth
decomposable nonnegative circuit with singleton leaves and $O(d2^b)$
nodes. It satisfies $\KL(p_h\Vert q_b)\leq a^2d/b$, while for anchored
laws and $2\leq b<L$,
$\TV(p_h,q_b)\geq1-\exp\{-\beta(\kappa^4)(L-1)^2/(8b)\}$
(Appendix~\ref{app:tile-proof}). Thus
$b=\lfloor\log_2d\rfloor$ gives polynomial size and vanishing
per-coordinate KL, while TV tends to one.

We next quantify how larger leaves reduce the KL cost. Write
$\mathcal C_{M,\ell}$ for normalized smooth decomposable nonnegative
circuits with at most $M$ nodes after binarizing products and declared
leaf scopes of at most $\ell$ coordinates. Thus
$\mathcal C_M=\mathcal C_{M,1}$ is the univariate-leaf class.
Leaves are arbitrary nonnegative measurable functions; node counts exclude
their evaluation cost. Write $[x]_+=\max\{x,0\}$.

\begin{theorem}[KL cost and leaf scope]\label{thm:kl-size}
Let $p_h$ be a grid law with independent uniform spins, centered
channels $|h_e|\leq a<1$ and $\eta_e\geq\kappa>0$. For
$q\in\mathcal C_{M,\ell}$ and any integer $12\leq b\leq L$
with $\ell\leq4b(b-1)/3$,
\begin{equation}\label{eq:kl-blocks}
 \min\{\KL(p_h\Vert q),\KL(q\Vert p_h)\}
 \geq\left\lfloor\frac Lb\right\rfloor^2
 \left[\frac{b\beta(\kappa^4)}{3\sqrt3}-\log M\right]_+.
\end{equation}
Fix $r>1$ and $0<\alpha<2/3$. Uniformly over these laws and
integers $1\leq\ell\leq\lfloor\alpha d\rfloor$ as $L\to\infty$,
\begin{equation}\label{eq:kl-rate}
 \inf_{q\in\mathcal C_{\lfloor d^r\rfloor,\ell}}\KL(p_h\Vert q)
 \asymp
 \inf_{q\in\mathcal C_{\lfloor d^r\rfloor,\ell}}\KL(q\Vert p_h)
 \asymp\frac d{\max\{\log d,\sqrt\ell\}}.
\end{equation}
The constants and the dimension threshold depend only on $a,\kappa,r,\alpha$.
The rate~\eqref{eq:kl-rate} also holds without smoothness.
\end{theorem}

For the lower bound, we choose disjoint subgrids so that each leaf contains
at most two-thirds of any subgrid's readings. Their reading laws are
independent, so we can add their KL costs. Tiled circuits give the matching upper bound
(Appendix~\ref{app:kl-size}).
Leaves of at most $(\log d)^2$ coordinates do not improve the order
$d/\log d$; larger leaves reduce it by storing larger tiles.
The proof needs $\alpha<2/3$; at $\ell=d$ one leaf represents the law.
For nonsmooth circuits, smoothing replaces $\log M$ in
\eqref{eq:kl-blocks} by $\log(4dM^2)$ (Appendix~\ref{app:smoothing}).
The finite bound is uninformative at the sizes in Section~\ref{sec:experiments}.

\subsection{Joint Prediction of Missing Readings}\label{sec:block-prediction}
Conditioning can remove dependence, so joint-density lower bounds need
not apply to prediction of missing readings.
For a fixed hidden set $S\subseteq E$ and $O=E\setminus S$, let
$p_{S\mid u_O}$ denote the conditional density of $U_S$ given $U_O=u_O$. Write
$\mathcal R_S(p,q)=\E_{p_O}\KL(p_{S\mid U_O}\Vert q_{S\mid U_O})$.
We set the conditional loss to $\infty$ where $q_O=0$, so
$\mathcal R_S(p,q)=\infty$ if this evidence has positive $p_O$ probability.

We consider three masks, chosen independently of readings. The patch
leaves untouched hidden subgrids; the other masks spread observations
across the grid and require us to track dependence after conditioning.
Index vertices by $(i,j)$ from zero.
\begin{enumerate}
\setlength{\itemsep}{0pt}
\item A patch: $S$ is the induced edge set of a centered
$\lfloor\rho L\rfloor\times\lfloor\rho L\rfloor$ vertex square,
for fixed $0<\rho<1$.
\item A staggered mask, with $L$ even:
$O$ consists of horizontal edges $(i,j)(i,j+1)$ with $i+j$ even,
and boundary vertical edges $(i,j)(i+1,j)$ with
$j\in\{0,L-1\}$ and $i$ even; set $S=E\setminus O$.
\item An aligned matching, with $L$ even: observe
$O=\{(i,2j)(i,2j+1):0\leq i<L,\ 0\leq j<L/2\}$ and hide $S=E\setminus O$.
\end{enumerate}

\begin{samepage}
\begin{theorem}[Joint prediction under fixed masks]\label{thm:block-prediction}
Under the channel assumptions of Theorem~\ref{thm:kl-size}, fix $r>1$
and any of the three masks above. Uniformly for
$1\leq\ell\leq\lfloor(\log d)^2\rfloor$ as $L\to\infty$,
through even integers for the last two masks,
\[
 \inf_{q\in\mathcal C_{\lfloor d^r\rfloor,\ell}}\mathcal R_S(p_h,q)
 \asymp\frac d{\log d}.
\]
Constants and the dimension threshold depend only on $a,\kappa,r$,
and on $\rho$ for the patch.
\end{theorem}
\end{samepage}

Each mask hides $\Theta(d)$ coordinates. Thus the optimal total
prediction loss grows, while its loss per hidden coordinate is
$\Theta(1/\log d)$ and vanishes.
Smoothness is unnecessary (Appendix~\ref{app:kl-size}). For almost every
evidence value, untouched subgrids give the patch lower bound and hidden
six-edge cycles give the staggered bound.
Under the aligned matching, hidden edges form disjoint narrow ladders.
Observed edges reconnect them, but a reading near a channel
zero gives little information about its bond. The resulting crossing-cycle
strengths can therefore vanish. On each subgrid, we lower-bound the costs
across all of the circuit's fixed cuts with probability at least $1/2$,
then average over evidence (Appendix~\ref{app:kl-size}).
All three upper bounds average over evidence.
\section{Witness Geometry and Learning}\label{sec:learning}
The representation bounds use cycles crossing a partition. Learning uses
intersecting cycles to calibrate local channels from observed moments.
We identify the observable channel products, then relate their error to
divergence between laws.

\subsection{Blocks and Witnesses}\label{sec:witnesses}
Let $G$ be a finite simple graph with centered channels bounded by
$a<1$. Vertex sign flips $h_{vw}\mapsto\sigma_v\sigma_wh_{vw}$,
$\sigma_v\in\{-1,1\}$, preserve every even-set product; positive
reliabilities remove this sign ambiguity. Bridges lie on no cycle and have no
effect on the law. Non-bridge edges are \emph{in series} if they lie
on the same simple cycles. We call these classes \emph{observable blocks}
(or blocks), distinct from prediction blocks and biconnected graph blocks.
An even set contains all of a block or
none of it, so with
\[
 h_A=\bigotimes_{e\in A}h_e,\qquad B_A=\prod_{e\in A}B_e,\qquad
 a_A=a^{|A|},
\]
the law depends on the channels only through the block channels,
\begin{equation}\label{eq:block-density}
 p_h(u)=\E\prod_A\{1+B_Ah_A(u_A)\},\qquad |h_A|\leq a_A .
\end{equation}
Rescaling a block's channels by positive factors with product one
preserves the law, so even infinite data need not identify individual
channels. We estimate the block products $\mathcal O(h)=(h_A)_A$, the
\emph{anchored observable coordinates}: between anchored laws they are
exactly what the law determines
(Appendix~\ref{app:graph-identification}). On the $L\times L$ grid, $L\geq3$, the blocks are the $d-8$ scalar
edges away from the corners and the four corner pairs. A corner pair
$A=\{i,j\}$ has channel $h_A=h_i\otimes h_j$, obtained by averaging
out its degree-two spin.

A block that is itself a cycle has $B_A=1$ and contributes an
independent factor $1+h_A(u_A)$ to~\eqref{eq:block-density}
(Remark~\ref{rem:cycle-blocks}). Every other block is a proper subset of
each simple cycle through it and hence a forest, so its bonds are
independent fair signs, $U_A$ is uniform and
$\E[B_A\mid U_A]=h_A(U_A)$.

A local regression gives a block channel multiplied by an unknown scale.
Cycle moments are products of reliabilities, so their logarithms add.
A \emph{witness} chooses a linear combination that isolates the required
scale (Lemma~\ref{lem:witness-identities}).

\begin{definition}[Witnesses]\label{def:witness}
Let $A$ be a block that is not a cycle. A \emph{witness} for $A$ is a
simple cycle $C_A\supseteq A$ with real weights $\vartheta_{A,F}$ on a
finite family $\mathcal F_A$ of even sets such that
\begin{equation}\label{eq:witness}
 \mathbf 1_{C_A\setminus A}=\sum_{F\in\mathcal F_A}\vartheta_{A,F}\mathbf 1_F
 \qquad\text{in }\R^E ,
\end{equation}
where $\mathbf 1_F$ is the edge-incidence vector of $F$.
Its \emph{window} is $W_A=C_A\cup\bigcup_{F\in\mathcal F_A}F$, and for an
anchor $\kappa\in(0,1]$ its \emph{constant} is
\[
 \Gamma_A=\kappa^{-|C_A\setminus A|}
 +a_A\sum_{F\in\mathcal F_A}|\vartheta_{A,F}|\,\kappa^{-|F|}.
\]
A \emph{witness system with $\chi$ colors} gives every block a witness
and one of $\chi$ colors such that blocks of the same color have
vertex-disjoint windows.
\end{definition}

\begin{lemma}[Witness identities]\label{lem:witness-identities}
Let $p_h$ be a law~\eqref{eq:density} with centered channels bounded by
$a<1$ and $\eta_e>0$ on every non-bridge edge. Every even set $F$ has
$\theta_F=\prod_{f\in F}\eta_f$. If $A$ has a witness, then with
$\gamma_A=\prod_{f\in C_A\setminus A}\eta_f$,
\begin{equation}\label{eq:block-regression}
 \E\bigl[\phi_{C_A\setminus A}\bigm|U_A=u_A\bigr]=\gamma_Ah_A(u_A),
 \qquad
 \log\gamma_A=\sum_{F\in\mathcal F_A}\vartheta_{A,F}\log\theta_F .
\end{equation}
Hence $h_A$ is a function of the law of $U_{W_A}$.
\end{lemma}

Witnesses exist (Appendix~\ref{app:graph-identification}). The constant
$\Gamma_A$ bounds how errors in the observed moments and regression
affect the recovered block (Theorem~\ref{thm:witness-geometry}); short
cycles and small weights reduce this bound.

On the grid, faces and six-cycles around adjacent faces supply all
calibration moments. For an interior scalar edge shared by faces $C,C'$,
$\gamma_A=\sqrt{\theta_C\theta_{C\triangle C'}/\theta_{C'}}$,
where $\triangle$ denotes symmetric difference.
Boundary edges and corner pairs use the ratios in
Lemma~\ref{lem:identify} (Appendix~\ref{app:learning-proofs}). Every
block has a witness with
\[
 \Gamma_A\leq\Gamma:=\kappa^{-3}+3a\kappa^{-4}+\tfrac32a\kappa^{-6}.
\]

Grid windows use at most $13$ edges and $24$ colors
(Lemma~\ref{lem:witnesses}). Windows of one color have independent reading
vectors, which lets the next theorem combine their information with
constants independent of grid size.
\subsection{Distance in Observable Coordinates}\label{sec:global-geometry}
The upper bound below converts local block error into joint KL.
The lower bound shows when close laws must have close observable
channels and therefore controls identification.

\begin{theorem}[Witness geometry]\label{thm:witness-geometry}
Let $G$ be a finite simple graph with no cycle block and a witness system with $\chi$ colors
and constants $\Gamma_A$ at anchor $\kappa$ (Definition~\ref{def:witness}).
Let $p_h,p_{h'}$ be laws~\eqref{eq:density} on $G$ with centered channels
bounded by $a<1$ and reliability at least $\kappa$ on non-bridge edges.
For either directed divergence
$D\in\{\KL(p_h\Vert p_{h'}),\KL(p_{h'}\Vert p_h)\}$,
\begin{equation}\label{eq:global-divergence}
 \sum_A\frac{\|h_A-h'_A\|_2^2}{8\chi\,\Gamma_A^2}
 \ \leq\ -\log\Aff(p_h,p_{h'})\ \leq\ \tfrac12D\ \leq\
 \sum_A\frac{\|h_A-h'_A\|_2^2}{4(1-a_A^2)},
\end{equation}
and $H^2(p_h,p_{h'})\leq\sum_A\|h_A-h'_A\|_2^2/\{8(1-a_A^2)\}$. The
upper bounds use neither the witnesses nor the reliabilities.
\end{theorem}

We share latent bonds for the upper bound. For the lower bound, we group
witness windows into independent color classes, as in dependency-graph
arguments \citep{janson2004}, and multiply their affinities
(Appendix~\ref{app:global-geometry}).
No smoothness is needed; Remark~\ref{rem:cycle-blocks} treats cycle blocks.

Write $\Delta^2(f,f')=\sum_A\|f_A-f'_A\|_2^2$ for block collections and
$\Delta(h,h')=\Delta(\mathcal O(h),\mathcal O(h'))$ for channels.
On grids, this sums errors in scalar channels and corner products.

\begin{corollary}[Grid geometry]\label{cor:grid-geometry}
Let $p_h,p_{h'}$ be anchored laws on the $L\times L$ grid, $L\geq4$,
with amplitude bound $a$ and anchor $\kappa$, and put
$c_\Gamma=(8\cdot24\,\Gamma^2)^{-1}$ with $\Gamma$ as in
Lemma~\ref{lem:identify}. For either directed divergence $D$,
\[
 c_\Gamma\,\Delta^2(h,h')\leq-\log\Aff(p_h,p_{h'})\leq\tfrac12D
 \leq\frac{\Delta^2(h,h')}{4(1-a^2)} ,
\]
so $1-e^{-c_\Gamma\Delta^2}\leq H^2(p_h,p_{h'})\leq\TV(p_h,p_{h'})$ and
$H^2(p_h,p_{h'})\leq\min\{1,\Delta^2/(8(1-a^2))\}$.
\end{corollary}

To cover fitted corner functions that need not factor into edge
channels, average out each degree-two corner spin (the contracted grid).
Call a block collection $f$ \emph{capped} by
$a_{\mathrm{in}}<1$ if $|f_A|\leq a_{\mathrm{in}}$, scalar blocks are
centered and corner blocks are centered in each argument. Its law
$p_f(u)=\E\prod_A\{1+B_Af_A(u_A)\}$ is positive and normalized, and
$p_h=p_{\mathcal O(h)}$.

\begin{lemma}[Shared-spin coupling]\label{lem:coupling}
If $f,f'$ are capped by $a_{\mathrm{in}}<1$, then for either directed
divergence $D$ between $p_f$ and $p_{f'}$,
$D\leq\Delta^2(f,f')/\{2(1-a_{\mathrm{in}}^2)\}$.
\end{lemma}

This anchor-free bound controls the fitted estimator's joint KL.
Weak reliability affects the reverse comparison: the lower
constant $c_\Gamma$ decays as $\kappa^{12}$. The next result shows that
this exponent cannot be lowered. Let $c^\star(a,\kappa)$ be the largest constant in
place of $c_\Gamma$ for which the left inequality of
Corollary~\ref{cor:grid-geometry} holds for all pairs of anchored laws
with amplitude bound $a$ and anchor $\kappa$ on every grid with
$L\geq4$.

\begin{proposition}[Sharpness of the anchor exponent]\label{prop:geometry-sharp}
For $0<\kappa<a<1$ with $\kappa\leq1/10$,
\[
 \frac{\kappa^{12}}{500\,a^2}\ \leq\ c^\star(a,\kappa)\ \leq\
 \frac{1.04\,\kappa^{12}}{a^2} ,
\]
so $c^\star=\Theta(\kappa^{12}/a^2)$. Restricting the
laws to $\mathcal P_L(a,R,\kappa)$ keeps the order: the best constant is
then at most $18\kappa^{12}/a_R^2$, $a_R=\min\{a,R/2\}$, for
$\kappa\leq1/20$ and $2\kappa<a_R$.
\end{proposition}

On a $4\times4$ grid, we weaken one strong edge while strengthening nearby
weak edges. These changes cancel in every four-cycle moment and every
even-set product through the strong edge. The first changing moment is
on the six-edge boundary of two adjacent faces. Its scale is $\kappa^6$,
so its squared signal, and hence the local information, is of order
$\kappa^{12}$ (Appendix~\ref{app:geometry-sharp}). The strong channel changes
at first order even as $\kappa\to0$. This coordinated change is
harder to detect than the isolated change in Remark~\ref{rem:bond-information}.
\subsection{A Spline Estimator and Its Risk}\label{sec:estimator}
We fit piecewise-linear channels on $K$ equal intervals. The estimator
takes $N$ observation vectors, the grid, an input cap $a_{\mathrm{in}}\in[a,1)$ and an input
anchor $0<\kappa_{\mathrm{in}}\leq\kappa$. For each edge $i$, choose the
face $C$ of Lemma~\ref{lem:identify} and put
$g_i(u)=\E[\phi_{C\setminus\{i\}}\mid U_i=u]$.
We estimate the regressions and scales, then project to ensure a valid density:
\begin{enumerate}
\setlength{\itemsep}{0pt}
\item Estimate each witness moment $\theta_F$, clip it to
$[\kappa_{\mathrm{in}}^{|F|},1]$, and form the clipped calibration
scales $\widehat\gamma_A$.
\item On the same vectors, estimate each regression $g_i$ by a centered
linear-spline projection using the known uniform marginal.
\item Project $\widehat g_i/\widehat\gamma_A$ onto centered vectors of knot values
bounded by $a_{\mathrm{in}}$. For a corner, divide the outer product of its two
bounded regressions by $\widehat\gamma_A\widehat\theta_C$ and project
onto separately centered tensors bounded by $a_{\mathrm{in}}^2$.
\end{enumerate}
Exact projections cost polynomial arithmetic time (Lemma~\ref{lem:spline-time});
our corner implementation is iterative (Section~\ref{sec:implementation}).
Fits are positive, normalized, retain uniform forest marginals and use
the queries of~\eqref{eq:evidence-zero}. They may lie outside the target
class: scalar channels need
be neither anchored nor $R$-Lipschitz, and corner projection need not
preserve rank one. Lemma~\ref{lem:coupling} covers this larger family.

\begin{theorem}[Global risk]\label{thm:learning}
Let $p_h\in\mathcal P_L(a,R,\kappa)$, $L\geq4$, $N\geq4$ and $K\geq1$.
With inputs $a_{\mathrm{in}}\in[a,1)$ and $0<\kappa_{\mathrm{in}}\leq\kappa$,
the spline estimator satisfies
\[
 \max\bigl\{\E\KL(p_h\Vert\widehat p),\ \E\KL(\widehat p\Vert p_h)\bigr\}
 \ \leq\ C\,d\left(\frac KN+\frac1{K^2}+\frac1N\right).
\]
Here $C$ depends only on $R,a_{\mathrm{in}},\kappa_{\mathrm{in}}$.
The estimator stores $O(dK+K^2)$ coefficients and has risk
$O(dN^{-2/3})$ when $K\asymp N^{1/3}$.
\end{theorem}

The terms $K/N$, $K^{-2}$ and $N^{-1}$ are regression variance, squared
bias and calibration error. Increasing $K$ reduces approximation bias
but raises variance; $K\asymp N^{1/3}$ balances these terms.
The constant depends on the input anchor: at common reliability $\eta$,
a six-edge sign product has mean $\eta^6$ and variance at most one,
giving the squared signal-to-noise scale $N\eta^{12}$ (Lemma~\ref{lem:witness-identities}).
If each channel is a linear spline on the knot grid, risk is $O(d(K+1)/N)$ with no bias;
this includes the $\tau=1/4$ ramp when $8$ divides $K$
(Appendix~\ref{app:learning-proofs}, Exact splines).

\begin{corollary}[Conditional prediction]\label{cor:conditional-risk}
Under Theorem~\ref{thm:learning}, every fixed hidden set $S\subseteq E$
has expected joint conditional KL bounded by the same risk:
\[
 \E_{\mathrm{train}}\mathcal R_S(p_h,\widehat p)
 \leq\E_{\mathrm{train}}\KL(p_h\Vert\widehat p).
\]
\end{corollary}
\begin{proof}
Put $O=S^c$. The chain rule gives
$\mathcal R_S(p_h,\widehat p)=\KL(p_h\Vert\widehat p)-\KL(p_{h,O}\Vert\widehat p_O)$,
and the second term is nonnegative.
\end{proof}
Thus a single joint fit controls prediction for any fixed missing set,
without retraining. The guarantee averages over observed evidence;
it gives no uniform bound for rare evidence values.
The chain-rule inequality holds for every estimator. With
Theorem~\ref{thm:anchor-free}, it gives risk $CdN^{-2/3}$ without an
anchor, but no polynomial query guarantee.

With a fixed input anchor, spline prediction risk is
$O(dN^{-2/3})=o(d/\log d)$ whenever
$N/(\log d)^{3/2}\to\infty$, with polynomial estimation and two
planar queries per conditional density.
Theorem~\ref{thm:block-prediction} gives the circuit comparison for its
three masks and leaf scopes $\ell\leq(\log d)^2$; unrestricted
predictors and separate coordinate heads are not covered.

\subsection{Minimax Rates and the Cost of Identification}\label{sec:minimax}
The spline estimator divides by cycle moments, which become small at
weak reliability. We next ask whether estimating the density itself
requires this calibration.
For forward and reverse KL, take $D(p,q)=\KL(p\Vert q)$ and
$D(p,q)=\KL(q\Vert p)$, respectively.
Write $\mathcal R_N(D,\mathcal P)=\inf_{\widetilde p}\sup_{p\in\mathcal P}
\E_pD(p,\widetilde p)$, with the infimum over probability-law estimators,
and abbreviate $\mathcal R_N(D)=\mathcal R_N(D,\mathcal P_L(a,R,\kappa))$.
Let $N_{\mathrm{TV}}(\delta)=\min\{N:\mathcal R_N(\TV)\leq\delta\}$.
For the lower bounds, we perturb one channel on each of $\Theta(d)$
vertex-disjoint squares and apply Fano's inequality
(Theorem~\ref{thm:minimax}, Appendix~\ref{app:learning-proofs}).
The condition $\kappa<\eta_*(a,R)$ leaves room for these perturbations.

\begin{theorem}[Hellinger rate without an anchor]\label{thm:anchor-free}
Fix $0<a<1$, $R>0$ and $0<\kappa\leq\eta_*(a,R)/2$.
For either $\mathcal P=\mathcal P_L(a,R,\kappa)$ or
$\mathcal P=\mathcal P^0_L(a,R)$, and all $L\geq4$, $N\geq1$,
\[
 c\min\{1,dN^{-2/3}\}
 \leq\mathcal R_N(H^2,\mathcal P)
 \leq C\min\{1,dN^{-2/3}\},
\]
with $c,C>0$ depending only on $a,R$. Forward KL risk is at most
$CdN^{-2/3}$. For $0<\delta<1/8$, the minimax sample size for expected
TV error $\delta$ lies between $cd^{3/2}$ and $C\delta^{-3}d^{3/2}$.
All upper bounds hold with the same $C$ on every finite simple graph
with $d$ edges.
\end{theorem}

The upper bound bypasses channel recovery by averaging predictive densities over
$\exp\{O(dN^{1/3})\}$ laws \citep[Theorem~2]{yangbarron1999}, using
shared-spin coupling (Appendix~\ref{app:anchor-free}). It gives
neither polynomial construction nor reverse-KL control.

This density guarantee does not imply accurate channel recovery. The next
result applies the coordinated perturbation of
Proposition~\ref{prop:geometry-sharp} to a recovery lower bound on one fixed grid.

\begin{corollary}[Observable-channel recovery]\label{cor:recovery-lower}
Put $a_R=\min\{a,R/2\}$. For $0<a<1$, $R>0$,
$0<\kappa\leq\min\{1/20,a_R/4\}$ and $N\geq1$,
\[
 \inf_{\widehat f}\sup_{p_h\in\mathcal P_4(a,R,\kappa)}
 \E_h\Delta^2(\widehat f,\mathcal O(h))
 \geq\frac{a_R^2}{65536}\min\{1,(N\kappa^{12})^{-1}\},
\]
where $\widehat f$ ranges over estimators of the observable block
channels from $N$ independent vectors.
\end{corollary}

Even on a fixed grid, vanishing worst-case block-recovery error requires
$N\kappa^{12}\to\infty$; the density rate is uniform in $\kappa$.
No matching recovery upper bound is proved (Appendix~\ref{app:geometry-sharp}).
In finite mixtures, converting density error into parameter error
likewise depends on identifiability \citep{honguyen2016,heinrichkahn2018}.
\section{Empirical Consequences and Limits}\label{sec:experiments}
We compare block error with joint divergence, measure query cost, and
examine when fitting dependence improves log loss. We then vary spline
resolution to study estimation error and use known-channel tiling to
isolate approximation error.
We use ramp and smooth loop channels with matched reliability
$\eta=0.875a$, and a Gaussian copula with correlation $0.5^{|i-j|}$.
Loop targets use $a=0.95$ unless stated otherwise; KL is in nats per vector.
Appendices~\ref{app:submission-protocol} and~\ref{app:protocols} give protocols and uncertainty.
The fitted dimensions are too small to test the asymptotic circuit bounds. For the ramp,
the circuit bound first exceeds three nodes at $d=38{,}364$
(Remark~\ref{rem:finite-constants}). Fitted caps exceed the sufficient
$a_{\mathrm{in}}<1/3$ condition in Proposition~\ref{prop:conditioning},
so we check queries against positive row transfer, which sums
nonnegative terms row by row.

\subsection{Observable Geometry and Query Cost in Practice}\label{sec:geometry-check}
\begin{figure}[htbp]\centering
\includegraphics[width=\linewidth]{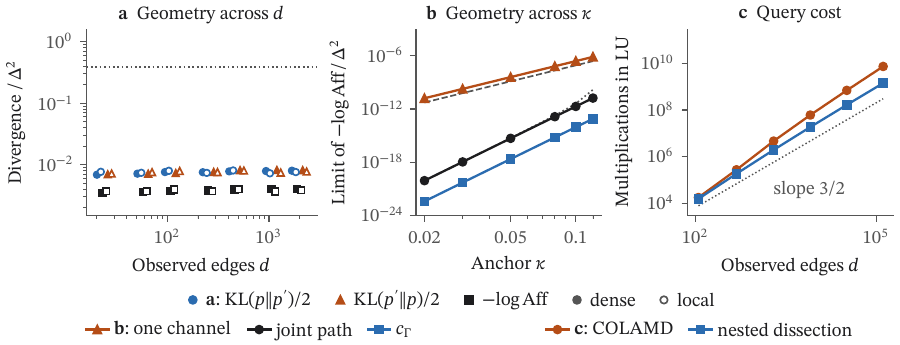}
\caption{Observable geometry and query cost. (a) Divergences
$\tfrac12\KL(p\Vert p')$, $\tfrac12\KL(p'\Vert p)$ and $-\log\Aff$
divided by $\Delta^2$, averaged over perturbation amplitudes $0.25,0.5$.
The upper constant of
Corollary~\ref{cor:grid-geometry} is dotted; the lower constant $c_\Gamma$
lies below the plotted range. Filled markers are dense and open
markers local perturbations; Monte Carlo bars are smaller than the
markers. (b) Local limits of $-\log\Aff/\Delta^2$ for one moving
channel (triangles, order $\kappa^6$) and the joint path (circles, order
$\kappa^{12}$), evaluated over all
sign patterns at $a=\DiagnosticValue{sharp-amplitude}$.
Squares give $c_\Gamma$; dashed and dotted lines give the respective
one-channel lower and joint-path upper bounds, the latter valid for
$511\kappa^3<1$. (c) Multiplications in the LU factorization of one Kac--Ward query under
nested-dissection and COLAMD orderings; the dotted segment has slope
$3/2$.}
\label{fig:geometry-queries}
\end{figure}

To check how block error relates to joint divergence
(Corollary~\ref{cor:grid-geometry}), we perturb one central scalar channel
(local) or all scalar channels (dense), preserving reliability, up to
$L=\DiagnosticValue{geometry-max-side}$. Estimates satisfy
\eqref{eq:global-divergence}: $\KL/\Delta^2$
lies in $[\DiagnosticValue{geometry-kl-min},\DiagnosticValue{geometry-kl-max}]$
and $-\log\Aff/\Delta^2$ in
$[\DiagnosticValue{geometry-aff-min},\DiagnosticValue{geometry-aff-max}]$,
with no visible trend in $d$. Figure~\ref{fig:geometry-queries}a plots
half of each KL ratio to match the scale in~\eqref{eq:global-divergence}.
Exact sign enumeration on the sharpness instance gives anchor
exponents $\DiagnosticValue{sharp-single-slope}$ for one moving
channel and $\DiagnosticValue{sharp-slope}$ for the joint path
(Figure~\ref{fig:geometry-queries}b), consistent with
Proposition~\ref{prop:geometry-sharp} and Remark~\ref{rem:bond-information}.
A corner rescaling has zero
observable distance and log-density discrepancy at most
$\DiagnosticValue{geometry-corner-err}$, despite positive per-edge error
(Section~\ref{app:rate-diagnostics}). This illustrates why we compare
observable blocks rather than individual channels.

We measure query arithmetic up to $L=\DiagnosticValue{query-max-side}$
($d=\DiagnosticValue{query-max-d}$). Over the three largest grids,
LU multiplication counts scale as $d^{\DiagnosticValue{query-nd-mult-slope}}$
under nested dissection and $d^{\DiagnosticValue{query-colamd-mult-slope}}$
under column approximate minimum degree (COLAMD) ordering
(Figure~\ref{fig:geometry-queries}c).
The former is consistent with Proposition~\ref{prop:queries}'s exponent,
but measures floating-point LU rather than exact Gram $LDL^*$.
It does not test accuracy on the largest grids.

\subsection{Weak Signals and Likelihood Refinement}\label{sec:shrinkage}
\noindent\emph{Weak dependence.}
We compare fitting dependence with using independence when calibration
moments are weak. Six-edge calibration has squared signal-to-noise scale $N\eta^{12}$
(Section~\ref{sec:estimator}); ignoring dependence costs at most
$\log L_G(\xi)$ by~\eqref{eq:loop-energy}. We select shrinkage
$h\mapsto\lambda h$ by validation, including independence at $\lambda=0$.
For ramp channels at $d=112$,
moment fits without shrinkage have higher mean KL than independence at every tested point
with $N\eta^{12}\leq2$. At $\eta=0.5$, $N=8192$, shrinkage reduces
mean KL from the paired independence value
$\SubmissionValue{weak-ramp-e0.5-n8192-paired-independence-kl}$ to
$\SubmissionValue{weak-ramp-e0.5-n8192-shrunk-kl}$ nats per vector.
Shrinkage also gives lower mean KL than independence at $\eta=0.375$, $N=32768$, where
$N\eta^{12}\approx0.253$. These settings do not locate a sharp
threshold (Figure~\ref{fig:calibration}). With heterogeneous
linear--sine channels and weak anchors, all
\SubmissionValue{nonramp-fits} fits have higher KL than independence under every tested
calibration, including true scales (Section~\ref{app:nonramp}). These results
support including $\lambda=0$ in validation; no oracle bound is proved.
The calibration scales have mean-squared error $O(1/(N\kappa_{\mathrm{in}}^{12}))$;
subsequent divisions add anchor factors (Appendix~\ref{app:learning-proofs}).
Theorem~\ref{thm:learning} fixes the input anchor.

\phantomsection\noindent\emph{Likelihood refinement and the forest floor.}\label{sec:matched}
At $d=24,112$, we compare the moment estimator with direct likelihood
optimization and expectation--maximization (EM) to assess whether
refinement improves the fit. Baselines are a hidden Chow--Liu tree (HCLT),
Gaussian mixture (GMM), monotone probabilistic circuit (MPC),
orthogonal sum-of-squares circuit (OSOS), and independence.
We use ten seeds (Table~\ref{tab:matched-joint}).
Under the budgets and starts in Section~\ref{app:submission-protocol},
the paired loop-task comparisons favor the moment estimator over every
generic baseline and independence. Likelihood and EM improve on moments
(Bonferroni-adjusted intervals, Table~\ref{tab:submission-contrasts}).
At $d=112$ none of the four generic
baselines improves mean KL over independence on either loop target.
The selected one-component Gaussian mixture has independence as its
population optimum (Proposition~\ref{prop:forest-blindness}). HCLT chooses
its structure from pairwise statistics, which miss loop dependence
(Section~\ref{sec:loop-expansion}).
All selected $d=112$ loop-task OSOS fits reach the time cap (Table~\ref{tab:matched-joint}).
We did not test neural autoregressive models or flows, so these results
do not establish an advantage over general density estimators. On Gaussian targets,
uniform forest marginals force a positive KL gap for every loop law
(Section~\ref{app:matched-protocol});
the fitted laws stay near independence loss. The tree, mixture and monotone
circuit have KL $\SubmissionValue{t2-gmm-kl}$--$\SubmissionValue{t5-mpc-kl}$
nats.

\noindent\emph{Predicting missing readings.}
To check whether the learned joint densities also predict missing readings,
we score the same fits without retraining under four masks: nested 25\% and 50\%
masks from one random coordinate permutation, a corner patch, and an
observed forest of all horizontal edges (Table~\ref{tab:conditional}).
The chain rule bounds each fit's conditional KL by its joint KL
(Corollary~\ref{cor:conditional-risk}). These masks differ from those
in Theorem~\ref{thm:block-prediction}, so this comparison does not test
that lower bound; seed intervals are unadjusted.
For loop targets and fitted loop laws, the horizontal-edge evidence is
uniform under both laws. The KL chain rule then equates conditional and
joint KL (Proposition~\ref{prop:forest-blindness}); this need not hold for
the Gaussian target or the generic baselines.

\subsection{Resolution and Average versus Global Error}\label{sec:fixed-resolution}
To examine the variance--bias tradeoff in Theorem~\ref{thm:learning},
we compare $K=8$ with $K=8\lceil N^{1/3}/8\rceil$ for $N$
from 8192 to 524288 (Section~\ref{app:rate-diagnostics}).
The ramp has zero spline bias on every tested grid; $K=8$ gives lower
mean KL throughout. Growing $K$ gives lower mean KL on the smooth
target from $N=131072$. With growing $K$, joint KL on both targets
decays more slowly than $N^{-2/3}$ over the tested range
(Figure~\ref{fig:submission-rates}). Joint KL stays below
Lemma~\ref{lem:coupling}'s block-error bound; these curves do not identify
the asymptotic rate.

To compare average KL and global TV (Proposition~\ref{prop:blocks}),
we use known ramp channels and $q_b$ with $b=\lfloor\log_2d\rfloor$,
from $d=60$ to $d=1{,}984$ (Figure~\ref{fig:tiling-gap}, Table~\ref{tab:tiling-gap}).
Over \TilingDraws{} draws, KL$/d$ ranges from $\TilingKLMin$ to
$\TilingKLMax$ nats. The certified face-parity lower bound on TV rises from
$\TilingTVBoundMin$ to $\TilingTVBoundMax$, while estimated TV rises
from $\TilingSixTV$ to near one.

For a centered half-side patch with known channels and paired draws,
conditional KL stays between the hidden-parity and deleted-edge bounds
at every tested tile width (Table~\ref{tab:masked-tiling}). These bounds
concern the chosen tiled law, so this check does not test the general
circuit lower bound.

For staggered and aligned-matching masks, \eqref{eq:forest-loss} equates
joint and conditional tiling log ratios. The same estimates and standard
errors give $\mathcal R_S(p_h,q_b)$; multiply KL$/d$ by $d/|S|$ for loss
per hidden edge.
\section{Discussion}\label{sec:discussion}
Polynomial-size circuits can have vanishing per-coordinate KL while TV
tends to one (Proposition~\ref{prop:blocks}). Thus small average log loss
does not certify accurate joint probabilities.
Larger leaves reduce KL cost; node counts exclude leaf computation
(Theorem~\ref{thm:kl-size}).

The model assumes known graphs and margins and cannot represent dependent
forest marginals (Proposition~\ref{prop:forest-blindness}). Estimated
margins \citep{liu2009nonparanormal} would require additional density
and tail bounds to control KL before the marginal transform.
The query bound counts arithmetic operations (Proposition~\ref{prop:queries}).
The representation bounds are uninformative at our fitted dimensions
(Section~\ref{sec:experiments}).

A convex mixture with a tractable tree copula on the same margins would
allow pairwise dependence. Linearity of integration makes its evidence
likelihood the same mixture of component likelihoods. Learning and
representation costs for this larger class remain open.

Can an efficient estimator reach the anchor-free rate without recovering
weak channels? Theorem~\ref{thm:anchor-free} gives no polynomial algorithm.
Theorem~\ref{thm:block-prediction} covers three masks with leaf scopes
$\ell\leq(\log d)^2$. Extending these bounds to other masks and larger
leaves remains open.
The finite-energy barrier also permits
dependent latent spins (Theorem~\ref{thm:dependent-barrier}); our learning
theory does not cover those priors.

Code and implementation:
\url{https://github.com/anindex/noisy_loop}.
\FloatBarrier
\label{page:main-end}
\noindent\begin{minipage}{\linewidth}
\acks{The author is supported by the VinUniversity Center for AI Research.
The author declares no competing interests.}
\end{minipage}
\par
\appendix
\section{Queries, Conditioning and the Proof of Theorem~\ref{thm:separation}}\label{app:model-details}
We prove the forest-marginal identities and query bounds, then combine
the learning and representation results to prove the main separation.
\subsection{Forest Marginals and Likelihood}
\phantomsection\label{sec:forest-blindness}\begin{proposition}[Forest marginals and a likelihood floor]\label{prop:forest-blindness}
Let $p=p_h$ be~\eqref{eq:density} on a simple graph. For every forest
$F\subseteq E$, $p_F=\mu_F$. If a positive probability density $q$
relative to $\mu$ satisfies $\log q=\sum_{j=1}^m\psi_j(u_{F_j})$, where
every $F_j$ is a forest and $\psi_j\in L^1(\mu_{F_j})$, then
\begin{equation}\label{eq:forest-kl-floor}
 \KL(p\Vert q)=\KL(p\Vert\mu)+\KL(\mu\Vert q).
\end{equation}
Consequently $\mu$ is the unique best approximation in any such class
that contains it.
\end{proposition}
The examples in Section~\ref{sec:loop-expansion} satisfy this lower bound;
hidden trees and mixtures have log-sum densities and need not.
Conversely, if $q$ is a law~\eqref{eq:density} or a fitted block law,
its forest marginals are uniform. For any target density $p$, data processing gives
$\KL(p\Vert q)\geq\max_{F\text{ forest}}\KL(p_F\Vert\mu_F)$ for every
such $q$. A target with a dependent pair therefore has a positive KL gap.
\begin{proof}
Since $p_F=\mu_F$, $\E_p\psi_j=\E_\mu\psi_j$ for every summand and
$\E_p\log q=\E_\mu\log q$. Subtracting this from $\E_p\log p$
gives~\eqref{eq:forest-kl-floor}; all terms are finite because $\log p$
is bounded on a finite graph and each $\psi_j$ is integrable.
Nonnegativity of KL and its equality case give uniqueness.
\end{proof}

\subsection{Exact Factorized Queries}
\begin{samepage}
\begin{proposition}[Exact factorized queries]\label{prop:queries}
Let $G$ be planar with a straight-line embedding and bounded degree, and
let $\mathcal E$ specify missing, point or interval evidence.
Given its local integrals $(\bar\ell_e,t_e)$, the positive likelihood
in~\eqref{eq:evidence-zero} can be computed in $O(d^{3/2})$ exact
arithmetic operations and one positive fourth root.
The resulting Gram $LDL^*$ factors give all unobserved bond means
$\bar B_e=\E[B_e\mid\mathcal E]$ and conditional densities
$1+\bar B_eh_e$ in $O(d^2\log d)$ further operations.
Constants depend only on the maximum degree.
\end{proposition}
\end{samepage}

The Kac--Ward identity is polynomial in $w$ and needs no convergent
series. For $|w_e|<1$, the spin sum gives $L_G(w)>0$, including for
negative weights.

\noindent\emph{Marginal likelihood.}
For a nonnegative coordinate likelihood $\ell_e$, let
$\bar\ell_e=\int\ell_e(u)\,du$ and assume $0<\bar\ell_e<\infty$.
The channel average $t_e=\bar\ell_e^{-1}\int\ell_e(u)h_e(u)\,du$
satisfies $|t_e|\leq a_e<1$. Conditional independence and finite spin
summation yield
\[
 \int p_h(u)\prod_e\ell_e(u_e)\,du
 =\E\prod_e\bar\ell_e(1+t_eB_e)
 =\Bigl(\prod_e\bar\ell_e\Bigr)L_G(t).
\]
Point observations use the corresponding emission density, and no field
term on an individual spin appears. Since
$L_G(t)=2^{-|V|}\sum_s\prod_e(1+t_eB_e)$ with every factor positive,
we have $L_G(t)>0$.

\noindent\emph{Factorization.}
By~\eqref{eq:kw}, $\det(I-T\diag(t))=L_G(t)^2>0$, so the
matrix is invertible and its Gram matrix is Hermitian positive definite
with determinant $L_G(t)^4$. In exact arithmetic, its $LDL^*$
factorization needs no pivoting in any order. Let $A=I-T\diag(t)$. The sparsity graphs of $A$ and $A^*A$
are contained in the graph $\mathcal S$ on directed edges that joins
orientations of edges sharing a vertex of $G$. For $A$ this follows
from nonbacktracking adjacency. For $A^*A$, a nonzero entry comes
from two columns sharing a nonzero row: their edges share either
the starting or ending vertex of that row's edge.

Let every vertex of $G$ have degree at most $\deg_{\max}$. We construct
a separator for any set $X$ of directed edges. Let $G_X$ be the subgraph
of $G$ formed by their undirected edges. Give each vertex of $G_X$ the weight $1/|X|$ times
the number of edges of $X$ that begin there. The graph $G_X$ is planar
with at most $2|X|$ vertices, so the planar separator theorem
\citep{liptontarjan1979} gives a vertex set $Z$ with
$|Z|\leq2\sqrt2\sqrt{2|X|}=4\sqrt{|X|}$ whose removal leaves two vertex
sets with no edge of $G_X$ between them, each of weight at most $2/3$.
At most $2\deg_{\max}|Z|$ edges of $X$ touch $Z$. Every other edge of $X$
has both endpoints in one of the two sets, and edges in different sets
share no vertex. Removing the $O(\sqrt{|X|})$ edges that touch $Z$
therefore splits the subgraph of $\mathcal S$ on $X$, and hence every
subgraph of $\mathcal S$ with vertex set $X$, into two parts with at most
$2|X|/3$ vertices each. The graphs of the matrix and of its Gram matrix
are subgraphs of $\mathcal S$, whose maximum degree is at most
$4\deg_{\max}-3$. These hereditary square-root separators and bounded
degree allow generalized nested dissection
\citep{lipton1979,gilberttarjan1987} to order the $2d$ unknowns such that $LDL^*$
factorization of the Gram matrix costs $O(d^{3/2})$ arithmetic
operations with $O(d\log d)$ fill. The determinant is the product of the
pivots, and its positive fourth root is $L_G(t)$.

Turns must lie in $(-\pi,\pi)$; a $2\pi$ shift changes an entry sign.
Unpivoted LU also works under strict row diagonal dominance
(Proposition~\ref{prop:conditioning}).

\noindent\emph{Bond means and sampling.}
For an unobserved edge $e$, we have $t_e=0$. Differentiating the
spin sum therefore removes its factor $1+t_eB_e$ and inserts $B_e$.
The bond mean is
$\bar B_e=\E[B_e\mid\text{evidence}]=\partial_{t_e}\log L_G(t)
=-\tfrac12\sum_{\vec e}[(I-T\diag(t))^{-1}T]_{\vec e\vec e}$,
summed over both orientations of $e$ \citep[compare][Proposition 2]{johnson} with
$T$ the turn matrix of~\eqref{eq:kw}. Each column of
$(I-T\diag(t))^{-1}T$ follows from the $LDL^*$ factors by
one forward and one backward triangular solve, since
$(I-T\diag(t))^{-1}$ is the inverse Gram matrix times the
adjoint. The $O(d\log d)$ fill gives an $O(d^2\log d)$ upper bound
for all $2d$ columns; this cost need not be optimal. The bond mean gives $U_e$ the conditional density
$1+\bar B_eh_e(u)$ and distribution function $u+\bar B_e\int_0^uh_e$, with
density floor $1-a_e$. To sample unconditionally, we draw
the spins, then each reading from $1+B_eh_e$. For linear-spline channels
such as the ramp, we invert the piecewise-quadratic distribution function.
Otherwise we use rejection sampling from the uniform law with envelope
$1+a_e$, requiring $O(d)$ expected channel evaluations.

\noindent\emph{Fitted corner blocks and implementation.}
After contracting a degree-two corner, a separately centered bounded
block channel $h_A(u_1,u_2)$ also contributes $\bar\ell(1+tB)$ after
integrating factorized evidence on its two coordinates, so the fitted
corner blocks of Theorem~\ref{thm:learning} admit the same query algorithm.
With input cap $a_{\mathrm{in}}$, we realize the block on the original grid by
edge weights $a_{\mathrm{in}}$ and $t/a_{\mathrm{in}}$, both bounded by
$a_{\mathrm{in}}$ since $|t|\leq a_{\mathrm{in}}^2$; summing the
intervening spin recovers $1+tB$. Linear and bilinear spline integrals are finite sums of polynomial
integrals. The implementation instead uses weights
$\sqrt{|t|}$ and $\sign(t)\sqrt{|t|}$, also bounded by $a_{\mathrm{in}}$,
and sparse LU in COLAMD order;
its safeguards and limitations are in Section~\ref{sec:implementation}.

\subsection{Conditioning of the Kac--Ward Matrix}
An exact-arithmetic cost bound does not control numerical stability.
We give a sufficient small-channel condition that bounds the matrix
condition number.
For an invertible matrix $A$, write $\|A\|_\infty=\max_i\sum_j|A_{ij}|$ and
$\operatorname{cond}_\infty(A)=\|A\|_\infty\|A^{-1}\|_\infty$.
\begin{proposition}[Neumann bound]\label{prop:conditioning}\label{cor:grid-conditioning}
Put $\nu=\|T\diag(w)\|_\infty$. If $\nu<1$, then
\[
 \|(I-T\diag(w))^{-1}\|_\infty\leq(1-\nu)^{-1},\qquad
 \operatorname{cond}_\infty(I-T\diag(w))\leq\frac{1+\nu}{1-\nu}.
\]
For maximum degree $\deg_{\max}$,
$\nu\leq(\deg_{\max}-1)\max_e|w_e|$.
Thus on the square grid, every factorized evidence query with channel
cap $a<1/3$ has condition number at most $(1+3a)/(1-3a)$.
The same bound holds for fitted laws with input cap $a_{\mathrm{in}}<1/3$,
with $a_{\mathrm{in}}$ in place of $a$.
\end{proposition}
\begin{proof}
The Neumann series bounds the inverse norm by $\sum_{k\geq0}\nu^k$,
and $\|I-T\diag(w)\|_\infty\leq1+\nu$ gives the condition number.
Each directed edge has at most $\deg_{\max}-1$ nonbacktracking
successors, with unit-modulus turn factors. Evidence weights are
averages of channels, hence bounded by $a$; the two-edge realization
above bounds fitted corner weights by $a_{\mathrm{in}}$.
\end{proof}
The condition number does not bound the relative determinant error
uniformly in $d$; Section~\ref{sec:implementation} gives numerical limits.

\subsection{Largest Reliability}\label{app:reliability-ceiling}
Centering and Lipschitz continuity give
$h(\tfrac12+t)-h(\tfrac12-t)\leq\min\{2Rt,2a\}$.
Integrating over $0\leq t\leq1/2$ bounds reliability by
$\eta_*(a,R)=a-a^2/R$ if $R\geq2a$, and by $R/4$ otherwise.
The channel $\clip(R(u-\tfrac12),-a,a)$ gives equality.

\subsection{Proof of the Main Separation Theorem}
\begin{proof}(Theorem~\ref{thm:separation}).
For item 1, we take $K=\lceil N^{1/3}\rceil$ in
Theorem~\ref{thm:learning} and apply Theorem~\ref{thm:minimax}.
For $N\leq3$, use $H^2\leq1$ and $d\geq24$.
Proposition~\ref{prop:queries} gives item 2 because the grid is planar
with degree at most four. Corollary~\ref{cor:grid-circuit} gives
item 3, with smoothing from Appendix~\ref{app:smoothing} and the
row-frontier upper bound.
\end{proof}
\section{Learning, Geometry and Minimax Proofs}\label{app:learning-proofs}
We first identify the observable parameters and prove the distance bounds.
We then analyze the spline estimator and derive the minimax rates.
\subsection{Observable Parameters on a General Graph}\label{app:graph-identification}
We characterize the channel changes that leave the observed law unchanged.
\begin{theorem}[Observable parameters of a graph]\label{thm:graph-quotient}
Let $h,h'$ be centered channels on a finite simple graph, bounded by
$a<1$, with $\int\phi h_e>0$ and $\int\phi h'_e>0$ on every edge. They
induce the same observed law if and only if, on every non-bridge edge,
$h'_e=e^{b_e}h_e$ for real constants $b_e$ with
\[
 \sum_{e\in\mathcal Q}b_e=0\qquad\text{for every series class }\mathcal Q;
\]
bridge channels may change arbitrarily within the assumptions.
Equivalently, the observable parameters are the normalized shapes
$h_e/\eta_e$ on non-bridge edges and the products
$\prod_{e\in\mathcal Q}\eta_e$ over series classes.
\end{theorem}

For a simple cycle $C\ni e$, observable moments give
\begin{equation}\label{eq:cycle-shape}
 \theta_C=\E\prod_{f\in C}\phi(U_f)=\prod_{f\in C}\eta_f,
 \qquad
 \frac{\E[\prod_{f\in C\setminus\{e\}}\phi(U_f)\mid U_e=u]}{\theta_C}
 =\frac{h_e(u)}{\eta_e},
\end{equation}
so shapes are identified directly, while amplitudes obey linear equations
in their logarithms with the \emph{unsigned real} cycle-incidence matrix.

\begin{proof}
The bond product of a cycle is one; removing $e$ leaves a path with bond
product $B_e$, and $\E[B_e\mid U_e=u]=h_e(u)$. With
$\E[\phi(U_e)\mid B_e]=\eta_eB_e$ this proves~\eqref{eq:cycle-shape}.
Equal laws therefore have equal normalized shapes and equal cycle
amplitude products. On a non-bridge edge the shapes agree and
$\eta_e$ and $\eta'_e=\int\phi h'_e$ are positive, so $h'_e=e^{b_e}h_e$
with $b_e=\log(\eta'_e/\eta_e)$. The remaining conditions are
$\sum_{e\in C}b_e=0$ for every simple cycle $C$.

We prove the real-valued version of the cycle-span result
stated over $\mathbb Q$ by
\citet[Proposition~1.1, attributed there to Goddyn]{averkov2021}; the
real span follows since cycle incidence vectors are rational.
Choose an even edge set $F$ uniformly from the binary cycle space. For
each non-bridge edge, $X_e=1_{\{e\in F\}}$ is a nonzero binary linear
functional of $F$ and is unbiased. Two such functionals are identical
exactly when their edges are in series, since an even set decomposes into
edge-disjoint simple cycles; otherwise they have rank two and are
pairwise independent. If $b$ sums to zero on every simple cycle, it sums
to zero on $F$. With $b_{\mathcal Q}=\sum_{e\in\mathcal Q}b_e$,
\[
 0=\Var\Bigl(\sum_eb_eX_e\Bigr)=\tfrac14\sum_{\mathcal Q}b_{\mathcal Q}^2,
\]
so every $b_{\mathcal Q}=0$. Conversely, each simple cycle contains all or none of
each series class, so the class equations imply the cycle equations and
leave every term of $L_G(h)$ unchanged; bridges appear in no term.
The same kernel characterization proves witness existence: $\mathbf1_{C_A\setminus A}$
is constant on each series class and zero on bridges, so it is
orthogonal to the kernel just identified and lies in the real span
of cycle-incidence vectors.

\end{proof}
\subsection{Proof of Lemmas~\ref{lem:witness-identities} and~\ref{lem:identify}}
\begin{lemma}[Witnesses on the grid]\label{lem:identify}
Let $p_h$ be anchored on the $L\times L$ grid, $L\geq4$. For a block
$A$ choose a square face $C\supseteq A$, so that
$\gamma_A=\prod_{f\in C\setminus A}\eta_f\geq\kappa^{4-|A|}$. For an
interior edge $A=\{e\}$, its other incident face $C'$ gives
$\gamma_A=\sqrt{\theta_C\theta_{C\triangle C'}/\theta_{C'}}$. For a
boundary scalar edge or corner pair, every $f\in C\setminus A$ has
another incident face $C_f$, and
\begin{equation}\label{eq:short-calibration}
 \eta_f=\sqrt{\frac{\theta_C\theta_{C_f}}{\theta_{C\triangle C_f}}}.
\end{equation}
These are witnesses whose weights are multiples of $\tfrac12$ on faces
and on six-cycles around two adjacent faces, and
\[
 \Gamma_A\leq\Gamma:=\kappa^{-3}+3a\kappa^{-4}+\tfrac32a\kappa^{-6}.
\]
\end{lemma}

For estimation, we also use one-dimensional regressions. Under the
lemma's assumptions, for each $i\in A$,
$g_i(u)=\E[\phi_{C\setminus\{i\}}\mid U_i=u]
=(\prod_{f\in C\setminus\{i\}}\eta_f)h_i(u)$ recovers $h_A=g_i/\gamma_A$
when $A=\{i\}$, and a corner pair $A=\{1,2\}$ satisfies
\begin{equation}\label{eq:corner-recovery}
 h_A(u_1,u_2)=\frac{g_1(u_1)g_2(u_2)}{\gamma_A\theta_C}.
\end{equation}

\begin{proof}
Given a bond, $\E[\phi(U_e)\mid B_e]=\eta_eB_e$, and readings are
independent given the spins. Every even edge set has bond product one,
which gives $\theta_F=\prod_{f\in F}\eta_f$. Let $A$ have a witness.
Since $C_A$ is a cycle, the bonds on $C_A\setminus A$ multiply to $B_A$,
so the conditional mean of $\phi_{C_A\setminus A}$ given the spins and
$U_A$ is $\gamma_AB_A$. The block $A$ is a forest, so its bonds are
independent fair signs and each $U_e$, $e\in A$, depends on the spins
only through $B_e$; hence $\E[B_A\mid U_A=u_A]=h_A(u_A)$, which is the
first identity of~\eqref{eq:block-regression}. Even sets and $C_A$
contain no bridge, so pairing~\eqref{eq:witness} with the vector
$(\log\eta_f)_f$ over the non-bridge edges gives the second:
\[
 \sum_{F\in\mathcal F_A}\vartheta_{A,F}\log\theta_F
 =\sum_{f}\log\eta_f\sum_{F\in\mathcal F_A}\vartheta_{A,F}\mathbf 1_F(f)
 =\sum_{f\in C_A\setminus A}\log\eta_f=\log\gamma_A .
\]
The moments $\theta_F$, $F\in\mathcal F_A$, and the regression of
$\phi_{C_A\setminus A}$ on $U_A$ are functionals of the law of $U_{W_A}$.

On the grid, two faces sharing an interior edge meet only in that edge,
and their symmetric difference is a six-cycle, so
$\mathbf 1_{C\setminus\{e\}}=\tfrac12(\mathbf 1_C+\mathbf 1_{C\triangle C'}
-\mathbf 1_{C'})$ and
$\mathbf 1_{\{f\}}=\tfrac12(\mathbf 1_C+\mathbf 1_{C_f}-\mathbf 1_{C\triangle C_f})$;
summing the latter over $f\in C\setminus A$ gives~\eqref{eq:witness}, and
exponentiating gives the two-face ratio
and~\eqref{eq:short-calibration}, where anchoring selects the positive
root. For a top boundary edge between columns $j,j+1$ (zero-based),
corner pairs occupy $j=0$ and $j=L-2$, so $1\leq j\leq L-3$ and the
other three edges of its face each lie in a second face. A corner pair
removes both exterior edges of its face, leaving two edges that each lie
in a second face. Rotation covers the other sides. The weights
$|\vartheta_{A,F}|$ on four-cycles and on six-cycles sum to at most
$(1,\tfrac12)$ for an interior edge, $(2,1)$ for a corner pair and
$(3,\tfrac32)$ for a boundary scalar edge, with $|C\setminus A|\leq3$
and $a_A\leq a$, which gives $\Gamma_A\leq\Gamma$. The regressions $g_i$ follow from the same conditional-expectation
calculation applied to the single edge $i$; for a
corner, $g_1=\gamma_A\eta_2h_1$, $g_2=\gamma_A\eta_1h_2$ and
$\theta_C=\gamma_A\eta_1\eta_2$, which gives~\eqref{eq:corner-recovery}.
\end{proof}

\subsection{Grid Windows}
\begin{samepage}
\begin{lemma}[Grid windows]\label{lem:witnesses}
On the $L\times L$ grid with $L\geq4$ and the witnesses of
Lemma~\ref{lem:identify}, each window has at most $13$ edges, within
Chebyshev distance two of the first vertex $v_A$ of its block in
row-major order, and $24$ colors make windows of the same color vertex
disjoint. So the grid has a witness system with $24$ colors and
constants $\Gamma_A\leq\Gamma$.
\end{lemma}
\end{samepage}

\begin{proof}
The windows are two adjacent faces for an interior edge (7 edges),
three faces in an L shape at a corner pair (10 edges) and four faces in
a T shape at a boundary scalar edge (13 edges). In each case the vertices
lie within Chebyshev distance two of $v_A$.

Index vertices by (row, column). An interior horizontal edge with left
endpoint $(r,c)$ has its window vertices in rows $r-1,r,r+1$ and columns
$c,c+1$; color it by $(r\bmod3,c\bmod2)$. An interior vertical edge with
upper endpoint $(r,c)$ has them in rows $r,r+1$ and columns
$c-1,c,c+1$; color it by $(r\bmod2,c\bmod3)$. Two distinct edges in one
of these twelve classes differ by at least three in the coordinate taken
modulo three or by at least two in the other, so their windows are
vertex disjoint. The window of the boundary scalar edge at position
$j$ along a side, $1\leq j\leq L-3$, lies in positions $j-1,\dots,j+2$
along that side and in the three rows or columns nearest to it. If
$L\geq6$, color the edge by $j\bmod4$, with one palette for the top and
bottom sides and another for the left and right sides: on one side,
equal colors are at least four positions apart, and opposite sides use
disjoint rows or columns because $L-3\geq3$. If $L\leq5$ there are
$4(L-3)\leq8$ such edges, which receive distinct colors. Each corner
pair receives its own color. This uses at most $12+8+4=24$ colors.
\end{proof}

\subsection{Global Divergence Bounds}\label{app:global-geometry}
\noindent\emph{Local divergence bounds.}
For bounded densities, $(p-q)^2=(\sqrt p+\sqrt q)^2(\sqrt p-\sqrt q)^2$
gives
\begin{equation}\label{eq:local-hellinger}
 \|p-q\|_2^2\leq8\max\{\|p\|_\infty,\|q\|_\infty\}H^2(p,q),
\end{equation}
and both directed KL divergences dominate $-2\log\Aff\geq2H^2$. For real
$x,y$ with $|x|,|y|\leq a<1$,
\begin{equation}\label{eq:emission-kl}
 \tfrac12\sum_{b=\pm1}(1+bx)\log\frac{1+bx}{1+by}
 \leq\frac{(x-y)^2}{2(1-a^2)},\qquad
 1-\tfrac12\sum_{b=\pm1}\sqrt{(1+bx)(1+by)}
 \leq\frac{(x-y)^2}{8(1-a^2)} .
\end{equation}
The first left side is the KL divergence between the laws of a sign
with means $x$ and $y$; as a function of $x$ it vanishes with its
derivative at $x=y$ and has second derivative $1/(1-x^2)\leq1/(1-a^2)$.
For the second, put $s_\pm=\sqrt{1\pm x}$ and $t_\pm=\sqrt{1\pm y}$; the
left side is
$\tfrac14\sum_\pm(s_\pm-t_\pm)^2=\tfrac14(x-y)^2\sum_\pm(s_\pm+t_\pm)^{-2}$,
and $(s_\pm+t_\pm)^2\geq4s_\pm t_\pm$, $s_+t_++s_-t_-\leq2$ and
$s_+t_+s_-t_-=\sqrt{(1-x^2)(1-y^2)}\geq1-a^2$ bound the sum by
$1/\{2(1-a^2)\}$.

\noindent\emph{Upper bounds and Lemma~\ref{lem:coupling}.}
\begin{proof}
Let $Z=(B_A)_A$ be the block bonds of uniform spins.
By~\eqref{eq:block-density}, $p_h$ is the law of readings that, given
$Z$, are independent across blocks with densities $1+B_Ah_A(u_A)$ on
$[0,1]^A$; each is a probability density because $h_A$ is centered in
every argument and $|h_A|\leq a_A<1$. The law of $Z$ does not depend on
the channels, so $p_h$ and $p_{h'}$ are mixtures over the same latent
law, with emissions $1+B_Ah_A$ and $1+B_Ah'_A$, and each $B_A$ is a fair
sign because no block is a cycle. The KL chain rule adds the
conditional KL of the block emissions, and the first bound
of~\eqref{eq:emission-kl} at $x=h_A(u_A)$, $y=h'_A(u_A)$, integrated
over $u_A$, bounds the joint KL of $(Z,U)$ by
$\sum_A\|h_A-h'_A\|_2^2/\{2(1-a_A^2)\}$ in either direction;
marginalizing $Z$ does not increase KL. Likewise the affinity of
the two joint laws is $\E\prod_A\{1-H_A^2(Z)\}$, with $H_A^2(Z)$ the
squared Hellinger distance of the two emissions of block $A$ given
$Z$. We use $\prod_A(1-x_A)\geq1-\sum_Ax_A$ on $[0,1]$ and the second
bound of~\eqref{eq:emission-kl},
$\E H_A^2(Z)\leq\|h_A-h'_A\|_2^2/\{8(1-a_A^2)\}$. Marginalization
does not decrease affinity, so these bounds give the Hellinger claim. The same
argument on the contracted grid, with capped blocks $f_A,f'_A$ in place
of $h_A,h'_A$ and $a_{\mathrm{in}}$ in place of $a_A$, proves
Lemma~\ref{lem:coupling}.
\end{proof}
With the spins as latent variable and single edges in place of blocks
the argument uses only that every bond $B_e$ is a fair sign. It
therefore gives, on any finite simple
graph and for centered channels bounded by $a<1$,
\begin{equation}\label{eq:edge-coupling}
 \KL(p_h\Vert p_{h'})\leq\frac1{2(1-a^2)}\sum_{e\in E}\|h_e-h'_e\|_2^2 .
\end{equation}

\noindent\emph{Lower bound.}
\begin{proof}
Fix a block $A$ with witness $(C_A,\mathcal F_A,\vartheta_A)$, and let
$p,p'$ be the densities of $U_{W_A}$ under the two laws, with
$H_A^2=H^2(p,p')$. Write $\gamma_A,\theta_F$ and $\gamma'_A,\theta'_F$ for
the scales and moments of Lemma~\ref{lem:witness-identities} under the
two laws. Since $U_A$ is uniform under both, on each fiber
$\{U_A=u_A\}$ both densities integrate to one and
$\int(\sqrt p+\sqrt{p'})^2\leq4$ there. Writing
$p-p'=(\sqrt p-\sqrt{p'})(\sqrt p+\sqrt{p'})$, Cauchy--Schwarz on each
fiber and on the whole space, with $|\phi_F|\leq1$
and~\eqref{eq:block-regression}, gives
\[
 \|\gamma_Ah_A-\gamma'_Ah'_A\|_2\leq2\sqrt2\,H_A,\qquad
 |\theta_F-\theta'_F|\leq2\sqrt2\,H_A\quad(F\in\mathcal F_A).
\]
Both $\theta_F$ and $\theta'_F$ are at least $\kappa^{|F|}$, so
$|\log\theta_F-\log\theta'_F|\leq2\sqrt2\,\kappa^{-|F|}H_A$ and,
by~\eqref{eq:block-regression},
$|\log\gamma_A-\log\gamma'_A|\leq2\sqrt2\,H_A
\sum_F|\vartheta_{A,F}|\kappa^{-|F|}$. Exchanging the laws if needed, let
$\gamma_A\geq\gamma'_A$. Then
$h_A-h'_A=(\gamma_Ah_A-\gamma'_Ah'_A)/\gamma_A
+h'_A(\gamma'_A-\gamma_A)/\gamma_A$, and
$\gamma_A\geq\kappa^{|C_A\setminus A|}$, $\|h'_A\|_2\leq a_A$ and
$1-\gamma'_A/\gamma_A\leq\log(\gamma_A/\gamma'_A)$ give
\begin{equation}\label{eq:local-stability}
 \|h_A-h'_A\|_2\leq2\sqrt2\,\Gamma_AH_A .
\end{equation}

Fix a color. The vector $U_{W_A}$ is a function of the spins on the
vertices of $W_A$ and of the emission randomness on its edges, so the
vectors of one color, whose windows are vertex disjoint, are independent
under both laws. Marginalization does not decrease affinity. We multiply
affinities over the independent components to obtain
\[
 \Aff(p_h,p_{h'})\leq\prod_A(1-H_A^2)\leq\exp\Bigl(-\sum_AH_A^2\Bigr),
\]
with the product and sum over the blocks of that color. Averaging the
exponent over the $\chi$ colors and using~\eqref{eq:local-stability}, we obtain
\[
 -\log\Aff(p_h,p_{h'})\geq\frac1\chi\sum_AH_A^2
 \geq\sum_A\frac{\|h_A-h'_A\|_2^2}{8\chi\,\Gamma_A^2},
\]
now over all blocks. The middle inequality
of~\eqref{eq:global-divergence} is $-2\log\Aff\leq D$.
\end{proof}

\noindent\emph{Proof of Corollary~\ref{cor:grid-geometry}.}
Lemmas~\ref{lem:identify} and~\ref{lem:witnesses} give a witness system
with $24$ colors and $\Gamma_A\leq\Gamma$, no block of the grid is a
cycle, and $a_A\leq a$. The remaining claims follow from
$H^2=1-\Aff\leq-\log\Aff$, $H^2\leq\TV$ and $H^2\leq1$.

\subsection{Cycle Blocks and Single-Channel Information}
\begin{remark}[Cycle blocks]\label{rem:cycle-blocks}
A block $A$ that is a cycle factors from the remaining law:
$p_h=(1+h_A)\otimes p_{h|E\setminus A}$, so affinities multiply and KL
adds. For this factor, with $D$ denoting either directed KL divergence,
$\|h_A-h'_A\|_2^2/\{8(1+a_A)\}\leq H^2\leq\tfrac12D\leq
\|h_A-h'_A\|_2^2/\{4(1-a_A)\}$.
The lower bound follows from~\eqref{eq:local-hellinger}. Taylor's theorem
for $F(x)=(1+x)\log(1+x)$ bounds
$(1+x)\log\{(1+x)/(1+y)\}-(x-y)$ by $(x-y)^2/\{2(1-a_A)\}$
on $[-a_A,a_A]$; centering cancels $x-y$ after integration.
\end{remark}

\begin{remark}[Bond information from one cycle]\label{rem:bond-information}
On any graph, let $\varsigma=\E[B_e\mid U_{-e}]$, where
$U_{-e}=(U_f)_{f\ne e}$, and let $C$ be a cycle through $e$. Then
\[
 \E\varsigma^2\geq\prod_{f\in C\setminus\{e\}}\eta_f^2 ,
\]
and if $h'$ differs from $h$ only at $e$, with both channels bounded by
$a$,
\[
 \frac{\E\varsigma^2}{8(1+a)}\|h_e-h'_e\|_2^2\leq H^2(p_h,p_{h'})
 \leq-\log\Aff(p_h,p_{h'})\leq\frac{\E\varsigma^2}{2(1-a)}\|h_e-h'_e\|_2^2 .
\]
On the anchored grid every edge lies on a face, so
$\E\varsigma^2\geq\kappa^6$: one channel alone incurs a conditioning factor
$\kappa^{-6}$, and the $\kappa^{-12}$ of
Proposition~\ref{prop:geometry-sharp} arises only when several blocks
move together.
\end{remark}
\begin{proof}
The product $g=\prod_{f\in C\setminus\{e\}}\phi(U_f)$ is a function of
$U_{-e}$ with $|g|\leq1$. Since $\E[\phi(U_f)\mid B_f]=\eta_fB_f$ and
the bonds around $C$ multiply to one,
$\E[\varsigma g]=\E[B_eg]=\prod_{f\in C\setminus\{e\}}\eta_f$, and
Cauchy--Schwarz gives the first display. For the second display, we write
$p_h=p_{-e}\{1+h_e(u_e)\varsigma\}$, with $p_{-e}$ the common marginal
density of $U_{-e}$.
The inequality $\KL(p\Vert q)\leq\int(p-q)^2/q$ and
$1+h'_e\varsigma\geq1-a$ give the upper bound after integration;
use $-2\log\Aff\leq\KL$. For $|x|,|y|\leq a$,
$(\sqrt{1+x}-\sqrt{1+y})^2=(x-y)^2/(\sqrt{1+x}+\sqrt{1+y})^2
\geq(x-y)^2/\{4(1+a)\}$, so
\[
 H^2(p_h,p_{h'})=\tfrac12\,\E\int_0^1\Bigl(\sqrt{1+h_e\varsigma}
 -\sqrt{1+h'_e\varsigma}\Bigr)^2du_e
 \geq\frac{\E\varsigma^2}{8(1+a)}\|h_e-h'_e\|_2^2 ,
\]
with the expectation over $U_{-e}$, and $H^2\leq-\log\Aff$.
\end{proof}

\subsection{Proof of Proposition~\ref{prop:geometry-sharp}}\label{app:geometry-sharp}
\begin{proof}
\emph{Lower bound.} Corollary~\ref{cor:grid-geometry} gives
$c^\star\geq(192\Gamma^2)^{-1}$. Since $\kappa<a$, we have
$\kappa^{-3}\leq a\kappa^{-4}$, so
$\Gamma\leq a\kappa^{-6}(\tfrac32+4\kappa^2)$. For $\kappa\leq1/10$,
$4\kappa^2<1/10$, hence
$\Gamma\leq\tfrac85a\kappa^{-6}$ and
$(192\Gamma^2)^{-1}\geq\kappa^{12}/(500a^2)$.

\emph{A path of binary laws.} Take $L=4$, so $d=24$ and the cycle space
has dimension $d-16+1=9$. Index vertices by (row, column), let $w=(0,1)$,
$A=\{(0,1),(1,1)\}$, $e_1=\{(0,1),(0,2)\}$ and $e_2=\{(0,0),(0,1)\}$; the
edge $e_2$ belongs to the corner pair $Q$ at $(0,0)$, while $A$ and
$e_1$ are scalar blocks. Binary channels $h_e=\eta_e\phi$ with
$\kappa\leq\eta_e\leq a$ are centered, bounded by $a$ and anchored. For
$0\leq x\leq\log(a/\kappa)$ put $\eta_A(x)=ae^{-x}$,
$\eta_{e_1}(x)=\kappa e^x$, $\eta_e(x)=\kappa e^{x/2}$ for $e\in Q$ and
$\eta_e(x)=\kappa$ otherwise, so $\eta_e(x)=\eta_e(0)e^{xv_e}$ with
$v_A=-1$, $v_{e_1}=1$, $v_e=\tfrac12$ on $Q$ and $v_e=0$ otherwise. Every
$h(x)$ is anchored. With $\phi_F(u)=\prod_{e\in F}\phi(u_e)$, the
loop expansion of~\eqref{eq:density} gives
\[
 p_x(u)=\sum_{F\text{ even}}\theta_F(x)\phi_F(u),\qquad
 \theta_F(x)=\theta_F(0)\exp\bigl(x\langle v,\mathbf 1_F\rangle\bigr),
\]
and the functions $\phi_F$ are orthonormal in $L^2(\mu)$.

\emph{Parity at one vertex.} The edges incident to $w$ are
$\{A,e_1,e_2\}$. An even set
meets it in zero or two edges, and it contains both edges of $Q$ or
neither, since the corner $(0,0)$ has degree two. Hence for every even
$F$,
\[
 \langle v,\mathbf 1_F\rangle=\mathbf 1[e_1\in F]+\mathbf 1[e_2\in F]-\mathbf 1[A\in F]
 =2\cdot\mathbf 1[F\cap\{A,e_1,e_2\}=\{e_1,e_2\}].
\]
So $\dot\theta_F=2\theta_F$ on the family $\mathcal F$ of even sets that
pass through $w$ and avoid $A$, and $\dot\theta_F=0$ otherwise, where the
dot is the derivative at $x=0$. The two faces at $w$ meet the star in
$\{A,e_2\}$ and $\{A,e_1\}$, so $F\mapsto(\mathbf 1[e_1\in F],\mathbf 1[e_2\in F])$
maps the $512$ even sets onto $\{0,1\}^2$ and each fiber has $128$
elements; $\mathcal F$ is the fiber over $(1,1)$. The grid is bipartite,
so even sets have even size. Both faces at $w$ contain $A$, and the
six-cycles of the grid are the boundaries of two adjacent faces; exactly
one of them passes through $w$ and avoids $A$, the boundary of the two
faces at $w$; every other pair containing a face at $w$ has $A$ on its
boundary. Thus $\mathcal F$ has one set of size six and $127$ sets
of size at least eight, all avoiding $A$, so $\theta_F(0)=\kappa^{|F|}$
on $\mathcal F$ and
\[
 \sum_F\dot\theta_F^2=4\sum_{F\in\mathcal F}\kappa^{2|F|}
 \leq4\kappa^{12}(1+127\kappa^4).
\]

\emph{Information and distance.} Every nonempty even set has at least
four edges, at most one of which ($A$) has reliability above $\kappa$,
so $\theta_F(0)\leq\kappa^3$ and $p_0\geq1-511\kappa^3>0$.
Orthonormality gives
\[
 I=\int\frac{\dot p^2}{p_0}\,d\mu\leq\frac{\sum_F\dot\theta_F^2}{1-511\kappa^3}
 \leq\frac{4\kappa^{12}(1+127\kappa^4)}{1-511\kappa^3}.
\]
The density $p_x$ is a finite sum with smooth coefficients and is bounded
below near $x=0$, so expanding $\sqrt{p_x/p_0}$ to second order and using
$\int\dot p\,d\mu=\int\ddot p\,d\mu=0$ gives
$-\log\Aff(p_0,p_x)=Ix^2/8+O(x^3)$. The blocks that move are $A$, $e_1$
and the corner product on $Q$, whose channels are $ae^{-x}\phi$,
$\kappa e^x\phi$ and $\kappa^2e^x\phi\otimes\phi$, so
$\Delta^2(h(0),h(x))=(a^2+\kappa^2+\kappa^4)x^2+O(x^3)$. If
$c\Delta^2\leq-\log\Aff$ along the path, letting $x\downarrow0$ gives
\[
 c\leq\frac I{8(a^2+\kappa^2+\kappa^4)}
 \leq\frac{\kappa^{12}(1+127\kappa^4)}{2(1-511\kappa^3)\,a^2},
\]
which is at most $1.04\,\kappa^{12}/a^2$ for $\kappa\leq1/10$, since
then $511\kappa^3\leq0.511$ and $127\kappa^4\leq0.0127$. The two
bounds differ by a factor $520$, and the upper one is $o(\kappa^{10})$
at fixed $a$.

\emph{Lipschitz channels.} Replace $\phi$ by $g(u)=2u-1$, which is
centered and bounded by one with $\int\phi g=1/2$ and $\int g^2=1/3$,
and take $h_e=s_eg$ with $s_A(x)=a_Re^{-x}$,
$s_{e_1}(x)=2\kappa e^x$, $s_e(x)=2\kappa e^{x/2}$ on $Q$ and
$s_e=2\kappa$ otherwise. For $0\leq x\leq\log\{a_R/(2\kappa)\}$ every
channel is bounded by $a_R\leq a$, is $2a_R$-Lipschitz with $2a_R\leq R$,
and has reliability $s_e/2\geq\kappa$, so the path lies in
$\mathcal P_L(a,R,\kappa)$. The products $g_F=\prod_{e\in F}g(u_e)$ over
even sets are orthogonal in $L^2(\mu)$ with $\|g_F\|_2^2=3^{-|F|}$ and
$|g_F|\leq1$, and the same parity identity holds, so with
$\kappa'=2\kappa$ the steps above give
$p_0\geq1-511\kappa'^3$,
$I\leq4\kappa'^{12}3^{-6}(1+127\kappa'^4/9)/(1-511\kappa'^3)$ and
$\Delta^2=\tfrac13(a_R^2+\kappa'^2+\kappa'^4/3)x^2+O(x^3)$. Hence every
$c$ valid on this class satisfies
$c\leq(2\kappa)^{12}(1+127\cdot16\kappa^4/9)/\{486(1-4088\kappa^3)a_R^2\}$,
which is at most $18\kappa^{12}/a_R^2$ for $\kappa\leq1/20$.
\end{proof}

\begin{proof}(Corollary~\ref{cor:recovery-lower})
We use the preceding Lipschitz path on the $4\times4$ grid, with
$g(u)=2u-1$ and $0\leq x\leq1/4$. Since $4\kappa\leq a_R$,
every channel remains bounded by $a_R$, is $R$-Lipschitz and has
reliability at least $\kappa$. The parity identity gives
\[
 p_x-p_0=(e^{2x}-1)\sum_{F\in\mathcal F}(2\kappa)^{|F|}g_F,
\]
where $\mathcal F$ has one set of size six and $127$ of size at least
eight. Orthogonality, $\|g_F\|_2^2=3^{-|F|}$ and
$p_0\geq1-4088\kappa^3$ imply
\begin{align*}
 \KL(p_x\Vert p_0)&\leq\chi^2(p_x\Vert p_0)\\
 &\leq\frac{(e^{2x}-1)^2(2\kappa)^{12}}
 {3^6(1-4088\kappa^3)}
 \left(1+\frac{127(2\kappa)^4}{9}\right)
 \leq128\kappa^{12}x^2.
\end{align*}
For the last inequality use $e^{2x}-1\leq2e^{1/2}x$ and
$\kappa\leq1/20$. The scalar block $A$ alone gives
\[
 \Delta^2(h(0),h(x))\geq\frac{a_R^2}{3}(1-e^{-x})^2
 \geq\frac{a_R^2x^2}{5},
\]
since $1-e^{-x}\geq xe^{-1/4}$ and $e^{-1/2}/3>1/5$.
Choose $x=\min\{1/4,(32\sqrt N\,\kappa^6)^{-1}\}$. Additivity
of KL and Pinsker's inequality give
$\TV(p_0^{\otimes N},p_x^{\otimes N})\leq1/4$.
Assigning an estimated block collection to the nearer of the two
endpoints incurs squared loss at least $\Delta^2(h(0),h(x))/4$
on a testing error. Thus the larger endpoint risk is at least
$3\Delta^2(h(0),h(x))/32\geq3a_R^2x^2/160$.
As $x^2\geq\min\{1,(N\kappa^{12})^{-1}\}/1024$ and
$3/163840\geq1/65536$, the claimed bound follows.
\end{proof}

\subsection{Proof of Theorem~\ref{thm:learning}}
\noindent\emph{The estimator.}
We first specify the estimator with inputs $a_{\mathrm{in}}=a$ and
$\kappa_{\mathrm{in}}=\kappa$.
We use all $N$ observations in both stages, $N_0=N_1=N$; the proof also
covers the split $N_0=\lfloor N/2\rfloor$, $N_1=N-N_0$. Estimate each
cycle moment from $N_0$ vectors and clip it to
$[\kappa^{|F|},1]$. For an interior block, clip the two-face ratio
$\sqrt{\widehat\theta_C\widehat\theta_{C\triangle C'}/\widehat\theta_{C'}}$
to $[\kappa^3,1]$ to obtain $\widehat\gamma_A$. For a boundary block, clip each root
in~\eqref{eq:short-calibration} to $[\kappa,a]$ and multiply the two or
three reliabilities to obtain $\widehat\gamma_A$. On $N_1$ vectors
estimate each $g_i$ by the centered linear-spline $L^2$ projection on $K$
uniform cells,
\[
 \widehat g_i=\sum_{j=1}^{K}\Bigl\{\frac1{N_1}\sum_{r=1}^{N_1}
 \tilde b_j(U_i^{(r)})\prod_{f\in C\setminus\{i\}}\phi(U_f^{(r)})\Bigr\}\tilde b_j,
\]
where $\tilde b_1,\ldots,\tilde b_K$ is an orthonormal basis of the centered
spline space. We identify each spline with its $K+1$ nodal values and use
the trapezoidal masses $\omega=(1/2,1,\ldots,1,1/2)/K$ of the hat
functions $b_0,\ldots,b_K$. For a scalar
block, we project $\widehat g_i/\widehat\gamma_A$ in the weighted norm
$\|v\|_\omega^2=\sum_j\omega_jv_j^2$ onto
$\{v:\omega^\top v=0,\ |v_j|\leq a\}$; with $z=(z_0,\ldots,z_K)$ the nodal
values of $\widehat g_i/\widehat\gamma_A$, the solution is
$v_j=\clip(z_j-t,-a,a)$ with the scalar $t$ chosen to
make the weighted mean zero. For a corner, we project both regressions in $\|\cdot\|_\omega$ onto
centered vectors bounded by one (or any fixed bound between $a^4$ and
one), divide their outer product by
$\widehat\gamma_A\widehat\theta_C$, and project the result onto
\[
 \mathcal B_a=\{V:\ V\omega=0,\ \omega^{\top}V=0,\ |V_{ij}|\leq a^2\}
\]
with squared norm $\|V\|_{\omega\otimes\omega}^2=\sum_{ij}\omega_i\omega_jV_{ij}^2$.
Bilinear interpolation gives a bounded corner channel centered in each argument. All
projections are onto closed convex sets and are computed exactly
(Lemma~\ref{lem:spline-time}). The fitted law is a normalized member of
the planar block family, with uniform forest marginals. Integrating out
either corner coordinate deletes that block by separate centering.
Suppressing the retained degree-two corners of a forest leaves a forest,
whose loop polynomial is one.

\begin{proof}
For a linear spline $v$ with nodal values $v_i$, integrating $v^2$
over one cell gives $(v_i^2+v_iv_{i+1}+v_{i+1}^2)/(3K)$. Thus
$\tfrac13\|v\|_\omega^2\leq\|v\|_2^2\leq\|v\|_\omega^2$, and
$\tfrac19,1$ for tensors. For centered bounded $R$-Lipschitz $h$, the
hat averages $h_j^*=\int hb_j/\omega_j$ are centered, bounded by $a$,
and interpolate $h$ within $2R/K$; apply the same construction to each
$g_i$. A uniform design and a response bounded by one give total
coefficient variance at most $K/N_1$, so best approximation and norm
equivalence yield
$\E\|\widehat g_i-g_i^*\|_\omega^2\leq C_3(K/N_1+R^2/K^2)$. Projection in
$\|\cdot\|_\omega$ onto a closed convex set containing $g_i^*$ does not
increase this error.

Each clipped moment has mean-squared error at most $1/N_0$.
We bound the same clipped ratios in logarithmic coordinates.
For $0<l\leq u$, the map $t\mapsto\clip(e^t,l,u)$ is $u$-Lipschitz.
Also, $\log$ is $\kappa^{-|F|}$-Lipschitz on $[\kappa^{|F|},1]$.
Each reliability ratio has three log-moment coefficients $\pm1/2$ and
moment lengths at most six. Cauchy--Schwarz therefore gives
$\E|\widehat\eta_f-\eta_f|^2\leq9a^2/(4N_0\kappa^{12})$.
For $x_j,y_j\in[0,1]$,
$|\prod_{j\leq k}x_j-\prod_{j\leq k}y_j|^2\leq k\sum_{j\leq k}|x_j-y_j|^2$,
so with $k\leq3$ the boundary scale has mean-squared error at most
$81a^2/(4N_0\kappa^{12})$. For the interior ratio, the same argument
with upper cap one gives $9/(4N_0\kappa^{12})$.
No independence between overlapping moments is used.
Scalar denominators are at least $\kappa^3$; at a corner
$\widehat\gamma_A\widehat\theta_C\geq\kappa^6$, with mean-squared error at
most twice the sum of the two factor errors. For either nodal norm and
$\gamma,\widehat\gamma\geq c>0$,
\[
 \Bigl\|\frac{\widehat f}{\widehat\gamma}-\frac f\gamma\Bigr\|
 \leq\frac{\|\widehat f-f\|}{c}+\frac{\|f\|\,|\widehat\gamma-\gamma|}{c^2},
\]
so the separate mean-squared errors bound the expected quotient
error. This argument does not require independent numerators and
denominators, so we can use the same vectors in both stages. Use $c=\kappa^3$ for scalars and
$c=\kappa^6$ for corners; target regressions have norm at most one. For
bounded nodal vectors,
$\|uv^{\top}-xy^{\top}\|_{\omega\otimes\omega}^2
\leq2\|u-x\|_\omega^2\|v\|_\omega^2+2\|x\|_\omega^2\|v-y\|_\omega^2$.
Linearity of the hat averages preserves the regression identities at the
nodes, so $g_1^*g_2^{*\top}/(\gamma_A\theta_C)=h_1^*h_2^{*\top}\in\mathcal B_a$.
These bounds and nonexpansiveness of the final projection give squared
$L^2$ error $C_4(K/N+K^{-2}+N^{-1})$ for every block, including the
interpolation bias.

Both laws lie in the planar block family with channels bounded by $a$,
so Lemma~\ref{lem:coupling} sums these block errors into either
directed KL.
\end{proof}

For conservative inputs, use $a_{\mathrm{in}}$ in projection and clipping bounds
and $\kappa_{\mathrm{in}}$ in clipping floors. The true
moments and comparison functions remain feasible,
and Lemma~\ref{lem:coupling} uses $\{2(1-a_{\mathrm{in}}^2)\}^{-1}$.

\noindent\emph{Exact splines.}
If every $h_i$ is a centered linear spline on the chosen grid, so is each
$g_i$, and the true splines can replace the hat averages above. The
approximation terms vanish, giving $C(a_{\mathrm{in}},\kappa_{\mathrm{in}})d(K/N+1/N)$
in either direction.

\begin{lemma}[Running time]\label{lem:spline-time}
Counting real arithmetic operations, comparisons and square roots, and
ignoring bit complexity, the estimator of Theorem~\ref{thm:learning} is
computed exactly with $O(dN\log(K+1)+dK\log(K+1)+K^8\log(K+1))$
operations, $O(d)$ of them square roots.
\end{lemma}

\begin{proof}
\emph{Moments and regressions.} Each block uses at most seven cycle
moments of order four or six and at most three square roots, so
calibration costs $O(dN)$. There are $d$ regressions, one per scalar
block and two per corner pair. Bisection locates the cell of a reading
in $O(\log(K+1))$ comparisons, after which two hats are nonzero, so
$y_k=N_1^{-1}\sum_rb_k(U_i^{(r)})\phi_{C\setminus\{i\}}(U^{(r)})$,
$k=0,\ldots,K$, costs $O(N\log(K+1))$. The Gram matrix $Q$ of
$b_0,\ldots,b_K$ in $L^2[0,1]$ is tridiagonal with $Q\mathbf 1=\omega$,
where $\mathbf 1$ is the all-ones vector, since $\sum_kb_k=1$.
As $\widehat g_i$ represents the empirical
functional on the centered spline space, its nodal vector is the
$Q$-orthogonal projection of $Q^{-1}y$ onto $\{v:\omega^\top v=0\}$,
namely $Q^{-1}y-(\omega^\top Q^{-1}y)\mathbf 1$: one tridiagonal
solve.

\emph{Scalar projections.} For nodal values $z$ and bound $a$ (bound $1$
for the eight corner regressions),
$t\mapsto\sum_j\omega_j\clip(z_j-t,-a,a)$ is
continuous, nonincreasing and piecewise linear with breakpoints
$z_j\pm a$. After sorting the $2K+2$ breakpoints, a sweep updates the
value by the slope times the gap and the slope by $\pm\omega_j$, at
$O(1)$ cost per breakpoint, until the value is nonpositive; linear
interpolation then gives the exact shift $t$, in $O(K\log(K+1))$ per
projection.

\emph{Corner projections.} For the matrix $Z$ projected, the
substitution $x_{ij}=\omega_i\omega_j(V_{ij}+a^2)$ maps $\mathcal B_a$ onto the flows
on the complete bipartite network with $K+1$ row and $K+1$ column nodes,
supply $a^2\omega_i$ at row node $i$, demand $a^2\omega_j$ at column node
$j$ and capacities $0\leq x_{ij}\leq2a^2\omega_i\omega_j$. It maps the
objective $\sum_{ij}\omega_i\omega_j(V_{ij}-Z_{ij})^2$ to
$\sum_{ij}\{x_{ij}^2/(\omega_i\omega_j)-2x_{ij}(Z_{ij}+a^2)\}$ plus a
constant. By \citet[Theorem~20]{vegh2016}, this capacitated problem with
$(K+1)^2$ arcs has an exact solution in $O((K+1)^8\log(K+1))$
arithmetic operations and comparisons; the outer products add $O(K^2)$.
\end{proof}

\subsection{Proof of Theorem~\ref{thm:minimax}}
\begin{theorem}[Minimax lower bounds]\label{thm:minimax}
Fix $0<a<1$, $R>0$ and $0<\kappa<\eta_*(a,R)$. There is
$c=c(a,R,\kappa)>0$ such that for $L\geq4$, $N\geq1$,
\[
 \mathcal R_N(H^2)\geq c\min\{1,dN^{-2/3}\},\qquad
 N\leq cd^{3/2}\ \Longrightarrow\ \mathcal R_N(\TV)\geq1/8.
\]
Together with Theorem~\ref{thm:learning}, these bounds give the
Hellinger, directed KL and TV sample rates in Theorem~\ref{thm:separation}.
\end{theorem}
\begin{proof}
\emph{Reliability slack.} Since $\kappa<\eta_*(a,R)$, we can shrink
the extremal ramp of Appendix~\ref{app:reliability-ceiling} slightly
to choose $a_\star<a$ and $\tau\in(0,1]$ with
$2a_\star/\tau<R$ and $a_\star(1-\tau/2)>\kappa$.

\emph{Packing.} Fix such $a_\star,\tau$, put $h_0=a_\star\ramp_\tau$, and
choose $m=\lfloor L/2\rfloor^2$ vertex-disjoint square faces. Set every
channel to $h_0$ and vary one edge of each selected face. Let $\psi$ be
the piecewise-linear function with values $0,1,-1,0$ at $0,1/4,3/4,1$,
extended by zero; it has integral zero, squared integral $1/3$ and
Lipschitz constant four. For signs $\sigma_j\in\{-1,1\}$ put
\[
 h_\sigma(u)=h_0(u)+\frac{\varepsilon_\psi}{K}\sum_{j=0}^{K-1}
 \sigma_j\psi\bigl(2K(u-1/2)-j\bigr),
\]
with fixed $\varepsilon_\psi>0$ such that $a_\star+\varepsilon_\psi<a$ and
$2a_\star/\tau+8\varepsilon_\psi<R$. The bumps have disjoint supports in
$[1/2,1]$, where $\phi=1$, so they preserve centering and the anchor
$a_\star(1-\tau/2)>\kappa$, and every channel stays in the class. One
differing sign contributes $2\varepsilon_\psi^2/(3K^3)$ to the squared
channel distance.

Greedy deletion of Hamming balls of radius $\lfloor mK/4\rfloor$ gives a
code $\mathcal V\subset\{-1,1\}^{mK}$ with pairwise distance at least
$mK/4$ and $\log|\mathcal V|\geq c_{\mathcal V}mK$, where
$c_{\mathcal V}=\log2-\mathsf H(1/4)>0$ and $\mathsf H$ is binary entropy in nats.
Distinct codewords have total squared channel distance between
$\varepsilon_\psi^2m/(6K^2)$ and $2\varepsilon_\psi^2m/(3K^2)$.

On a selected face the marginal density is
$1+h_\sigma(u_1)h_0(u_2)h_0(u_3)h_0(u_4)$, bounded by $1+a^4$, and
$\|h_0\|_2^2=a_\star^2(1-2\tau/3)$. The selected face margins use
disjoint spins, so, as in the proof of
Theorem~\ref{thm:witness-geometry}, marginalization, tensorization of
affinity and~\eqref{eq:local-hellinger} give
\[
 H^2(p_\sigma,p_{\sigma'})\geq1-\exp(-c_{\mathrm{TV}}m/K^2),\qquad
 c_{\mathrm{TV}}=\frac{\varepsilon_\psi^2\{a_\star^2(1-2\tau/3)\}^3}{48(1+a^4)},
\]
and $\TV\geq H^2$ gives the same bound for TV. We obtain this separation
directly from face marginals with three known channels $h_0$. No calibration
is needed, and the constant $c_{\mathrm{TV}}$ contains no negative power
of $\kappa$.
The edgewise coupling bound~\eqref{eq:edge-coupling} gives
$\KL(p_\sigma^{\otimes N}\Vert p_{\sigma'}^{\otimes N})\leq
C_0Nm/K^2$ with $C_0=\varepsilon_\psi^2/[3(1-a^2)]$.

\emph{Lower bounds.} Take $K=\lceil c_KN^{1/3}\rceil$ with $c_K\geq4$ and
$c_K^3\geq8C_0/c_{\mathcal V}$; the packing is admissible for every $K$.
With $\sigma$ uniform on the code and
$\mathbf U=(U^{(1)},\ldots,U^{(N)})$ the sample, Fano's inequality
\citep{scarlett} gives
$\mathsf H(\sigma\mid\mathbf U)\leq\log2+P_{\mathrm{err}}\log|\mathcal V|$
for every decoder, and $\mathsf I(\sigma;\mathbf U)$ is at most the
largest pairwise sample KL. Its ratio to the code entropy is at most
$C_0N/(c_{\mathcal V}K^3)\leq1/8$. Since $K\geq4$ and $m\geq4$ for
$L\geq4$, and $c_{\mathcal V}=\tfrac34\log3-\log2$,
$\log2/(c_{\mathcal V}mK)\leq\log2/(16c_{\mathcal V})\leq3/8$, the last
step being $2^{14}\leq3^9$. So every decoder errs with probability at
least $1/2$.

Since $K\leq2c_KN^{1/3}$ and $m\geq d/10$ for $L\geq4$, distinct code
laws have $H^2$ and TV at least $\underline H^2=1-e^{-x}$, with
$x=c_{\mathrm{TV}}m/K^2\geq c_0dN^{-2/3}$ and
$c_0=c_{\mathrm{TV}}/(40c_K^2)$. Decode an estimate $\widetilde p$ to the
nearest code law in the Hellinger metric $\sqrt{H^2}$, respectively in
TV; on an error the triangle inequality gives
$H^2(p_\sigma,\widetilde p)\geq\underline H^2/4$, respectively
$\TV(p_\sigma,\widetilde p)\geq\underline H^2/2$. The worst-case expected
losses are therefore at least
$\underline H^2/8\geq\tfrac18(1-e^{-1})\min\{1,c_0dN^{-2/3}\}$ and
$\underline H^2/4$. The latter is at least $1/8$ when $x\geq\log2$, in
particular when $N\leq(c_0/\log2)^{3/2}d^{3/2}$; take $c$ in the theorem
as the smaller of $\tfrac18(1-e^{-1})\min\{1,c_0\}$ and
$(c_0/\log2)^{3/2}$. Both KL losses dominate
$2H^2$, which gives $\mathcal R_N(H^2)\leq\tfrac12\mathcal R_N(D)$ and,
for $N\geq d^{3/2}$, the KL lower bound; $H^2\leq1$ gives the other cap.

\emph{Upper bounds.} With $K=\lceil N^{1/3}\rceil$,
$K/N+K^{-2}+N^{-1}\leq4N^{-2/3}$. Theorem~\ref{thm:learning}
bounds either KL risk by $CdN^{-2/3}$ for $N\geq4$.
Pinsker and Jensen give $\E\TV\leq\sqrt{CdN^{-2/3}/2}$,
so $N_{\mathrm{TV}}(\delta)\leq C'\delta^{-3}d^{3/2}$ for
$0<\delta<1/8$. The lower TV bound gives $N_{\mathrm{TV}}(\delta)>cd^{3/2}$.
\end{proof}

\subsection{Proof of Theorem~\ref{thm:anchor-free}}\label{app:anchor-free}
\begin{proof}
\emph{Lower bounds.} Put $\kappa_1=\eta_*(a,R)/2$. For $0<\kappa\leq\kappa_1$,
$\mathcal P_L(a,R,\kappa_1)\subseteq\mathcal P_L(a,R,\kappa)\subseteq
\mathcal P^0_L(a,R)$, so every minimax risk over the last two classes is
at least the risk over the first. Theorem~\ref{thm:minimax} at the anchor
$\kappa_1$, which is a function of $a$ and $R$, gives
$\mathcal R_N(H^2)\geq c\min\{1,dN^{-2/3}\}$ for $N\geq1$ and
$\mathcal R_N(\TV)\geq1/8$ for $N\leq cd^{3/2}$, with $c=c(a,R)$.

\emph{A net of channels.} Let $\mathcal H$ be the set of centered
$R$-Lipschitz functions $[0,1]\to[-a,a]$. If $h\in\mathcal H$ and
$|h(u_0)|=\|h\|_\infty$ with, say, $h(u_0)>0$, then
$h(u)\geq\|h\|_\infty-R|u-u_0|$, and integrating gives
$0\geq\|h\|_\infty-R/2$; hence $\|h\|_\infty\leq R/2$.
We construct the usual Lipschitz net \citep[\S2]{kolmogorov1959} directly.
For $r\geq R/2$ take $\mathcal H_r=\{0\}$. Otherwise put
$m=\lceil4R/r\rceil$ and round each value $h(j/m)$ to the nearest
multiple of $r/4$, with a fixed rule for ties. There are at most
$4R/r+2$ choices for the first value and at most five for each next
value, since adjacent rounded values differ by at most $r/2$.
Two functions with the same rounded values differ by at most $r/4$
at a grid point and at most $r/4+2R/m\leq3r/4$ everywhere.
Choose one member of $\mathcal H$ for each pattern that occurs. This gives
a finite sup-norm $r$-net $\mathcal H_r\subseteq\mathcal H$ with
$\log|\mathcal H_r|\leq\log(4R/r+2)+m\log5\leq c_1R/r$
for an absolute constant $c_1$.

\emph{Mixture.} Let $G$ be a finite simple graph with $d$ edges and let
$\mathcal N$ be the set of laws $p_g$ on $G$ with $g_e\in\mathcal H_r$
for every edge, so $\log|\mathcal N|\leq c_1Rd/r$. For $p=p_h$ with
channels in $\mathcal H$, choose $p_g\in\mathcal N$ with
$\|h_e-g_e\|_\infty\leq r$ on every edge;~\eqref{eq:edge-coupling} gives
$\KL(p\Vert p_g)\leq dr^2/\{2(1-a^2)\}$. Let
$q_n=|\mathcal N|^{-1}\sum_{p_g\in\mathcal N}p_g^{\otimes n}$ be the uniform mixture of
product laws on $n$ vectors, with $q_0=1$, let
\[
 \widehat q_i(u)=\frac{q_{i+1}(U^{(1)},\dots,U^{(i)},u)}
 {q_i(U^{(1)},\dots,U^{(i)})},\qquad
 \widehat p=\frac1N\sum_{i=0}^{N-1}\widehat q_i ,
\]
be its predictive density after $i$ observations and their average.
Following \citet{barron1987} and the proof of Theorem~2 in
\citet{yangbarron1999}, convexity of KL and the chain rule give
\[
 \E\KL(p\Vert\widehat p)
 \leq\frac1N\sum_{i=0}^{N-1}\E\KL(p\Vert\widehat q_i)
 =\frac1N\KL(p^{\otimes N}\Vert q_N)
 \leq\frac{\log|\mathcal N|}N+\KL(p\Vert p_g),
\]
where the last step uses $q_N\geq|\mathcal N|^{-1}p_g^{\otimes N}$. With
$r=\{R(1-a^2)/N\}^{1/3}$ both terms are multiples of
$R^{2/3}(1-a^2)^{-1/3}dN^{-2/3}$, so
\[
 \E\KL(p\Vert\widehat p)\leq C_2\,dN^{-2/3},\qquad
 C_2=(\tfrac12+c_1)R^{2/3}(1-a^2)^{-1/3},
\]
for every $N\geq1$, every such graph and every law with channels in
$\mathcal H$. This bound requires neither an anchor nor a grid.

\emph{Upper bounds.} On the grid this is
$\mathcal R_N(D,\mathcal P^0_L(a,R))\leq C_2dN^{-2/3}$ for the forward
loss, and $H^2\leq\min\{1,\tfrac12\KL\}$ gives the Hellinger bound.
Pinsker's and Jensen's inequalities give $\E\TV\leq\delta$ once
$N\geq(C_2d/(2\delta^2))^{3/2}$, so the minimax sample size over either
class is at most $\{(C_2/2)^{3/2}+1\}\delta^{-3}d^{3/2}$; take
$C=\max\{C_2,(C_2/2)^{3/2}+1\}$.
\end{proof}
\section{Positive Representation Proofs}\label{app:representation-proofs}
We first decompose a circuit into products across fixed cuts, then use
crossing cycles to bound their affinity with the target law.

\subsection{Balanced Products}
We state the decomposition after conditioning on observations outside a
selected block. This form supports both the density and prediction bounds.
\begin{lemma}[Balanced products on a selected block]\label{lem:balance}\label{lem:block-balance}
Let $q$ be a normalized smooth decomposable nonnegative circuit with
binary products and $M$ nodes. Let $T\subseteq E$, $|T|\geq2$, and
$0<\zeta\leq1/3$. Suppose every leaf scope meets $T$ in at most
$(1-\zeta)|T|$ coordinates. For $O\subseteq E\setminus T$ and almost
every $u_O$ with $0<q_O(u_O)<\infty$, the conditional marginal
$q_{T\mid u_O}$ is a mixture of at most $M$ normalized products
across $\zeta$-balanced cuts of $T$. The cuts may differ between terms.
They belong to a family $\mathcal A(q,T)$ of at most $M$ cuts fixed by
the declared scopes, independently of $O$ and $u_O$.
\end{lemma}
Taking $T=E$, $O=\varnothing$ and $\zeta=\zeta_\alpha$ gives the
full-scope decomposition used in Proposition~\ref{prop:balanced-affinity}.
\begin{proof}
We adapt the balanced decompositions of \citet[Theorem~38]{martens} and
\citet[Theorem~6]{bova2016} to the scope measure $|A\cap T|$.
Start with root context one. At sums, pass the weighted context to
each child. At products, follow the child of largest measure, breaking
ties by a fixed node order, and multiply the context by its sibling.
Stop when the measure first falls to at most $(1-\zeta)|T|$.
Leaves meet this threshold and sums preserve scope, so stopping follows
a product step. The stopped measure exceeds
$(1-\zeta)|T|/2\geq\zeta|T|$.

Each induced tree follows exactly one such path and stops at exactly one
node. Decomposability keeps contexts outside the current scope. Adding
contexts at shared stopped nodes gives at most $M$ balanced terms
$c_v(u_{E\setminus A_v})f_v(u_{A_v})$. Fix $u_O$ and integrate out
the other coordinates outside $T$. Tonelli factors each term across
$T\cap A_v\mid T\setminus A_v$; their integrals sum to $q_O(u_O)$.
Zero-integral terms vanish almost everywhere; positive terms have
finite positive factor integrals. Normalize the factors and divide
weights by $q_O(u_O)$. These cuts depend only on declared scopes,
including constant coordinates. Unused nodes need no normalization.
\end{proof}

\subsection{Proof of Proposition~\ref{prop:balanced-affinity}}\label{app:affinity-proof}
\begin{proof}
Write $q=\sum_{i=1}^m\omega_i r_i$, where $m\leq M$ and the $r_i$
are normalized balanced products (Lemma~\ref{lem:balance}). Then
\[
 \Aff(p,q)\leq\sum_{i=1}^m\sqrt{\omega_i}\,\Aff(p,r_i)
 \leq e^{-\Lambda/2}\sum_{i=1}^m\sqrt{\omega_i}
 \leq\sqrt M\,e^{-\Lambda/2}.
\]
The first step uses square-root subadditivity; the last uses
Cauchy--Schwarz and $\sum_i\omega_i=1$. Together with $\Aff\leq1$,
this proves~\eqref{eq:affinity}.
Jensen gives $-2\log\Aff\leq\KL$ in either direction, and
$1-\TV\leq\Aff$. For use in Lemma~\ref{lem:conditional-affinity},
affinity cannot decrease under a Markov kernel $\mathcal T$ to a finite
set: Cauchy--Schwarz gives
$\int\sqrt{pq}\,\mathcal T(z\mid\cdot)\leq
\sqrt{(\mathcal Tp)(z)(\mathcal Tq)(z)}$; sum over $z$.
\end{proof}

\subsection{Conditional Witnesses and Dependent Spins}\label{app:dependent-witnesses}
We prove the conditional affinity bound, then apply it to crossing cycles.
For a binary pair $(X,Y)$ with mass matrix $P$, let $A_P$ be its entrywise
square root. We use $\|\cdot\|_{\mathrm{op}}$ and $\|\cdot\|_{\mathrm F}$
for the spectral and Frobenius matrix norms. Square roots of product
probabilities are nonnegative unit vectors, and $A_P$ has nonnegative
maximizing singular vectors. Hence
$\sup_{f,g}\Aff(P,f\otimes g)=\|A_P\|_{\mathrm{op}}$.
Write $v=\operatorname{Cov}(X,Y)$.
For signs $X,Y\in\{\pm1\}$, $v=4\det P$, while
$\|A_P\|_{\mathrm F}^2=1$. If $v\ne0$, the arithmetic--geometric mean
inequality gives
\[
 |\det A_P|
 =\frac{|\det P|}{\sqrt{P_{++}P_{--}}+\sqrt{P_{+-}P_{-+}}}
 \geq2|\det P|=|v|/2.
\]
The two squared singular values sum to one and have product
$(\det A_P)^2$. Consequently
\begin{equation}\label{eq:pair-covariance-affinity}
 \sup_{f,g}\Aff(P,f\otimes g)^2
 \leq\frac{1+\sqrt{1-v^2}}2=e^{-\beta(v)}.
\end{equation}
For $v=0$ the bound follows from affinity being at most one.
For fair signs with correlation $\theta$, equality holds: the largest
singular value is $(\sqrt{1+\theta}+\sqrt{1-\theta})/2=e^{-\beta(\theta)/2}$.
Minimizing the order-$1/2$ R\'enyi divergence over product laws gives
the dependence measure of \citet[Eq.~(1)]{lapidoth2019}, expressed here
in nats rather than bits.

\begin{proof}(Lemma~\ref{lem:conditional-affinity}).
The kernels map every product law across the cut to a law
$Q_X\otimes Q_Y$. Conditional on earlier pairs, its next pair is still
a product:
\[
 Q(Z_j\mid Z_{<j})=Q_X(X_j\mid X_{<j})Q_Y(Y_j\mid Y_{<j}).
\]
We sum the affinity over the last pair first. By
\eqref{eq:pair-covariance-affinity} and monotonicity of $\beta$ in
absolute value, this conditional sum is at most
$e^{-\beta(c_k)/2}$. Repeating backwards bounds the joint affinity by
$\exp\{-\tfrac12\sum_j\beta(c_j)\}$. Histories having zero probability
under either law contribute zero. Affinity data processing, as in
Appendix~\ref{app:affinity-proof}, transfers the bound to the original
product law. Take the supremum over products and the infimum over cuts.
\end{proof}

\begin{proof}(Theorem~\ref{thm:dependent-barrier}).
Let $\sigma_e=\phi(U_e)$ and $\nu_e=B_e\sigma_e$. Centering $h_e$ gives
\[
 \Pr(\sigma_e=z\mid S)=\tfrac12(1+zB_e\eta_e),\qquad
 \Pr(\nu_e=z\mid S)=\tfrac12(1+z\eta_e).
\]
Thus the $\nu_e$ are mutually independent and independent of all spins.

Fix a balanced cut and choose vertex-disjoint crossing cycles
$F_1,\ldots,F_k$ of length at most $s$. Set $F_j^{(i)}=F_j\cap E_i$.
The two half-cycle bond products agree, since $\prod_{e\in F_j}B_e=1$;
denote their common value by $A_j$. The observed half-sign products are
\[
 X_j=A_j\prod_{e\in F_j^{(1)}}\nu_e,\qquad
 Y_j=A_j\prod_{e\in F_j^{(2)}}\nu_e.
\]
Write $Z_j=(X_j,Y_j)$, and put $b_j=\E[A_j\mid Z_{<j}]$ and
$r_j^{(i)}=\prod_{e\in F_j^{(i)}}\eta_e$. The noise products in this
display are independent of each other, $A_j$ and the earlier pairs.
The conditional pair law is therefore
\[
 \Pr(X_j=x,Y_j=y\mid Z_{<j})
 =\tfrac14\{1+b_jr_j^{(1)}x+b_jr_j^{(2)}y
                  +r_j^{(1)}r_j^{(2)}xy\}.
\]

Every nonempty proper edge subset of a simple cycle has an odd-degree
vertex $v$. Choose one in $F_j^{(1)}$. Then $A_j$ is $S_v$ times a
function of the other spins. Vertex-disjointness ensures that earlier
pairs depend only on those other spins and on independent emission
noise. Conditional expectation first over $S_v$, then over the other
spins and earlier noise, gives $|b_j|\leq\epsilon_\pi$. Hence
\[
 \operatorname{Cov}(X_j,Y_j\mid Z_{<j})
 =r_j^{(1)}r_j^{(2)}(1-b_j^2)
 \geq\kappa^s(1-\epsilon_\pi^2).
\]
Lemma~\ref{lem:conditional-affinity} proves the bound on
$\mathcal I_{\zeta_\alpha}$. Proposition~\ref{prop:balanced-affinity}
gives the circuit bounds, and Appendix~\ref{app:smoothing} removes
smoothness. Finally $\beta(t)\geq t^2/4$ gives the displayed explicit
lower bound.
\end{proof}

\subsection{Removing Smoothness}\label{app:smoothing}
We use standard smoothing \citep{shih2019} with reference densities;
\citet[Lemma~2]{broadrick2024} restate this step as a conversion from
likelihood polynomials to network polynomials.
In a finite decomposable nonnegative DAG with binary products,
internal scopes are unions of child scopes. If the root omits
coordinates, first multiply it by value-one singleton leaves on them.
At each sum--child wire, multiply the child by value-one singleton leaves
on the coordinates of the sum's scope that it misses. They integrate to
one under the reference probabilities.
A separate wrapper at each wire leaves shared children intact.
The padding scopes are disjoint, so binary products remain decomposable.
Processing sums bottom-up preserves the function and original scopes
and makes the DAG smooth. New singleton leaves respect the cap $\alpha d\geq1$.
The root and each of the $M_w$ wires add at most $2d$ nodes (at most $d$
singleton leaves and $d$ binary products), so $M'\leq M+2d(1+M_w)$.
Merging parallel sum wires gives $M_w\leq M(M-1)$, hence $M'\leq4dM^2$.
Applying~\eqref{eq:size} to $M'$ gives
$\log M\geq\tfrac12\{\Lambda+2\log(1-\delta)-\log(4d)\}$
at TV error $\delta<1$.

\subsection{Packing Crossing Cycles}
We first bound how many cycles a balanced cut must split. Adjacent
unsplit cycles have the same color. Thus two large sides of a cut need
enough split cycles to separate them across the array.
\begin{lemma}[Square arrays of cycles]\label{lem:face-array}
Let $T$ be the union of a $b\times b$ array of simple $f$-cycles.
Suppose adjacent indices in each row or column share an edge, indices
differing by at least two in either coordinate have disjoint vertex
sets, and each edge lies in at most two cycles. Let $n_\partial$ count
edges in exactly one cycle. If each side of a cut contains at least
$\zeta|T|$ edges, its maximum number $k$ of vertex-disjoint crossing
cycles satisfies
\[
 4k^2+4(1-1/f)k\geq\zeta b^2-(1-\zeta)n_\partial/f.
\]
\end{lemma}
\begin{proof}
Color the cut sides 1 and 2. Write $n_1,n_2$ for pure-cycle counts
and $m$ for the mixed-cycle count. If color $j$ occupies $e_j$ edges,
then $2e_j\leq fn_j+(f-1)m+n_\partial$: a mixed cycle uses at most
$f-1$ edges of either color, and boundary edges are counted only once.
Using $e_j\geq\zeta|T|$ and $2|T|=fb^2+n_\partial$ gives
$\zeta b^2-(1-\zeta)n_\partial/f\leq\min(n_1,n_2)+(1-1/f)m$.
Within each of the four index-parity classes, cycles are vertex-disjoint,
so $m\leq4k$.
Choosing one mixed cycle in each occupied row of a fixed parity gives
at most $k$ cycles; hence at most $2k$ rows contain mixed cycles.
The same holds for columns. If every row or every column contains a
mixed cycle, then $b\leq2k$, so $\min(n_1,n_2)\leq b^2\leq4k^2$.
Otherwise a mixed-free row and column are monochromatic, since
adjacent pure cycles share an edge. Their intersection forces the
same color, also on every other mixed-free row or column. Pure cycles
of the opposite color lie among at most $(2k)^2$ intersections of
mixed rows and columns. Thus $\min(n_1,n_2)\leq4k^2$ in all cases.
\end{proof}

\subsection{Mixed Grid Faces}
\begin{lemma}[Mixed grid faces]\label{lem:grid}
Let $L\geq2$. Every split $E_1\mid E_2$ of the edges of an $L\times L$
grid with at least $\zeta_\alpha d$ edges on each side has at least
$k_{L,\alpha}$ vertex-disjoint square faces with edges in both $E_1$ and
$E_2$, with $k_{L,\alpha}$ as in~\eqref{eq:count}.
In particular $k_{L,\alpha}\geq\sqrt{\zeta_\alpha}L/3$ for
$L\geq4/\zeta_\alpha$.
\end{lemma}
\begin{proof}
Apply Lemma~\ref{lem:face-array} with $b=L-1$, $f=4$ and
$n_\partial=4b$. Its right side is
$(L-1)(\zeta_\alpha L-1)$, giving $k\geq k_{L,\alpha}$.

For $L\geq4/\zeta_\alpha$ we have $L\geq12$, so
$\zeta_\alpha L-1\geq\tfrac34\zeta_\alpha L$ and $L-1\geq\tfrac{11}{12}L$,
hence $(L-1)(\zeta_\alpha L-1)\geq\tfrac{11}{16}\zeta_\alpha L^2$. With
$k_0=\sqrt{\zeta_\alpha}L/3$,
$4k_0^2+3k_0=\tfrac49\zeta_\alpha L^2+\sqrt{\zeta_\alpha}L
\leq\tfrac{11}{16}\zeta_\alpha L^2$, because
$\sqrt{\zeta_\alpha}L\geq4/\sqrt{\zeta_\alpha}\geq4\sqrt3>144/35$.
Every integer $k<k_0$ thus has $4k^2+3k<(L-1)(\zeta_\alpha L-1)$, so
$k_{L,\alpha}\geq k_0$.
\end{proof}

\begin{proof}(Corollary~\ref{cor:grid-circuit}).
A square face is a simple cycle of length four, so
Lemma~\ref{lem:grid} gives $k_{\alpha,4}(G)\geq k_{L,\alpha}$, and
Theorem~\ref{thm:witness-barrier} with $s=4$ gives the lower bounds with
$\Lambda_L$ in place of $\Lambda$. Lemma~\ref{lem:grid} and
$\beta(\kappa^4)\geq\kappa^8/4$ give
$\Lambda_L\geq\kappa^8\sqrt{\zeta_\alpha}L/12$ for $L\geq4/\zeta_\alpha$.

For the upper bound, we carry the $L$ spins of one row in $2^L$ states.
A state's value is the joint density of the processed readings and that
row assignment, with earlier spins summed out. Initialize the first row
by the factors $1+s_is_{i+1}h_e(u_e)$ of its horizontal edges, the first
one multiplied by $2^{-L}$. To advance one row,
replace the old spins $s_i$ by new spins $s'_i$ from left to right: with
$\mathbf g_0$ the current state vector and $e_i$ the $i$th vertical edge,
\[
 \mathbf g_i(s'_{\leq i},s_{>i})
 =\frac12\sum_{s_i=\pm1}
 \mathbf g_{i-1}(s'_{<i},s_i,s_{>i})\{1+s_is'_ih_{e_i}(u_{e_i})\}.
\]
The factor $1/2$ draws the new spin uniformly and the sum removes the old
one, preserving the invariant. After $L$ updates, process the new row's
horizontal edges; after the last row, sum all states. Each step multiplies
by nonnegative univariate leaves on a new coordinate, and both summands of
an update have the same scope. With $2^L$ states and two predecessors per
state, the circuit is smooth, decomposable and nonnegative, and
structured by the linear vtree of its processing order, with
$O(d2^L)$ nodes; its singleton leaves meet the scope cap since
$\alpha d\geq1$. Since $\Lambda_L=\Theta(L)$ for fixed $\kappa,\alpha$, the
exponential orders match.
\end{proof}

\begin{remark}[Finite constants]\label{rem:finite-constants}
For the ramp with $a=0.95$ and $\tau=1/4$ (so $\kappa=\eta=0.83125$),
$\alpha=2/3$ and $\delta=1/2$, the smooth-circuit bound of
Corollary~\ref{cor:grid-circuit} first gives $\log M>0$ at $L=80$ ($d=12{,}640$)
and first exceeds the trivial count $M\geq3$ (two leaves and a product)
at $L=139$ ($d=38{,}364$). Both exceed the fitted dimensions, so our
experiments do not test the circuit bound. The example satisfies the hypothesis of
Theorem~\ref{thm:separation} exactly when $R>2a/\tau=7.6$.
\end{remark}
\subsection{Average versus Global Error}\label{app:tile-proof}
For an integer $b\geq1$ let $q_b$ be the tiled law of
Section~\ref{sec:average-global}, with tiles of at most $b\times b$
vertices.

\begin{proposition}[Average versus global error]\label{prop:blocks}
For centered channels with $|h_e|\leq a<1$, $q_b$ has a smooth
decomposable nonnegative circuit with singleton leaves and $O(d2^b)$
nodes, and $\KL(p_h\Vert q_b)\leq\sum_{e\text{ deleted}}\xi_e\leq a^2d/b$.
Also $\KL(q_b\Vert p_h)\leq a^2d/\{2(1-a^2)b\}$.
If moreover $\eta_e\geq\kappa>0$ on every
edge and $2\leq b\leq L-1$, then
$\TV(p_h,q_b)\geq1-\exp\{-\beta(\kappa^4)(L-1)^2/(8b)\}$.
\end{proposition}

\begin{proof}
The construction deletes $2L\lfloor(L-1)/b\rfloor\leq d/b$ edges. The
tiles use disjoint vertex sets, so $q_b$ is the product of the tile laws
and uniform factors on the deleted coordinates. We apply the row-frontier
construction of Corollary~\ref{cor:grid-circuit} within each tile. Its
width is at most $b$, so it carries at most $2^b$ states. Multiplying
tile circuits and uniform singleton leaves gives a smooth decomposable
nonnegative circuit with $O(d2^b)$ nodes. We couple $p_h$ and $q_b$ to the same latent
spins. Only the deleted emissions differ, and each has
$\KL(1+B_eh_e\Vert1)\leq\int h_e^2=\xi_e\leq a^2$; adding these
conditional terms bounds the KL between the joint laws of spins and
readings, and marginalizing the spins (data processing) gives
$\KL(p_h\Vert q_b)\leq\sum_{e\text{ deleted}}\xi_e\leq a^2d/b$.
For reverse KL, let $P$ and $Q$ include the same independent uniform
latent spins. Under $Q$, replace only the deleted-edge emissions by
uniform readings. Its reading marginal is $q_b$. Each bond is a fair
sign under $Q$, and the unchanged emissions cancel in the density
ratio. Thus, with $\xi_e=\int h_e^2$,
\[
 \KL(q_b\Vert p_h)\leq\KL(Q\Vert P)
 =-\frac12\sum_{e\text{ deleted}}\int\log(1-h_e^2)
 \leq\frac{\sum_{e\text{ deleted}}\xi_e}{2(1-a^2)}
 \leq\frac{a^2d}{2(1-a^2)b}.
\]
Here $-\log(1-y)\leq y/(1-y)$ for $0\leq y<1$, and every emission
is bounded below by $1-a$.

For TV, let $2\leq b\leq L-1$. In each of the $\lfloor(L-1)/b\rfloor$
face rows between zero-based vertex rows $jb-1$ and $jb$,
$1\leq j\leq\lfloor(L-1)/b\rfloor$, take the faces in the even
face columns among $0,\dots,L-2$. This gives
$k=\lfloor(L-1)/b\rfloor\lceil(L-1)/2\rceil\geq(L-1)^2/(4b)$ faces
$C_1,\ldots,C_k$, using $\lfloor x\rfloor\geq x/2$ for $x\geq1$. They are
vertex-disjoint: alternate faces in one face row share no vertex, and
$b\geq2$ separates the face rows. Let $\phi_{C_j}$ be their thresholded
parities, as in the proof of Corollary~\ref{cor:grid-circuit}. Under $p_h$,
conditionally on the spins, $\phi_{C_j}$ has mean
$\theta_{C_j}\prod_{e\in C_j}B_e=\theta_{C_j}\geq\kappa^4$, independent of the
spin values. The parities use disjoint emissions, so they are independent
with these means. Under $q_b$, each $C_j$ contains two deleted vertical
edges whose readings are uniform and independent of all others, so the
$\phi_{C_j}$ are independent fair signs. Affinity does not decrease under
$u\mapsto(\phi_{C_1},\ldots,\phi_{C_k})$, and a sign with mean $\theta$
has affinity $(\sqrt{1+\theta}+\sqrt{1-\theta})/2=e^{-\beta(\theta)/2}$
with a fair sign. Since $\beta$ is increasing on $[0,1)$,
\[
 \Aff(p_h,q_b)\leq\prod_{j\leq k}e^{-\beta(\theta_{C_j})/2}
 \leq e^{-k\beta(\kappa^4)/2}
 \leq\exp\{-\beta(\kappa^4)(L-1)^2/(8b)\},
\]
and $1-\TV\leq\Aff$ gives the bound. By data processing,
$\TV(p_h,q_b)$ is also at least the TV between the two parity laws; when
every $\theta_{C_j}$ equals $\theta$, as for the ramp, the count of
positive parities is sufficient, so this is the TV between
$\mathrm{Bin}(k,(1+\theta)/2)$ and $\mathrm{Bin}(k,1/2)$, the certified
lower bound in Table~\ref{tab:tiling-gap}. At $b=\lfloor\log_2d\rfloor$, which lies in
$[2,L-1]$ for $L\geq6$, the exponent grows as $d/\log d$.
\end{proof}
\subsection{KL Approximation and Block Prediction}\label{app:kl-size}
We apply balanced products to disjoint coordinate blocks and add their KL
costs with the chain rule. These are prediction blocks, not the observable
channel blocks of Section~\ref{sec:witnesses}. Conditioning makes each
block's cost depend on the observed readings, so we keep that dependence
until the final expectation.

\begin{lemma}[Adding block costs]\label{lem:conditional-block-cost}\label{lem:kl-direct-sum}
Let $O,T_1,\ldots,T_m$ be disjoint coordinate sets, $|T_i|\geq2$,
and put $T_{<i}=\bigcup_{j<i}T_j$ and $O_i=O\cup T_{<i}$.
Let $q$ be a normalized smooth decomposable nonnegative circuit with $M$ nodes after binarizing products,
whose leaves meet each
$T_i$ in at most $2|T_i|/3$ coordinates. Write
$\mathcal A_i=\mathcal A(q,T_i)$ for its fixed cut family from
Lemma~\ref{lem:block-balance}. Suppose measurable $w_{i,A}(u_{O_i})\geq0$
satisfy, for every $A\in\mathcal A_i$, $p_{O_i}$-almost every history,
and all product densities
across $A\mid T_i\setminus A$,
\[
 \Aff(p_{T_i\mid u_{O_i}},f_A\otimes g_{T_i\setminus A})
 \leq e^{-w_{i,A}(u_{O_i})/2}.
\]
Then, for $S=E\setminus O$,
\begin{equation}\label{eq:random-block-cost}
 \mathcal R_S(p,q)\geq\sum_i\E_{p_{O_i}}
 \left(\left[\min_{A\in\mathcal A_i}w_{i,A}(U_{O_i})-\log M\right]_+\right).
\end{equation}
The cut minimum and positive part are taken before averaging.
\end{lemma}
The order matters: the evidence may make different cuts weak on
different samples. We must control the weakest cut for each history
before using the KL chain rule.
\begin{proof}
Fix $U_O$, marginalize to $\cup_iT_i$, and apply the KL chain rule.
Each conditional of $q$ mixes at most $M$ products across
$\mathcal A_i$ (Lemma~\ref{lem:block-balance}). Square-root
subadditivity bounds its affinity with the target conditional by
$\sqrt M\exp\{-\min_A w_{i,A}/2\}$. KL dominates zero and minus
twice the log affinity. Sum and average to obtain~\eqref{eq:random-block-cost}.
For independent target blocks, the target conditional is always
$p_{T_i}$. The same argument applies in either KL direction, with
histories averaged under its first argument. Undefined conditionals of
the second argument on positive first-argument mass give infinite risk.
All chain rules use nonnegative extended quantities.
\end{proof}

We use two consequences. If
$\mathcal I_{1/3}(p_{T_i\mid u_{O_i}})\geq\lambda_i$ almost surely for
constants $\lambda_i\geq0$, then the lower bound is
$\sum_i[\lambda_i-\log M]_+$. It holds for $p_O$-almost every fixed
$u_O$ when these hypotheses hold for almost every subsequent history
under $p_{T_{<i}\mid u_O}$. If the target blocks are independent,
$p_{\cup_iT_i}=\bigotimes_i p_{T_i}$, the same circuit assumptions give
$\KL(p\Vert q),\KL(q\Vert p)
\geq\sum_i[\mathcal I_{1/3}(p_{T_i})-\log M]_+$.
In each case, balance is relative to $|T_i|$.

\begin{proof}(Theorem~\ref{thm:kl-size}).
We pack $\lfloor L/b\rfloor^2$ vertex-disjoint $b\times b$ subgrids.
Integrating an unused reading removes its emission factor, since
$\int h_e=0$. The remaining subgrids use disjoint independent spins
and emission noises, so their reading marginals are independent.
For each subgrid, Theorem~\ref{thm:witness-barrier} and
Lemma~\ref{lem:grid}, with leaf fraction $2/3$ and balance $1/3$, give
$\mathcal I_{1/3}\geq b\beta(\kappa^4)/(3\sqrt3)$ when $b\geq12$.
Each block has $2b(b-1)$ readings, so the leaf condition in
Lemma~\ref{lem:kl-direct-sum} holds when $\ell\leq4b(b-1)/3$.
That lemma proves~\eqref{eq:kl-blocks} in each KL direction.

Put $c_0=\beta(\kappa^4)/(3\sqrt3)$ and choose
\[
 b=\max\left\{12,\left\lceil\frac{2r\log d}{c_0}\right\rceil,
 \left\lceil\frac{1+\sqrt{1+3\ell}}2\right\rceil\right\}.
\]
The last term ensures the leaf condition. Since $\ell\leq\alpha d$,
it is at most $2+\sqrt{3\alpha/2}\,L$. Thus $b\leq L$ for all
sufficiently large $L$, uniformly in $\ell$, because $\alpha<2/3$.
The size budget gives $\log M\leq r\log d$, so the bracket in
\eqref{eq:kl-blocks} is at least $c_0b/2$. Using
$\lfloor L/b\rfloor\geq L/(2b)$ and $d\leq2L^2$ gives the lower bound
$c_0d/(16b)$. Finally $b\asymp\max\{\log d,\sqrt\ell\}$ uniformly
over the stated range.

For the upper bound take $t=\lfloor(r-1)\log_2d/2\rfloor$.
Proposition~\ref{prop:blocks} gives $q_t$ with
$O(d2^t)=O(d^{(r+1)/2})$ nodes and both KL divergences $O(d/t)$.
Here $t\asymp\log d$ with $1\leq t\leq L$ for large $d$.
Alternatively, choose the largest integer $t\geq1$ with
$2t(t-1)\leq\ell$. Every tile, including boundary tiles, has at most
$\ell$ internal edges. Multiply one exact-density leaf per nonempty tile
and one uniform leaf per deleted reading. Their scopes partition $E$,
so this normalized, decomposable circuit is smooth and has $O(d)$ nodes.
Since $t\asymp\sqrt\ell$, Proposition~\ref{prop:blocks} gives both KL
bounds $O(d/\sqrt\ell)$. Both constructions fit $\lfloor d^r\rfloor$
for large $d$; choosing the larger width proves~\eqref{eq:kl-rate}.
\end{proof}

\begin{lemma}[Faces across an aligned matching]\label{lem:matching-faces}
In a $2b\times2b$ vertex square, observe the aligned horizontal
matching and let $T$ be the hidden edges. If $b\geq8$, every
$1/3$-balanced cut of $T$ splits the hidden edges of at least $b/2$
elementary faces. At least $\lceil b/12\rceil$ such faces touch disjoint
observed matching components.
\end{lemma}
\begin{proof}
Index a face by its top-left vertex $(i,c)$, from zero. It has two
hidden verticals and, when $c$ is odd, two hidden horizontals.
Adjacent faces share a hidden edge along every row and along odd
columns. A row or odd column without a mixed face is therefore
monochromatic. There are $2b-1$ rows and $b-1$ odd columns.
If each row or each odd column contains a mixed face, the claim follows.
Otherwise choose a pure row and a pure odd column. They intersect,
so have the same color, say 1. Every pure row and pure odd column
has that color. If there are $m$ mixed faces, a pure color-2 face in
an odd column must lie in one of at most $m$ mixed rows and $m$ mixed
odd columns. There are at most $m^2$ such faces.
Every hidden edge belongs to an odd-column face except the $4b-2$
vertical edges on the left and right boundaries. Thus color 2 has at
most $4(m^2+m)+4b-2$ edges. Since $|T|=6b^2-4b$, balance gives
\[
4m^2+4m\geq2b^2-16b/3+2.
\]
If $m<b/2$, this implies $b^2-22b/3+2<0$, impossible for $b\geq8$.
Finally, faces in one class $(i\bmod2,c\bmod3)$ touch disjoint matching
components: their row pairs or their pair-column sets
$\{\lfloor c/2\rfloor,\lfloor(c+1)/2\rfloor\}$ are disjoint.
One of the six classes contains at least $\lceil m/6\rceil$ mixed faces.
\end{proof}

\begin{proof}(Theorem~\ref{thm:block-prediction}).
Each $b$ below is at least $12$ and, since $\beta\leq\log2$,
at least $2\log d$ for large $d$. Each block has at least $2b(b-1)$
coordinates, so $\ell\leq(\log d)^2$ gives the leaf condition of
Lemma~\ref{lem:conditional-block-cost}.

\emph{Patch.} Put $m=\lfloor\rho L\rfloor$ and choose
$b=\lceil6\sqrt3\,r\log d/\beta(\kappa^4)\rceil$.
The inner $(m-2)\times(m-2)$ vertex square has no incident observed edge.
Pack it with $\lfloor(m-2)/b\rfloor^2$ vertex-disjoint $b\times b$
subgrids. Their reading laws are independent of one another and of
$U_O$, since integrating unused readings removes their emissions.
Lemma~\ref{lem:conditional-block-cost} gives the lower bound from
Theorem~\ref{thm:kl-size}, with the block count using $m-2$ in place of $L$.

\emph{Staggered mask.} Every vertex in columns $1,\ldots,L-2$,
including the top and bottom rows, has exactly one observed incident edge. The observed components are horizontal dimers and two-edge boundary
paths; hence $O$ is a forest. Counting its edges gives $|O|=L(L+1)/2$ and
$|S|=L(3L-5)/2$. Each vertex touches $O$.

We use hidden cycles of length six around two vertically adjacent cells.
Index them by $(t,c)$, where $0\leq t\leq L/2-2$ and
$1\leq c\leq L-3$, with top-left vertex
$(2t+((c+1)\bmod2),c)$. The top-row index plus the column index is odd, so both horizontal
edges are hidden. The vertical edges avoid the observed boundary columns. Consecutive indices in a row or column share an edge;
indices differing by at least two in either coordinate give disjoint
vertex sets. Each horizontal edge occurs in at most two cycles with
the same $c$. A vertical edge in column $x$ occurs in at most one cycle
with $c=x-1$ and one with $c=x$. Thus each edge lies in at most two
cycles. A square $b\times b$ array has $n_\partial=8b-2$ edges in exactly
one cycle: there are $b(b-1)$ shared horizontal edges and
$(b-1)(2b-1)$ shared vertical edges, including those between diagonally
adjacent indices. Thus
$6b^2-2\{b(b-1)+(b-1)(2b-1)\}=8b-2$.
Lemma~\ref{lem:face-array} at $f=6$, $\zeta=1/3$ gives
$4k^2+10k/3\geq b^2/3-(8b-2)/9$, and therefore $k\geq b/6$
for $b\geq12$.

One unused index between successive arrays in both directions gives
$\lfloor L/\{2(b+1)\}\rfloor\lfloor(L-2)/(b+1)\rfloor$
arrays. This is $\Theta(L^2/b^2)$ when $b+1\leq L/4$.
Let $T_i$ be their edge sets. The arrays are vertex-disjoint. Fix the readings on
$O_i=O\cup\bigcup_{j<i}T_j$. The posterior spin law is
\[
 \pi_{O_i}(s\mid u_{O_i})\ \propto\
 \prod_{e\in O_i}(1+B_eh_e(u_e))
 =\frac{\exp\{\sum_{e\in O_i}\artanh(h_e(u_e))B_e\}}
 {\prod_{e\in O_i}\cosh(\artanh(h_e(u_e)))}.
\]
At a vertex of $T_i$, exactly one incident edge belongs to $O_i$;
earlier arrays touch no such vertex. Its conditional spin mean has
absolute value at most $a$. The tower property preserves this bound
after marginalizing spins outside the vertex set of $T_i$, even when
the observed edge joins a vertex outside that set. The conditional target law
on $T_i$ thus satisfies Theorem~\ref{thm:dependent-barrier} with
$\epsilon_\pi\leq a$ and $s=6$. Uniformly in the history,
\[
 \mathcal I_{1/3}(p_{h,T_i\mid u_{O_i}})
 \geq (b/6)\beta((1-a^2)\kappa^6).
\]
Write $\beta_*=\beta((1-a^2)\kappa^6)>0$ and choose
$b=\lceil12r\log d/\beta_*\rceil$. Lemma~\ref{lem:conditional-block-cost}
gives $\Omega(L^2\beta_*/b)=\Omega(d/\log d)$ for circuits of size
at most $d^r$, for almost every evidence value.

\emph{Aligned matching.} Conditioned on $U_O=u_O$, latent spins
factor over matched pairs $o=\{v,w\}$ with mass
$(1+s_vs_wh_o(u_o))/4$. Put
$b=\lceil384\log2\,(r\log d+\log2)/\kappa^8\rceil$. Place
$\lfloor L/(2b)\rfloor^2$ disjoint $2b\times2b$ squares at offsets in
$(2b\mathbb Z)^2$, and let
$T_i$ be their induced hidden edges. Each observed pair lies wholly
inside one square or outside all squares. Integrating unused hidden
readings removes their emission factors, so the $T_i$ are independent
given $U_O$. Earlier target blocks do not change their conditional laws.

Fix $A\in\mathcal A_i$, a $1/3$-balanced cut of $T_i$, and choose
$k=\lceil b/12\rceil$ mixed faces from Lemma~\ref{lem:matching-faces},
independently of the evidence. On each side, take the sign product of
the hidden readings of a face. A face with four hidden edges touches
four distinct matching components. Its spins are independent fair
signs, so the two products are fair with covariance $c_C=\prod_{e\in C}\eta_e$.
A face with hidden verticals $e_1,e_2$ and observed horizontals $o_1,o_2$
gives fair products with covariance
\begin{equation}\label{eq:matching-covariance}
 c_C(u_O)=\eta_{e_1}\eta_{e_2}h_{o_1}(u_{o_1})h_{o_2}(u_{o_2}).
\end{equation}
Faces touching disjoint matching components give independent pairs
conditionally on $U_O$. Applying the fixed-cut argument in
Lemma~\ref{lem:conditional-affinity} to $|c_C|$ gives the measurable
cost $W_A(U_O)=\sum_C\beta(c_C(U_O))$, since $\beta$ is even.

Since $O$ is a forest, its readings are independent uniform variables
under $p_{h,O}$. The summands in
$W_A$ are independent, lie in $[0,\log2]$, and have means at least
$\kappa^8/4$. For a four-hidden face this follows from
$\beta(\kappa^4)\geq\kappa^8/4$; for a two-hidden face use
$\xi_o\geq\eta_o^2\geq\kappa^2$ in~\eqref{eq:matching-covariance}.
Put $\lambda=k\kappa^8/8\geq b\kappa^8/96$. For independent
$X_j\in[0,B]$, convexity gives
$\E e^{-X_j/B}\leq\exp\{-(1-e^{-1})\E X_j/B\}$.
Markov's inequality, $\E W_A\geq2\lambda$ and
$2(1-e^{-1})-1>1/4$ give
$\Pr\{W_A<\lambda\}\leq e^{-\lambda/(4\log2)}$.
There are at most $M$ cuts in $\mathcal A_i$, fixed before the evidence.
Our choice of $b$ gives $\lambda\geq4\log2\log(2M)\geq2\log M$ when $M\leq d^r$.
A union bound gives $\min_{A\in\mathcal A_i}W_A\geq\lambda$ with
probability at least $1/2$. Equation~\eqref{eq:random-block-cost}
contributes at least $\lambda/4\geq b\kappa^8/384$ per block.
Summing gives $\Omega(d/\log d)$ for sufficiently large $L$.

\emph{Upper bounds.} The univariate-leaf class is contained in
$\mathcal C_{M,\ell}$. For all three masks, its positive tiled law $q_t$ satisfies
$\mathcal R_S(p_h,q_t)\leq\KL(p_h\Vert q_t)=O(d/\log d)$.
Each mask has $|S|=\Theta(d)$; for the aligned matching,
$|S|=L(3L-4)/2$. For both forest masks, $p_{h,O}=q_{t,O}=1$, so
\begin{equation}\label{eq:forest-loss}
 \log\frac{p_h(u_S\mid u_O)}{q_t(u_S\mid u_O)}
 =\log\frac{p_h(u)}{q_t(u)},\qquad
 \mathcal R_S(p_h,q_t)=\KL(p_h\Vert q_t).
\end{equation}
Deleting edges preserves the forest property, proving the identity.
\end{proof}

For nonsmooth circuits, we apply the argument to the smoothed
circuit of size $M'\leq4dM^2$ (Appendix~\ref{app:smoothing}). The same
rates hold with $\log M\leq r\log d$ replaced by
$\log M'\leq(2r+1)\log d+\log4$, including in the choice of $b$.
\FloatBarrier
\section{Implementation and Experimental Protocol}\label{sec:implementation}\label{app:submission-protocol}
Our bounds use exact projections and arithmetic
(Theorem~\ref{thm:learning}, Proposition~\ref{prop:queries});
we implement projections and determinant queries in floating point.
We therefore check projection errors and compare query results with
independent calculations.
Spline evidence integrals need no quadrature (Appendix~\ref{app:model-details});
equal interval endpoints specify point densities.

\noindent\emph{Numerical projection.}
Suppose feasible block-$A$ coefficients differ from the exact estimator
(Appendix~\ref{app:learning-proofs}) by at most $\varepsilon_A$ in weighted
nodal norm, including preliminary corner projection errors, with cap
$a_{\mathrm{in}}\in[a,1)$. Nodal norm equivalence,
$(x+y)^2\leq2x^2+2y^2$ and~\eqref{eq:emission-kl} give, after enlarging $C$,
for either directed KL divergence $D$ between $p_h$ and $\widetilde p$,
\[
 \E D
 \leq 2C d(K/N+K^{-2}+N^{-1})
       +\frac{1}{1-a_{\mathrm{in}}^2}\sum_A\E\varepsilon_A^2 .
\]
We use 60 scalar bisection steps and Dykstra iterations for corners.
We check feasibility and a primal--dual gap: for objective
$\tfrac12\|x-v\|_\omega^2$ (using $\omega\otimes\omega$ for corners),
a feasible point's squared projection error is at most twice the gap.
We propagate preliminary corner gaps and reject unmet tolerances.
These bounds use no interval enclosures and give no iteration or
bit-complexity bound.

\phantomsection\noindent\emph{Numerical queries.}\label{app:numerical-diagnostics}
Above cap $1/3$, the Neumann bound may be inconclusive. Phase and
posterior-range guards detect some failures but do not certify forward
error. We compare determinants and bond solves with positive transfer
on forest, disjoint-square and strip cases through amplitude $0.9999$
(Table~\ref{tab:submission-numerical}); Section~\ref{sec:geometry-check}
compares sparse LU orders.
\begin{table}[htbp]\centering\small
\begin{tabular}{llrrrrrr}\toprule
 & & & & \multicolumn{2}{c}{Returned / rejected} & \multicolumn{2}{c}{Maximum error}\\
\cmidrule(lr){5-6}\cmidrule(l){7-8}
Graph & $a$ & $d$ & Jobs & Determinant & Bonds & Log & Bond \\\midrule
Forest & 0.25 & 112--32{,}512 & 6/6 & 12/0 & 12/0 & $0$ & $1.7\times10^{-18}$ \\
Forest & 0.8 & 112--32{,}512 & 6/6 & 12/0 & 12/0 & $4.4\times10^{-13}$ & $1.7\times10^{-16}$ \\
Forest & 0.95 & 112--32{,}512 & 6/6 & 12/0 & 12/0 & $4.5\times10^{-13}$ & $0$ \\
Forest & 0.9999 & 112--32{,}512 & 6/6 & 12/0 & 12/0 & $6.5\times10^{-13}$ & $2.2\times10^{-16}$ \\
\midrule
Disjoint squares & 0.25 & 112--32{,}512 & 6/6 & 12/0 & 12/0 & $2.3\times10^{-13}$ & $1.7\times10^{-18}$ \\
Disjoint squares & 0.8 & 112--32{,}512 & 6/6 & 12/0 & 12/0 & $1.0\times10^{-12}$ & $1.7\times10^{-16}$ \\
Disjoint squares & 0.95 & 112--32{,}512 & 6/6 & 12/0 & 12/0 & $1.8\times10^{-12}$ & $4.4\times10^{-16}$ \\
Disjoint squares & 0.9999 & 112--32{,}512 & 6/6 & 12/0 & 12/0 & $2.5\times10^{-9}$ & $6.1\times10^{-13}$ \\
\midrule
Full strip & 0.25 & 220--14{,}332 & 6/6 & 12/0 & 12/0 & $5.5\times10^{-14}$ & $6.9\times10^{-16}$ \\
Full strip & 0.8 & 220--14{,}332 & 6/6 & 12/0 & 12/0 & $6.3\times10^{-13}$ & $3.2\times10^{-14}$ \\
Full strip & 0.95 & 220--14{,}332 & 6/6 & 12/0 & 12/0 & $2.3\times10^{-12}$ & $2.6\times10^{-13}$ \\
Full strip & 0.9999 & 220--14{,}332 & 6/6 & 10/2 & 12/0 & $2.7\times10^{-8}$ & $1.9\times10^{-3}$ \\
\bottomrule\end{tabular}
\caption{Determinant contractions on fixed numerical cases with two mask rates and point and interval evidence. The amplitude $a$ sets the conditioning regime. The Jobs column gives completed/scheduled runs. Determinant and bond columns count returned/rejected queries; each bond query checks three selected bonds by sparse solves. Rejections are retained guard outcomes. Errors are maximum absolute discrepancies from analytic or positive-transfer references; bond errors include every returned solve, even when the determinant is rejected. These cases do not extend the sufficient conditioning regime.}\label{tab:submission-numerical}
\end{table}
For the 4 prespecified large-strip queries, 80/160-digit positive transfer agrees within $2.0\times10^{-76}$; the largest float64 transfer discrepancy is $1.4\times10^{-11}$. The maximum discrepancy among 3 returned determinants is $4.6\times10^{-9}$. This is a precision comparison, not a directed-rounding certificate.

Across \SubmissionValue{moment-query-fits} moment fits, we check
\SubmissionValue{moment-query-checks} point, missing-coordinate
and interval queries on four test vectors per fit, using positive
transfer with shared block integrals. The maximum log-likelihood discrepancy is
$\SubmissionValue{moment-query-max-error}$. For the
\SubmissionValue{quadrature-fits} loop-target fits, doubling quadrature order
changes block integrated squared error (ISE) by at most
$\SubmissionValue{quadrature-max-error}$.

Of \SubmissionSupport{numerical-queries} fixed queries, the determinant
phase guard rejects \SubmissionSupport{numerical-rejected} full-strip
points at $a=0.9999$ in dimensions
\SubmissionSupport{numerical-rejected-dim-1} and
\SubmissionSupport{numerical-rejected-dim-2}. Bond solves return estimates
with maximum errors
$\SubmissionSupport{numerical-rejected-bond-error-1}$ and
$\SubmissionSupport{numerical-rejected-bond-error-2}$, respectively.
For the \SubmissionSupport{numerical-accepted} accepted determinants,
maximum log and corresponding bond errors are
$\SubmissionSupport{numerical-accepted-log-error}$ and
$\SubmissionSupport{numerical-accepted-bond-error}$.

\noindent\emph{Scoring.}
We use known square grids, known uniform margins and independent
vectors. The held-out mean of $\log(p/\widehat p)(U)$ estimates joint KL
in nats per vector; we retain negative estimates. Seed intervals include
fitting and evaluation variation; Monte Carlo errors concern a fixed fit.

\noindent\emph{Matched comparison.}
Our six tasks combine $L=4,8$ ($d=24,112$) with ramp, smooth and Gaussian
targets. Loop channels have $a=0.95$ and shapes
$h=a\ramp_{1/4}$ and $h=a\tanh((2u-1)/\tau')$, with
$\tau'\log\cosh(1/\tau')=0.875$; both have $\eta=0.83125$.
Selection minimizes mean validation negative log likelihood over two valid
development seeds, breaking ties by list order. Ten confirmation seeds share
$32768/4096/4096$ training/validation/test vectors from independent random streams.
Gaussian-based models include the probit Jacobian.

\begin{table}[htbp]\centering\footnotesize
\setlength{\tabcolsep}{4pt}
\begin{tabular}{@{}lp{0.82\linewidth}@{}}\toprule
Moments & Observable scalar and corner blocks, $K=8$, input anchor
$0.3$, cap $0.98$; all training vectors serve short-cycle calibration and regression.\\
Likelihood, EM & Centered scalar splines, $K=8$, cap $0.98$, no anchor
constraint; validation selects the checkpoint and a cold or moment start.
Likelihood uses projected Adam \citep{kingma2015adam}: rate $0.01$,
batch 256, validation every 32 updates, patience 12, at most 10,000 updates.
Generalized EM \citep{dempster1977} uses full E steps and up to 20
backtracking projected-gradient M-step updates; validation after each
iteration, patience six.\\
HCLT & HCLTs \citep{liu2021hclt} in PyJuice \citep{liu2024pyjuice},
with Gaussian leaves in probit coordinates;
minimum standard deviation $0.05$; structure
learned from the first 4096 training vectors; widths $32,64$, full-batch
EM or Anemone \citep{liu2026learning}. Batches of 256; Anemone accumulates 4096 vectors per update.
Validation each epoch, patience 15, at most 10,000 epochs.\\
GMM & Full covariances in probit coordinates; $8,16$ components for
loop targets at $d=24$, otherwise $1,2$. Covariance regularization $10^{-5}$, tolerance $10^{-4}$, one initialization,
at most 200 iterations.\\
MPC & Linear-tree region graph over the midpoint order, Gaussian probit leaves; widths $8,16$ for loop
targets and $16,32$ for Gaussian targets. Adam at rate $0.01$.\\
OSOS & Same region graph, unit-period complex Fourier leaves,
landing-Adam optimizer (landing methods: \citealp{ablin2022landing,ablin2024infeasible})
at rates $0.01,0.05$;
regularization $0.1$, safe-step parameter $0.5$; widths $16,32$ for loop
targets and $8,16$ for Gaussian targets. Validation projects weights
by SVD and scores the actual Gram normalizer.\\
\bottomrule\end{tabular}
\caption{Procedures and development candidates for the matched comparison.
Circuit width is the number of sum units; MPC has the same number of
input units and OSOS has one more. Both use batches of 128, validation
every 128 updates and at most 8192 updates.}
\label{tab:submission-grid}
\end{table}

MPC and OSOS use \citet{loconte2026orthogonal}'s code with our region
graph and Fourier leaves.\footnote{\url{https://github.com/april-tools/squared-unitary-pcs}.}
Moment starts use the corner control: leading weighted singular components
and centered box projection. Cold starts use
$0.2(2u-1)$ plus independent $N(0,0.02^2)$ noise, then the same projection.
We score loop fits by positive row transfer, exponential in grid width.
The optimizers have no global-optimum guarantee.

\noindent\emph{Budgets and inference.}
Caps of 180 seconds in development and 600 seconds otherwise include
initialization, validation and checkpointing. We keep the best valid
checkpoint, excluding failed starts. Likelihood and EM use two starts
per task and seed; others use one. Budgets and capacities differ.
Two single-thread CPU workers and one GPU worker use an Intel Core Ultra 9
285K and NVIDIA RTX 5090 under a 72-hour limit.

The \SubmissionValue{primary-planned} primary contrasts are raw moment joint KL minus
validation-selected likelihood, EM, HCLT, GMM, MPC, OSOS or independence,
on each of six tasks. Each requires ten complete pairs. Intervals are
\[
 \bar D\ \pm\ t_{9,\,1-0.05/(2\cdot\SubmissionValue{primary-planned})}\,s_D/\sqrt{10},
\]
where $\bar D,s_D$ are the mean and standard deviation of ten paired
seed differences, assumed approximately normal. Positive intervals favor
the comparator. Secondary intervals are unadjusted 95\% with $n-1$ degrees
of freedom for $n\geq2$ pairs unless stated otherwise; sample counts are reported.

\section{Supporting Experimental Results}\label{app:protocols}
We give the settings and results behind Section~\ref{sec:experiments},
including conditional prediction, resolution choices and weak-channel fits.
\subsection{Matched Fits and Paired Comparisons}\label{app:matched-protocol}
Selected loop-task widths for HCLT, MPC and OSOS are
\SubmissionValue{loop-gpu-width}, \SubmissionValue{loop-mpc-width} and
\SubmissionValue{loop-osos-width}. GMM component counts (ramp, smooth)
are $(\SubmissionValue{t0-gmm-components},\SubmissionValue{t1-gmm-components})$
at $d=24$ and $(\SubmissionValue{t3-gmm-components},\SubmissionValue{t4-gmm-components})$
at $d=112$. Of
\SubmissionValue{confirmation-trajectories} confirmation trajectories,
\SubmissionValue{confirmation-capped} reach the cap with valid
checkpoints; likelihood and EM each use two starts.

\noindent\emph{Conditional prediction.}
For observed $O$, the mean of
$\log(p/\widehat p)(U)-\log(p_O/\widehat p_O)(U_O)$ estimates
conditional KL under $p_O$.
One fixed permutation gives nested 25\% and 50\% masks. The patch hides
edge midpoints below $L/2$ in both coordinates (eight edges at $d=24$,
32 at $d=112$); the forest observes horizontal edges. Masks are shared
across seeds and methods; fits and test sets are shared across masks.
The 192 task--method--mask cells report ten seed means and descriptive,
unadjusted 95\% intervals, retaining negative values.
Independent baseline checks cover all masks and coordinate permutations
in four dimensions, including changed missing values. Full-dimensional
baseline scores lack an independent reference; fitted block scores are
checked against positive transfer.

\begin{table}[htbp]\centering\small
\begin{tabular}{@{}llrrr@{\hspace{12pt}}rrrr@{\hspace{12pt}}r@{}}\toprule
 & & \multicolumn{3}{c}{Noisy-loop fits} & \multicolumn{4}{c}{Baselines} & Reference\\
\cmidrule(lr){3-5}\cmidrule(lr){6-9}\cmidrule(l){10-10}
$d$ & Target & Mom. & Lik. & EM & HCLT & GMM & MPC & OSOS & Indep. \\\midrule
24 & Ramp & $0.017$ & $0.010$ & $\mathbf{0.003}$ & $1.946$ & $1.913$ & $1.984$ & $2.806$ & $1.971$ \\
24 & Smooth & $0.039$ & $0.031$ & $\mathbf{0.022}$ & $1.639$ & $1.736$ & $1.797$ & $2.615$ & $1.785$ \\
24 & Gaussian & $3.324$ & $3.323$ & $3.324$ & $0.033$ & $\mathbf{0.004}$ & $0.097$ & $2.242$ & $3.324$ \\
\midrule
112 & Ramp & $0.116$ & $0.073$ & $\mathbf{0.014}$ & $11.862$ & $11.918$ & $11.881$ & $13.786$ & $11.820$ \\
112 & Smooth & $0.254$ & $0.173$ & $\mathbf{0.124}$ & $10.728$ & $10.768$ & $10.731$ & $12.606$ & $10.668$ \\
112 & Gaussian & $15.964$ & $15.963$ & $15.963$ & $0.168$ & $\mathbf{0.101}$ & $0.493$ & $10.267$ & $15.963$ \\
\bottomrule\end{tabular}
\caption{Joint KL in nats per vector, averaged over ten seeds. Methods share training, validation and test observations. At $d=112$, GMM candidates have one or two components, and all \SubmissionValue{osos-large-loop-capped} selected OSOS fits on the loop targets reach their time cap. Bold marks the smallest displayed mean in complete rows, including ties; it does not indicate statistical significance. Table~\ref{tab:submission-contrasts} gives pair counts and adjusted comparisons.}\label{tab:matched-joint}
\end{table}
\begingroup\footnotesize\setlength{\tabcolsep}{2pt}
\setlength{\LTcapwidth}{\textwidth}
\begin{longtable}{@{}lccc@{\hspace{8pt}}ccc@{}}
\multicolumn{7}{l}{\tablename~\thetable}\\[3pt]\toprule & \multicolumn{3}{c}{$d=24$} & \multicolumn{3}{c}{$d=112$}\\\cmidrule(lr){2-4}\cmidrule(l){5-7}Method & Ramp & Smooth & Gaussian & Ramp & Smooth & Gaussian \\\midrule\endfirsthead
\multicolumn{7}{l}{\tablename~\thetable\ (continued)}\\[3pt]\toprule & \multicolumn{3}{c}{$d=24$} & \multicolumn{3}{c}{$d=112$}\\\cmidrule(lr){2-4}\cmidrule(l){5-7}Method & Ramp & Smooth & Gaussian & Ramp & Smooth & Gaussian \\\midrule\endhead
\midrule\endfoot
\bottomrule
\noalign{\vskip\abovecaptionskip}
\caption{Conditional KL in nats per vector: mean and unadjusted 95\% Student interval over ten seeds. These 192 secondary estimates are descriptive; negative estimates and interval endpoints are retained. Bold marks the smallest displayed mean within each task and mask, including ties, without a significance claim.}\label{tab:conditional}\\\endlastfoot
\multicolumn{7}{l}{\textit{Fixed random 25\% of coordinates hidden}}\label{tab:conditional-random25}\\*[4pt]
Moments & \shortstack{$0.013$\\{\footnotesize$[0.011, 0.016]$}} & \shortstack{$0.029$\\{\footnotesize$[0.026, 0.031]$}} & \shortstack{$1.238$\\{\footnotesize$[1.217, 1.259]$}} & \shortstack{$0.085$\\{\footnotesize$[0.081, 0.090]$}} & \shortstack{$0.182$\\{\footnotesize$[0.176, 0.188]$}} & \shortstack{$6.712$\\{\footnotesize$[6.677, 6.748]$}} \\*[1pt]
Likelihood & \shortstack{$0.008$\\{\footnotesize$[0.007, 0.009]$}} & \shortstack{$0.023$\\{\footnotesize$[0.020, 0.025]$}} & \shortstack{$1.237$\\{\footnotesize$[1.216, 1.259]$}} & \shortstack{$0.053$\\{\footnotesize$[0.050, 0.057]$}} & \shortstack{$0.117$\\{\footnotesize$[0.112, 0.122]$}} & \shortstack{$6.712$\\{\footnotesize$[6.676, 6.747]$}} \\*[1pt]
EM & \shortstack{$\mathbf{0.003}$\\{\footnotesize$[0.002, 0.004]$}} & \shortstack{$\mathbf{0.016}$\\{\footnotesize$[0.014, 0.017]$}} & \shortstack{$1.238$\\{\footnotesize$[1.217, 1.259]$}} & \shortstack{$\mathbf{0.010}$\\{\footnotesize$[0.009, 0.012]$}} & \shortstack{$\mathbf{0.080}$\\{\footnotesize$[0.077, 0.084]$}} & \shortstack{$6.712$\\{\footnotesize$[6.676, 6.747]$}} \\*[1pt]
HCLT & \shortstack{$1.439$\\{\footnotesize$[1.397, 1.481]$}} & \shortstack{$1.172$\\{\footnotesize$[1.074, 1.270]$}} & \shortstack{$0.013$\\{\footnotesize$[0.011, 0.014]$}} & \shortstack{$7.952$\\{\footnotesize$[7.918, 7.986]$}} & \shortstack{$7.218$\\{\footnotesize$[7.179, 7.257]$}} & \shortstack{$0.074$\\{\footnotesize$[0.069, 0.079]$}} \\*[1pt]
GMM & \shortstack{$1.395$\\{\footnotesize$[1.377, 1.413]$}} & \shortstack{$1.266$\\{\footnotesize$[1.245, 1.286]$}} & \shortstack{$\mathbf{0.001}$\\{\footnotesize$[0.001, 0.002]$}} & \shortstack{$7.974$\\{\footnotesize$[7.953, 7.996]$}} & \shortstack{$7.233$\\{\footnotesize$[7.194, 7.272]$}} & \shortstack{$\mathbf{0.044}$\\{\footnotesize$[0.042, 0.047]$}} \\*[1pt]
MPC & \shortstack{$1.461$\\{\footnotesize$[1.445, 1.476]$}} & \shortstack{$1.324$\\{\footnotesize$[1.306, 1.343]$}} & \shortstack{$0.033$\\{\footnotesize$[0.030, 0.037]$}} & \shortstack{$7.949$\\{\footnotesize$[7.928, 7.971]$}} & \shortstack{$7.204$\\{\footnotesize$[7.163, 7.244]$}} & \shortstack{$0.187$\\{\footnotesize$[0.179, 0.196]$}} \\*[1pt]
OSOS & \shortstack{$2.272$\\{\footnotesize$[2.245, 2.299]$}} & \shortstack{$2.130$\\{\footnotesize$[2.110, 2.150]$}} & \shortstack{$1.485$\\{\footnotesize$[1.453, 1.516]$}} & \shortstack{$9.764$\\{\footnotesize$[9.732, 9.795]$}} & \shortstack{$8.994$\\{\footnotesize$[8.944, 9.044]$}} & \shortstack{$6.230$\\{\footnotesize$[6.161, 6.299]$}} \\*[1pt]
Independence & \shortstack{$1.458$\\{\footnotesize$[1.442, 1.473]$}} & \shortstack{$1.321$\\{\footnotesize$[1.303, 1.339]$}} & \shortstack{$1.238$\\{\footnotesize$[1.216, 1.259]$}} & \shortstack{$7.931$\\{\footnotesize$[7.908, 7.954]$}} & \shortstack{$7.189$\\{\footnotesize$[7.148, 7.229]$}} & \shortstack{$6.712$\\{\footnotesize$[6.676, 6.747]$}} \\[4pt]
\multicolumn{7}{l}{\textit{Fixed random 50\% of coordinates hidden}}\label{tab:conditional-random50}\\*[4pt]
Moments & \shortstack{$0.016$\\{\footnotesize$[0.013, 0.018]$}} & \shortstack{$0.036$\\{\footnotesize$[0.033, 0.039]$}} & \shortstack{$2.521$\\{\footnotesize$[2.495, 2.548]$}} & \shortstack{$0.113$\\{\footnotesize$[0.106, 0.119]$}} & \shortstack{$0.247$\\{\footnotesize$[0.240, 0.254]$}} & \shortstack{$11.663$\\{\footnotesize$[11.606, 11.719]$}} \\*[1pt]
Likelihood & \shortstack{$0.010$\\{\footnotesize$[0.009, 0.011]$}} & \shortstack{$0.028$\\{\footnotesize$[0.025, 0.030]$}} & \shortstack{$2.520$\\{\footnotesize$[2.494, 2.547]$}} & \shortstack{$0.071$\\{\footnotesize$[0.069, 0.074]$}} & \shortstack{$0.166$\\{\footnotesize$[0.159, 0.174]$}} & \shortstack{$11.662$\\{\footnotesize$[11.605, 11.718]$}} \\*[1pt]
EM & \shortstack{$\mathbf{0.003}$\\{\footnotesize$[0.002, 0.004]$}} & \shortstack{$\mathbf{0.020}$\\{\footnotesize$[0.018, 0.022]$}} & \shortstack{$2.521$\\{\footnotesize$[2.495, 2.548]$}} & \shortstack{$\mathbf{0.014}$\\{\footnotesize$[0.012, 0.016]$}} & \shortstack{$\mathbf{0.119}$\\{\footnotesize$[0.113, 0.125]$}} & \shortstack{$11.662$\\{\footnotesize$[11.605, 11.718]$}} \\*[1pt]
HCLT & \shortstack{$1.755$\\{\footnotesize$[1.703, 1.807]$}} & \shortstack{$1.472$\\{\footnotesize$[1.368, 1.575]$}} & \shortstack{$0.024$\\{\footnotesize$[0.022, 0.026]$}} & \shortstack{$11.366$\\{\footnotesize$[11.322, 11.411]$}} & \shortstack{$10.282$\\{\footnotesize$[10.231, 10.332]$}} & \shortstack{$0.122$\\{\footnotesize$[0.117, 0.127]$}} \\*[1pt]
GMM & \shortstack{$1.740$\\{\footnotesize$[1.723, 1.757]$}} & \shortstack{$1.573$\\{\footnotesize$[1.554, 1.593]$}} & \shortstack{$\mathbf{0.003}$\\{\footnotesize$[0.002, 0.004]$}} & \shortstack{$11.419$\\{\footnotesize$[11.385, 11.452]$}} & \shortstack{$10.310$\\{\footnotesize$[10.260, 10.359]$}} & \shortstack{$\mathbf{0.074}$\\{\footnotesize$[0.071, 0.078]$}} \\*[1pt]
MPC & \shortstack{$1.793$\\{\footnotesize$[1.775, 1.810]$}} & \shortstack{$1.624$\\{\footnotesize$[1.606, 1.642]$}} & \shortstack{$0.065$\\{\footnotesize$[0.062, 0.069]$}} & \shortstack{$11.379$\\{\footnotesize$[11.345, 11.414]$}} & \shortstack{$10.268$\\{\footnotesize$[10.217, 10.319]$}} & \shortstack{$0.328$\\{\footnotesize$[0.317, 0.339]$}} \\*[1pt]
OSOS & \shortstack{$2.615$\\{\footnotesize$[2.587, 2.644]$}} & \shortstack{$2.441$\\{\footnotesize$[2.420, 2.462]$}} & \shortstack{$1.981$\\{\footnotesize$[1.943, 2.019]$}} & \shortstack{$13.252$\\{\footnotesize$[13.205, 13.299]$}} & \shortstack{$12.114$\\{\footnotesize$[12.059, 12.169]$}} & \shortstack{$8.477$\\{\footnotesize$[8.396, 8.558]$}} \\*[1pt]
Independence & \shortstack{$1.786$\\{\footnotesize$[1.768, 1.804]$}} & \shortstack{$1.618$\\{\footnotesize$[1.601, 1.636]$}} & \shortstack{$2.521$\\{\footnotesize$[2.494, 2.548]$}} & \shortstack{$11.346$\\{\footnotesize$[11.310, 11.382]$}} & \shortstack{$10.236$\\{\footnotesize$[10.185, 10.287]$}} & \shortstack{$11.662$\\{\footnotesize$[11.605, 11.718]$}} \\[4pt]
\multicolumn{7}{l}{\textit{Missing patch}}\label{tab:conditional-patch}\\*[4pt]
Moments & \shortstack{$0.010$\\{\footnotesize$[0.008, 0.012]$}} & \shortstack{$0.022$\\{\footnotesize$[0.019, 0.024]$}} & \shortstack{$1.557$\\{\footnotesize$[1.535, 1.579]$}} & \shortstack{$0.044$\\{\footnotesize$[0.041, 0.046]$}} & \shortstack{$0.100$\\{\footnotesize$[0.096, 0.104]$}} & \shortstack{$5.614$\\{\footnotesize$[5.571, 5.657]$}} \\*[1pt]
Likelihood & \shortstack{$0.006$\\{\footnotesize$[0.004, 0.007]$}} & \shortstack{$0.016$\\{\footnotesize$[0.014, 0.017]$}} & \shortstack{$1.557$\\{\footnotesize$[1.535, 1.578]$}} & \shortstack{$0.028$\\{\footnotesize$[0.027, 0.030]$}} & \shortstack{$0.066$\\{\footnotesize$[0.062, 0.070]$}} & \shortstack{$5.614$\\{\footnotesize$[5.571, 5.657]$}} \\*[1pt]
EM & \shortstack{$\mathbf{0.002}$\\{\footnotesize$[0.001, 0.003]$}} & \shortstack{$\mathbf{0.011}$\\{\footnotesize$[0.009, 0.012]$}} & \shortstack{$1.557$\\{\footnotesize$[1.535, 1.579]$}} & \shortstack{$\mathbf{0.005}$\\{\footnotesize$[0.003, 0.006]$}} & \shortstack{$\mathbf{0.046}$\\{\footnotesize$[0.042, 0.049]$}} & \shortstack{$5.614$\\{\footnotesize$[5.571, 5.657]$}} \\*[1pt]
HCLT & \shortstack{$0.956$\\{\footnotesize$[0.915, 0.998]$}} & \shortstack{$0.835$\\{\footnotesize$[0.784, 0.886]$}} & \shortstack{$0.015$\\{\footnotesize$[0.014, 0.017]$}} & \shortstack{$4.266$\\{\footnotesize$[4.231, 4.301]$}} & \shortstack{$3.844$\\{\footnotesize$[3.814, 3.873]$}} & \shortstack{$0.062$\\{\footnotesize$[0.057, 0.067]$}} \\*[1pt]
GMM & \shortstack{$1.000$\\{\footnotesize$[0.987, 1.014]$}} & \shortstack{$0.880$\\{\footnotesize$[0.871, 0.889]$}} & \shortstack{$\mathbf{0.003}$\\{\footnotesize$[0.002, 0.004]$}} & \shortstack{$4.275$\\{\footnotesize$[4.248, 4.301]$}} & \shortstack{$3.861$\\{\footnotesize$[3.833, 3.889]$}} & \shortstack{$\mathbf{0.049}$\\{\footnotesize$[0.046, 0.053]$}} \\*[1pt]
MPC & \shortstack{$0.982$\\{\footnotesize$[0.972, 0.992]$}} & \shortstack{$0.879$\\{\footnotesize$[0.873, 0.885]$}} & \shortstack{$0.040$\\{\footnotesize$[0.037, 0.043]$}} & \shortstack{$4.245$\\{\footnotesize$[4.219, 4.271]$}} & \shortstack{$3.830$\\{\footnotesize$[3.802, 3.858]$}} & \shortstack{$0.171$\\{\footnotesize$[0.160, 0.181]$}} \\*[1pt]
OSOS & \shortstack{$1.673$\\{\footnotesize$[1.652, 1.693]$}} & \shortstack{$1.573$\\{\footnotesize$[1.558, 1.588]$}} & \shortstack{$1.346$\\{\footnotesize$[1.326, 1.366]$}} & \shortstack{$5.468$\\{\footnotesize$[5.433, 5.504]$}} & \shortstack{$5.036$\\{\footnotesize$[5.002, 5.070]$}} & \shortstack{$4.459$\\{\footnotesize$[4.391, 4.526]$}} \\*[1pt]
Independence & \shortstack{$0.977$\\{\footnotesize$[0.968, 0.987]$}} & \shortstack{$0.874$\\{\footnotesize$[0.868, 0.880]$}} & \shortstack{$1.557$\\{\footnotesize$[1.535, 1.579]$}} & \shortstack{$4.228$\\{\footnotesize$[4.201, 4.255]$}} & \shortstack{$3.813$\\{\footnotesize$[3.784, 3.843]$}} & \shortstack{$5.614$\\{\footnotesize$[5.571, 5.657]$}} \\[4pt]
\multicolumn{7}{l}{\textit{Observed forest}}\label{tab:conditional-forest}\\*[4pt]
Moments & \shortstack{$0.017$\\{\footnotesize$[0.014, 0.019]$}} & \shortstack{$0.039$\\{\footnotesize$[0.036, 0.042]$}} & \shortstack{$2.162$\\{\footnotesize$[2.132, 2.192]$}} & \shortstack{$0.116$\\{\footnotesize$[0.109, 0.123]$}} & \shortstack{$0.254$\\{\footnotesize$[0.248, 0.261]$}} & \shortstack{$9.082$\\{\footnotesize$[9.030, 9.133]$}} \\*[1pt]
Likelihood & \shortstack{$0.010$\\{\footnotesize$[0.009, 0.011]$}} & \shortstack{$0.031$\\{\footnotesize$[0.028, 0.033]$}} & \shortstack{$2.161$\\{\footnotesize$[2.131, 2.191]$}} & \shortstack{$0.073$\\{\footnotesize$[0.070, 0.076]$}} & \shortstack{$0.173$\\{\footnotesize$[0.165, 0.180]$}} & \shortstack{$9.081$\\{\footnotesize$[9.029, 9.132]$}} \\*[1pt]
EM & \shortstack{$\mathbf{0.003}$\\{\footnotesize$[0.002, 0.004]$}} & \shortstack{$\mathbf{0.022}$\\{\footnotesize$[0.020, 0.024]$}} & \shortstack{$2.162$\\{\footnotesize$[2.131, 2.192]$}} & \shortstack{$\mathbf{0.014}$\\{\footnotesize$[0.013, 0.016]$}} & \shortstack{$\mathbf{0.124}$\\{\footnotesize$[0.118, 0.130]$}} & \shortstack{$9.081$\\{\footnotesize$[9.029, 9.132]$}} \\*[1pt]
HCLT & \shortstack{$1.939$\\{\footnotesize$[1.887, 1.991]$}} & \shortstack{$1.628$\\{\footnotesize$[1.521, 1.734]$}} & \shortstack{$0.024$\\{\footnotesize$[0.022, 0.026]$}} & \shortstack{$11.841$\\{\footnotesize$[11.796, 11.885]$}} & \shortstack{$10.715$\\{\footnotesize$[10.663, 10.767]$}} & \shortstack{$0.097$\\{\footnotesize$[0.091, 0.102]$}} \\*[1pt]
GMM & \shortstack{$1.898$\\{\footnotesize$[1.880, 1.916]$}} & \shortstack{$1.727$\\{\footnotesize$[1.710, 1.743]$}} & \shortstack{$\mathbf{0.003}$\\{\footnotesize$[0.002, 0.004]$}} & \shortstack{$11.892$\\{\footnotesize$[11.853, 11.932]$}} & \shortstack{$10.742$\\{\footnotesize$[10.691, 10.793]$}} & \shortstack{$\mathbf{0.077}$\\{\footnotesize$[0.073, 0.080]$}} \\*[1pt]
MPC & \shortstack{$1.979$\\{\footnotesize$[1.959, 1.998]$}} & \shortstack{$1.791$\\{\footnotesize$[1.775, 1.808]$}} & \shortstack{$0.064$\\{\footnotesize$[0.060, 0.067]$}} & \shortstack{$11.850$\\{\footnotesize$[11.813, 11.888]$}} & \shortstack{$10.701$\\{\footnotesize$[10.648, 10.754]$}} & \shortstack{$0.277$\\{\footnotesize$[0.264, 0.289]$}} \\*[1pt]
OSOS & \shortstack{$2.756$\\{\footnotesize$[2.732, 2.781]$}} & \shortstack{$2.564$\\{\footnotesize$[2.543, 2.586]$}} & \shortstack{$1.832$\\{\footnotesize$[1.798, 1.865]$}} & \shortstack{$13.576$\\{\footnotesize$[13.528, 13.625]$}} & \shortstack{$12.402$\\{\footnotesize$[12.342, 12.462]$}} & \shortstack{$7.168$\\{\footnotesize$[7.085, 7.252]$}} \\*[1pt]
Independence & \shortstack{$1.971$\\{\footnotesize$[1.951, 1.991]$}} & \shortstack{$1.785$\\{\footnotesize$[1.769, 1.801]$}} & \shortstack{$2.162$\\{\footnotesize$[2.131, 2.192]$}} & \shortstack{$11.820$\\{\footnotesize$[11.781, 11.859]$}} & \shortstack{$10.668$\\{\footnotesize$[10.616, 10.721]$}} & \shortstack{$9.081$\\{\footnotesize$[9.029, 9.132]$}} \\[4pt]
\end{longtable}\endgroup
\begin{table}[htbp]\centering\footnotesize\setlength{\tabcolsep}{4pt}
\begin{tabular}{@{}lcccc@{}}\toprule
Task & Likelihood & EM & HCLT & GMM \\\midrule
Ramp, $d=24$ & \shortstack{$0.00639$\\{\footnotesize$[0.00184, 0.01094]$}} & \shortstack{$0.01352$\\{\footnotesize$[0.00945, 0.01759]$}} & \shortstack{$-1.92878$\\{\footnotesize$[-2.02849, -1.82907]$}} & \shortstack{$-1.89660$\\{\footnotesize$[-1.93364, -1.85956]$}} \\[2pt]
Smooth, $d=24$ & \shortstack{$0.00871$\\{\footnotesize$[0.00362, 0.01379]$}} & \shortstack{$0.01732$\\{\footnotesize$[0.01416, 0.02047]$}} & \shortstack{$-1.59993$\\{\footnotesize$[-1.80944, -1.39041]$}} & \shortstack{$-1.69655$\\{\footnotesize$[-1.73104, -1.66206]$}} \\[2pt]
Gaussian, $d=24$ & \shortstack{$0.00096$\\{\footnotesize$[-0.00027, 0.00218]$}} & \shortstack{$0.00010$\\{\footnotesize$[-0.00032, 0.00051]$}} & \shortstack{$3.29079$\\{\footnotesize$[3.21565, 3.36593]$}} & \shortstack{$3.31944$\\{\footnotesize$[3.24318, 3.39569]$}} \\[2pt]
\midrule
Ramp, $d=112$ & \shortstack{$0.04314$\\{\footnotesize$[0.02997, 0.05631]$}} & \shortstack{$0.10157$\\{\footnotesize$[0.08894, 0.11420]$}} & \shortstack{$-11.74613$\\{\footnotesize$[-11.83042, -11.66183]$}} & \shortstack{$-11.80156$\\{\footnotesize$[-11.87463, -11.72849]$}} \\[2pt]
Smooth, $d=112$ & \shortstack{$0.08173$\\{\footnotesize$[0.06573, 0.09773]$}} & \shortstack{$0.13050$\\{\footnotesize$[0.11558, 0.14541]$}} & \shortstack{$-10.47365$\\{\footnotesize$[-10.58372, -10.36357]$}} & \shortstack{$-10.51351$\\{\footnotesize$[-10.61890, -10.40812]$}} \\[2pt]
Gaussian, $d=112$ & \shortstack{$0.00093$\\{\footnotesize$[-0.00022, 0.00207]$}} & \shortstack{$0.00093$\\{\footnotesize$[-0.00005, 0.00192]$}} & \shortstack{$15.79608$\\{\footnotesize$[15.66795, 15.92421]$}} & \shortstack{$15.86266$\\{\footnotesize$[15.72906, 15.99626]$}} \\[2pt]
\bottomrule\end{tabular}\par\medskip
\begin{tabular}{@{}lccc@{}}\toprule
Task & MPC & OSOS & Independence \\\midrule
Ramp, $d=24$ & \shortstack{$-1.96739$\\{\footnotesize$[-2.00758, -1.92721]$}} & \shortstack{$-2.78963$\\{\footnotesize$[-2.84613, -2.73313]$}} & \shortstack{$-1.95410$\\{\footnotesize$[-1.99450, -1.91371]$}} \\[2pt]
Smooth, $d=24$ & \shortstack{$-1.75785$\\{\footnotesize$[-1.79319, -1.72251]$}} & \shortstack{$-2.57585$\\{\footnotesize$[-2.61914, -2.53255]$}} & \shortstack{$-1.74589$\\{\footnotesize$[-1.78033, -1.71145]$}} \\[2pt]
Gaussian, $d=24$ & \shortstack{$3.22738$\\{\footnotesize$[3.14921, 3.30556]$}} & \shortstack{$1.08191$\\{\footnotesize$[0.95900, 1.20483]$}} & \shortstack{$0.00013$\\{\footnotesize$[-0.00028, 0.00053]$}} \\[2pt]
\midrule
Ramp, $d=112$ & \shortstack{$-11.76523$\\{\footnotesize$[-11.84039, -11.69008]$}} & \shortstack{$-13.66979$\\{\footnotesize$[-13.77320, -13.56638]$}} & \shortstack{$-11.70383$\\{\footnotesize$[-11.77899, -11.62866]$}} \\[2pt]
Smooth, $d=112$ & \shortstack{$-10.47645$\\{\footnotesize$[-10.58808, -10.36482]$}} & \shortstack{$-12.35166$\\{\footnotesize$[-12.46758, -12.23573]$}} & \shortstack{$-10.41408$\\{\footnotesize$[-10.52441, -10.30375]$}} \\[2pt]
Gaussian, $d=112$ & \shortstack{$15.47027$\\{\footnotesize$[15.34731, 15.59323]$}} & \shortstack{$5.69665$\\{\footnotesize$[5.50346, 5.88985]$}} & \shortstack{$0.00091$\\{\footnotesize$[-0.00005, 0.00188]$}} \\[2pt]
\bottomrule\end{tabular}\par\medskip
\caption{Moment-minus-comparator joint KL: mean and two-sided paired Student interval, with Bonferroni adjustment over all 42 contrasts. Positive values favor the comparator. Each reported interval uses ten pairs.}\label{tab:submission-contrasts}
\end{table}
\begin{table}[htbp]\centering\footnotesize\setlength{\tabcolsep}{4pt}
\begin{tabular}{@{}lcccc@{}}\toprule
Task & Mom. $-$ corner & Mom. $-$ shrunk & Corner $-$ lik. & Corner $-$ EM \\\midrule
Ramp, $d=24$ & \shortstack{$0.0006$\\{\footnotesize$[0.0004, 0.0009]$}} & \shortstack{$0.0000$\\{\footnotesize$[0.0000, 0.0000]$}} & \shortstack{$0.0058$\\{\footnotesize$[0.0036, 0.0079]$}} & \shortstack{$0.0129$\\{\footnotesize$[0.0110, 0.0148]$}} \\[2pt]
Smooth, $d=24$ & \shortstack{$0.0006$\\{\footnotesize$[0.0003, 0.0008]$}} & \shortstack{$0.0000$\\{\footnotesize$[0.0000, 0.0000]$}} & \shortstack{$0.0081$\\{\footnotesize$[0.0057, 0.0105]$}} & \shortstack{$0.0167$\\{\footnotesize$[0.0153, 0.0181]$}} \\[2pt]
Gaussian, $d=24$ & \shortstack{$0.0000$\\{\footnotesize$[-0.000002, 0.0001]$}} & \shortstack{$0.0001$\\{\footnotesize$[-0.00002, 0.0003]$}} & \shortstack{$0.0009$\\{\footnotesize$[0.0003, 0.0015]$}} & \shortstack{$0.0000$\\{\footnotesize$[-0.0002, 0.0003]$}} \\[2pt]
\midrule
Ramp, $d=112$ & \shortstack{$0.0007$\\{\footnotesize$[0.0002, 0.0011]$}} & \shortstack{$0.0000$\\{\footnotesize$[0.0000, 0.0000]$}} & \shortstack{$0.0425$\\{\footnotesize$[0.0362, 0.0488]$}} & \shortstack{$0.1009$\\{\footnotesize$[0.0948, 0.1070]$}} \\[2pt]
Smooth, $d=112$ & \shortstack{$0.0007$\\{\footnotesize$[0.0003, 0.0011]$}} & \shortstack{$0.0000$\\{\footnotesize$[0.0000, 0.0000]$}} & \shortstack{$0.0810$\\{\footnotesize$[0.0734, 0.0886]$}} & \shortstack{$0.1298$\\{\footnotesize$[0.1227, 0.1369]$}} \\[2pt]
Gaussian, $d=112$ & \shortstack{$0.0000$\\{\footnotesize$[-0.000004, 0.0001]$}} & \shortstack{$0.0009$\\{\footnotesize$[0.0004, 0.0013]$}} & \shortstack{$0.0009$\\{\footnotesize$[0.0003, 0.0014]$}} & \shortstack{$0.0009$\\{\footnotesize$[0.0004, 0.0014]$}} \\[2pt]
\bottomrule\end{tabular}
\caption{Secondary paired joint-KL differences: mean, unadjusted 95\% Student interval. Pair counts are ten unless shown. Positive values favor the second method. Corner projection changes only the four corner tensors; shrinkage is selected by validation.}\label{tab:submission-controls}
\end{table}

We did not test parameter sharing \citep{bricout2026patching} or
initialization and tree selection guided by latent-state utilization
\citep{shao2026utilization}.

\noindent\emph{Gaussian forest floor.}
For horizontal-edge positions $i_1<\cdots<i_m$ in the Gaussian target's
midpoint order, the marginal is Markov with successive correlations
$0.5^{i_{j+1}-i_j}$. Hence
$\det\Sigma_F=\prod_{j<m}(1-0.5^{2(i_{j+1}-i_j)})$.
Data processing gives joint KL to any uniform-forest law of at least
$-\tfrac12\log\det\Sigma_F$.
At $L=4,8$, the forest sizes
are \SubmissionValue{L4-gaussian-forest-size} and
\SubmissionValue{L8-gaussian-forest-size} and the bounds are
$\SubmissionValue{L4-gaussian-forest-floor}$ and
$\SubmissionValue{L8-gaussian-forest-floor}$ nats. These bounds do not establish
independence as the optimal loop-family approximation.

\subsection{Geometry, Query Scaling and Rates}\label{app:rate-diagnostics}
\noindent\emph{Observable geometry and query scaling.}
We use 16-cell splines with ramp amplitude \DiagnosticValue{geometry-amplitude},
width $1/2$ and reliability \DiagnosticValue{geometry-kappa}.
A centered even bump preserves reliability (Section~\ref{sec:geometry-check});
corner rescalings $0.8$ and $1/0.8$ preserve their product. Each direction uses
\DiagnosticValue{geometry-samples} exact draws at $L=4,6,8,12,16,24,32$.
Determinant scores use the fitted mean-zero control variate $p'/p-1$;
spline Gram matrices give $\Delta^2$. At $L=4$, a Walsh transform
enumerates $2^{24}$ signs for the sharpness path and single-channel control.

Query scaling uses one ramp draw with $a=0.95$, $\tau=1/16$ at
$L=8,16,\ldots,\DiagnosticValue{query-max-side}$, checked at $L=6$.
One-thread LU uses COLAMD or nested dissection. We count
$\sum_k\ell_ku_k$ multiplications from lower-column and upper-row
nonzero counts and fit slopes on the three largest grids.
Nested dissection uses pivot threshold zero and retains diagonal pivots in every run.

\noindent\emph{Rate diagnostics.}
Both shapes and dimensions use ten seeds, anchor $0.3$,
$N=8192$, $32768$, $131072$, $524288$. We use $K=8$ or
$K=8\lceil N^{1/3}/8\rceil\in\{24,32,56,88\}$. Knot grids contain
the ramp breakpoints. Each seed shares 65,536 test vectors across sizes
and resolutions. Gaussian quadrature, split at cells and ramp kinks,
gives block ISE; we double its order to check accuracy.
\begin{figure}[htbp]\centering
\includegraphics[width=\linewidth]{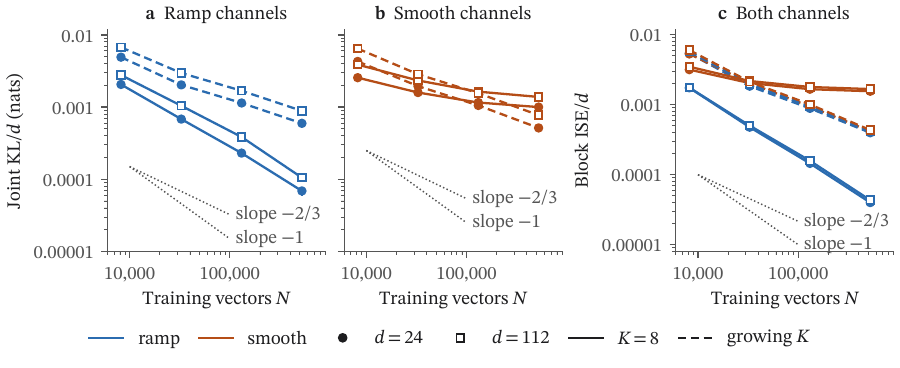}
\caption{Resolution and sample size: joint KL$/d$ (nats) for
ramp (a) and smooth (b) channels, and block ISE$/d$ (c).
Solid curves use $K=8$; dashed curves use $K=8\lceil N^{1/3}/8\rceil$.
Dotted slopes are $-1$ and $-2/3$; bars are 95\% Student intervals over
\SubmissionValue{rate-seeds} seeds. In (c), dimensions nearly coincide;
smooth lies above ramp at fixed $K$ and nearly overlaps it at growing $K$
(blue dashed under orange dashed). These curves do not identify an asymptotic exponent.}
\label{fig:submission-rates}
\end{figure}

Summed block ISE is the fitted-block version of $\Delta^2$ (Section~\ref{sec:global-geometry}).
To examine the slower-than-$N^{-2/3}$ joint-KL decay, we refit ramp models
over \DiagnosticValue{rate-diagnostic-seeds} seeds.
At small $N$, emission KL grows less than quadratically with channel
error, suggesting an effect of clipping on flat regions. True calibration
scales change the slope by at most
$\DiagnosticValue{rate-oracle-slope-change}$.
We did not refit the smooth target.

\subsection{Tiling and Missing-Patch Completion}\label{app:masked-tiling}
\noindent\emph{Tiling.} At $a=0.95$, $L=6,8,12,16,24,32$, we draw 10,000
exact ramp samples per grid by uniform spins and inverse CDFs.
Kac--Ward determinants score $p_h$ and each tile of $q_b$;
positive transfer checks \TilingCheckDraws{} draws per grid for
$L\leq\TilingTransferMaxSide$, with maximum log-density difference
$\TilingTransferMaxErr$. At $L=32$, dense--sparse LU differences are at most
$\TilingDenseMaxErr$ and 2-norm condition numbers at most $\TilingCondMax$.
KL and TV average $\log p_h-\log q_b$ and $(1-q_b/p_h)_+$
(Table~\ref{tab:tiling-gap}). TV estimates near one are not certified:
10,000 draws cannot exclude overlap on a region of $p_h$-probability
below about $3\times10^{-4}$.

\noindent\emph{Missing patch.}
For the ramp with $a=0.95$, $\tau=1/4$ and $L\in\{8,12,16,24,32\}$,
we hide the induced edges $S$ of the centered $\lfloor L/2\rfloor$-side
vertex square. We score 10,000 draws per grid under the uniform law and
known-channel tiled laws, $b\in\{2,4,\lfloor\log_2d\rfloor\}$, without
fitting. Seed namespace 20261116 gives one stream per grid.
We estimate $\mathcal R_S$ by averaging
$\log p_h(U)-\log p_{h,O}(U_O)-\log q_b(U)+\log q_{b,O}(U_O)$,
where $O=S^c$, with descriptive standard errors over independent draws
shared across tile widths.

We choose fully hidden, vertex-disjoint faces containing a deleted edge,
greedily in row-major order. Their sign parities are independent with
mean $\theta=\eta^4$ under $p_h$, independently of evidence, since each
cycle cancels its spins. Under $q_b$ each contains a distinct independent
uniform deleted reading, so the parities remain independent fair signs
after conditioning. Data processing and the joint upper bound give
\[
 k\left\{\frac{1+\theta}{2}\log(1+\theta)
       +\frac{1-\theta}{2}\log(1-\theta)\right\}
 \leq \mathcal R_S(p_h,q_b)\leq\xi d_{\mathrm{cut}},
 \qquad \xi=a^2(1-2\tau/3),
\]
where $k$ counts selected faces. These bounds concern this tiled law.
Tiled determinants zero deleted weights; evidence determinants also zero
hidden weights. On eight draws per grid, full and observed scores must
match per-tile determinants within $10^{-9}$ nats, with chain-rule residual
below $10^{-10}$. Positive transfer checks small instances. A deviation
beyond four standard errors from the bounds stops the study.
\begin{table}[htbp]\centering\small
\begin{tabular}{@{}rrrrrrrr@{}}\toprule
 & & & \multicolumn{3}{c}{Tiled risk $\mathcal R_S(p_h,q_b)$} & Per hidden edge & Uniform\\
\cmidrule(lr){4-6}\cmidrule(lr){7-7}\cmidrule(l){8-8}
$L$ & $|S|$ & $b$ & Lower & Estimate (SE) & Upper & $\mathcal R_S/|S|$ & $\mathcal R_S(p_h,\mu)$\\\midrule
8 & 24 & 2 & 0.23 & 5.17 (0.03) & 36.10 & 0.2156 & 5.90 (0.03)\\
 &  & 4 & 0.23 & 3.00 (0.02) & 12.04 & 0.1249 & \\
 &  & 6 & 0.00 & 2.12 (0.02) & 12.04 & 0.0883 & \\
\midrule
12 & 60 & 2 & 1.06 & 10.85 (0.04) & 90.25 & 0.1809 & 13.00 (0.05)\\
 &  & 4 & 0.95 & 8.30 (0.03) & 36.10 & 0.1383 & \\
 &  & 8 & 0.59 & 4.77 (0.02) & 18.05 & 0.0795 & \\
\midrule
16 & 112 & 2 & 1.42 & 19.48 (0.05) & 168.47 & 0.1739 & 22.37 (0.06)\\
 &  & 4 & 0.71 & 14.56 (0.04) & 72.20 & 0.1300 & \\
 &  & 8 & 0.71 & 6.64 (0.03) & 24.07 & 0.0593 & \\
\midrule
24 & 264 & 2 & 3.56 & 41.33 (0.08) & 397.10 & 0.1566 & 47.84 (0.09)\\
 &  & 4 & 2.85 & 26.11 (0.06) & 180.50 & 0.0989 & \\
 &  & 10 & 1.18 & 11.20 (0.04) & 72.20 & 0.0424 & \\
\midrule
32 & 480 & 2 & 6.65 & 70.25 (0.10) & 722.00 & 0.1463 & 81.82 (0.12)\\
 &  & 4 & 4.27 & 50.48 (0.08) & 336.94 & 0.1052 & \\
 &  & 10 & 3.08 & 25.61 (0.05) & 144.40 & 0.0534 & \\
\bottomrule\end{tabular}
\caption{Joint completion of a centered half-side patch, with known ramp channels $a=0.95$, $\tau=1/4$. Each grid uses 10,000 independent draws, paired across tile widths and the uniform law $\mu$. Parentheses give Monte Carlo standard errors. Grid size and the uniform reference appear once per group. The lower bound uses hidden face parities; the upper bound is the deleted-edge energy (Section~\ref{app:masked-tiling}). Bounds are rounded outwards. These finite sizes assess the tiled construction, not the asymptotic circuit lower bound.}\label{tab:masked-tiling}
\end{table}
\begin{table}[htbp]\centering\small
\begin{tabular}{@{}rrrrrrrr@{}}
\toprule
 & & & \multicolumn{3}{c}{KL$/d$} & \multicolumn{2}{c@{}}{TV}\\
\cmidrule(lr){4-6}\cmidrule(l){7-8}
$L$ & $d$ & $b$ & $a^2/b$ & $\xi d_{\mathrm{cut}}/d$ & estimate & lower bound & estimate\\
\midrule
6 & 60 & 5 & 0.180 & 0.150 & 0.0363 & 0.330 & 0.7303\\
8 & 112 & 6 & 0.150 & 0.107 & 0.0342 & 0.406 & 0.8587\\
12 & 264 & 8 & 0.113 & 0.068 & 0.0266 & 0.468 & 0.9549\\
16 & 480 & 8 & 0.113 & 0.050 & 0.0213 & 0.506 & 0.9835\\
24 & 1,104 & 10 & 0.090 & 0.065 & 0.0285 & 0.775 & $\approx1$\\
32 & 1,984 & 10 & 0.090 & 0.073 & 0.0315 & 0.915 & $\approx1$\\
\bottomrule
\end{tabular}
\caption{Tiled law $q_b$ with $b=\lfloor\log_2d\rfloor$ for the ramp with $a=0.95$, $\tau=1/4$. KL$/d$ bounds: $a^2/b$ and the sharper $\xi d_{\mathrm{cut}}/d$, with $d_{\mathrm{cut}}$ cut edges and $\xi=\int_0^1h^2=a^2(1-2\tau/3)$. Estimates are Monte Carlo from 10,000 draws of $p_h$ per row, with standard errors at most 0.0004 for KL$/d$ and 0.0035 for TV. TV lower bound: exact binomial TV of the face-parity witness count (proof of Proposition~\ref{prop:blocks}), rounded down; estimates near one are not certified.}\label{tab:tiling-gap}
\end{table}

\begin{figure}[htbp]\centering
\includegraphics[width=\linewidth]{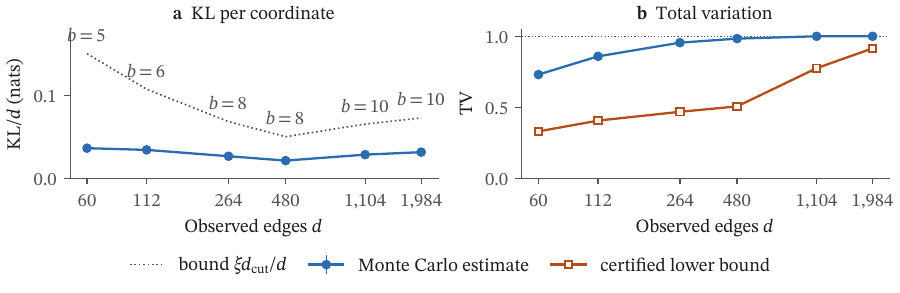}
\caption{Average versus global error for the tiled ramp law $q_b$,
$b=\lfloor\log_2d\rfloor$, $a=0.95$.
(a) Monte Carlo KL$/d$ and the bound
$\sum_{e\text{ deleted}}\xi_e/d=\xi d_{\mathrm{cut}}/d$ of
Proposition~\ref{prop:blocks}, where $d_{\mathrm{cut}}$ counts deleted
edges, $\xi=\int h^2$, and labels give $b$. Both rise from $d=1{,}104$,
where tiling needs multiple cuts per direction.
(b) Monte Carlo TV and the certified face-parity bound (exact binomial TV).}
\label{fig:tiling-gap}
\end{figure}

\subsection{Weak Signals and Weak Anchors}
\noindent\emph{Weak signals.}
At $d=112$, we use both loop shapes, reliabilities $0.28,0.375,0.5,0.83125$,
$N=8192,32768$, five seeds, $K=8$ and input anchor $0.2$.
We select $\lambda\in\{0,0.25,0.5,0.75,1\}$ on 4,096 independent
validation vectors, scaling block $A$ by $\lambda^{|A|}$.
Each seed has 65,536 independent test vectors. Separate $2^{20}$-vector
streams estimate independence loss, bounded by
$\log L_G(\xi)$ in~\eqref{eq:loop-energy}.
\begin{figure}[htbp]\centering
\includegraphics[width=\linewidth]{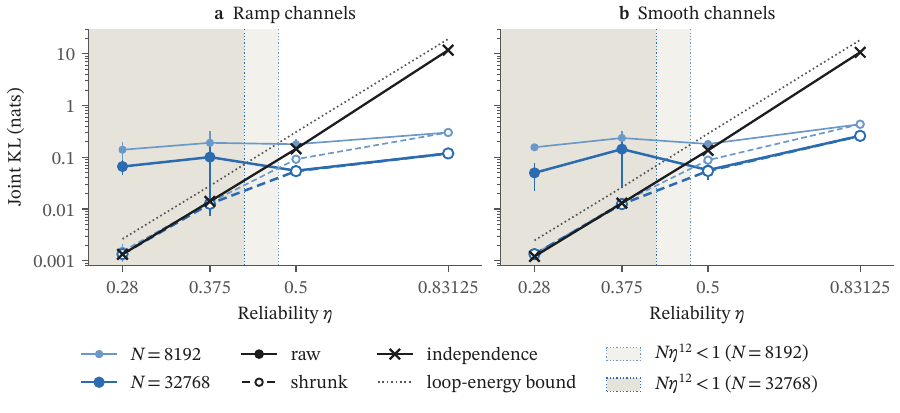}
\caption{Moments and validation-selected shrinkage for ramp (a) and
smooth (b) channels at $d=112$, $K=8$, with nested
training sets. Crosses give
independence KL from separate million-vector streams ($\pm1.96$ Monte
Carlo standard errors); other bars
are 95\% Student intervals over \SubmissionValue{weak-seeds} seeds.
Nested shaded regions with dotted edges mark $N\eta^{12}<1$:
$\eta<\SubmissionValue{weak-snr-threshold-n32768}$ at $N=32768$ and
$\eta<\SubmissionValue{weak-snr-threshold-n8192}$ at $N=8192$.}
\label{fig:calibration}
\end{figure}

\phantomsection\noindent\emph{Heterogeneous channels with weak anchors.}\label{app:nonramp}
The linear--sine channel is
$h_e(u)=b_e(2u-1)+0.15s_e\sin(4\pi u)$, with independent
$b_e\sim\mathrm{Unif}[0.7,0.8]$ and uniform signs $s_e$.
Its reliability is $b_e/2\in[0.35,0.4]$. Ten seeds at each $L=4,8,16$
redraw channels and observations, with nested training sizes
$N=512,2048,8192,32768$ and 2,048 independent test vectors per seed.
The longer-cycle fit splits data equally between calibration and regression, with
input anchor $\kappa_{\mathrm{in}}=0.3$, input cap $a_{\mathrm{in}}=0.98$, and
$K=\max\{4,\lceil N^{1/3}/2\rceil\}$.
The witness lengths $(|C|,|C'|,|C\triangle C'|)$ are $(4,4,6)$ for
interior edges, $(4,10,12)$ for boundary edges, and $(4,8,8)$ for corners.
Boundary edges lie on the outer cycle, excluding corner pairs;
other scalar edges are interior.
We estimate $\gamma_A^2=\theta_C\theta_{C\triangle C'}/\theta_{C'}$
from $C\cap C'=A$, clipping moments to $[\kappa_{\mathrm{in}}^{|F|},1]$
and the scale to $[\kappa_{\mathrm{in}}^{4-|A|},1]$.

\begin{table}[htbp]\centering\footnotesize
\begin{tabular}{@{}rrc@{\hspace{16pt}}rrr@{}}
\toprule $d$ & $N$ & Seeds & Fitted KL & Independence KL & Excess\\\midrule
24 & 512 & 10/10 & $0.044$\,{\footnotesize$[0.015, 0.078]$} & $0.006$\,{\footnotesize$[0.003, 0.008]$} & $0.038$\,{\footnotesize$[0.010, 0.073]$}\\*[2pt]
 & 2,048 & 10/10 & $0.061$\,{\footnotesize$[0.016, 0.144]$} &  & $0.055$\,{\footnotesize$[0.011, 0.139]$}\\*[2pt]
 & 8,192 & 10/10 & $0.056$\,{\footnotesize$[0.018, 0.098]$} &  & $0.051$\,{\footnotesize$[0.012, 0.095]$}\\*[2pt]
 & 32,768 & 10/10 & $0.042$\,{\footnotesize$[0.014, 0.071]$} &  & $0.037$\,{\footnotesize$[0.008, 0.064]$}\\*[2pt]
\midrule
112 & 512 & 10/10 & $0.236$\,{\footnotesize$[0.166, 0.306]$} & $0.044$\,{\footnotesize$[0.027, 0.059]$} & $0.191$\,{\footnotesize$[0.132, 0.246]$}\\*[2pt]
 & 2,048 & 10/10 & $0.312$\,{\footnotesize$[0.215, 0.447]$} &  & $0.268$\,{\footnotesize$[0.164, 0.401]$}\\*[2pt]
 & 8,192 & 10/10 & $0.281$\,{\footnotesize$[0.168, 0.421]$} &  & $0.236$\,{\footnotesize$[0.129, 0.384]$}\\*[2pt]
 & 32,768 & 10/10 & $0.224$\,{\footnotesize$[0.168, 0.284]$} &  & $0.179$\,{\footnotesize$[0.132, 0.247]$}\\*[2pt]
\midrule
480 & 512 & 10/10 & $1.165$\,{\footnotesize$[0.940, 1.490]$} & $0.186$\,{\footnotesize$[0.166, 0.210]$} & $0.980$\,{\footnotesize$[0.774, 1.285]$}\\*[2pt]
 & 2,048 & 10/10 & $1.517$\,{\footnotesize$[1.194, 1.913]$} &  & $1.331$\,{\footnotesize$[1.021, 1.737]$}\\*[2pt]
 & 8,192 & 10/10 & $1.382$\,{\footnotesize$[1.220, 1.616]$} &  & $1.196$\,{\footnotesize$[1.044, 1.424]$}\\*[2pt]
 & 32,768 & 10/10 & $1.056$\,{\footnotesize$[0.882, 1.352]$} &  & $0.871$\,{\footnotesize$[0.709, 1.165]$}\\*[2pt]
\bottomrule\end{tabular}
\caption{Joint KL and paired fitted-minus-independence loss in nats per vector for longer-cycle calibration with sample splitting on heterogeneous linear--sine channels. Entries are mean [minimum, maximum] across seeds; negative Monte Carlo estimates are retained. Positive excess favors independence. Each seed shares 2,048 test vectors across training sizes. The independence reference appears once per dimension.}\label{tab:nonramp-loss}
\end{table}
Longer-cycle fits lose to independence in
\SubmissionValue{nonramp-fits-above-independence} of
\SubmissionValue{nonramp-fits} fits (Table~\ref{tab:nonramp-loss}).
Full-sample short cycles lose in
\DiagnosticValue{nonramp-current-short-above-independence} of
\DiagnosticValue{nonramp-current-fits} fits with the same draws, $K$,
anchor and cap. At $d=24$, $N=32768$, mean KL is
$\DiagnosticValue{nonramp-current-d24-n32768-short-kl}$ (short cycles),
$\DiagnosticValue{nonramp-current-d24-n32768-legacy-kl}$ (longer cycles) and
$\DiagnosticValue{nonramp-current-d24-independence-kl}$ (independence).
True calibration scales and face moments, from reliabilities $b_e/2$,
still lose in
\DiagnosticValue{nonramp-current-oracle-above-independence} of
\DiagnosticValue{nonramp-current-fits} fits with estimated regressions.
Calibration error alone does not explain the excess KL; these refits do
not estimate an asymptotic rate.

\FloatBarrier
\FloatBarrier
\bibliography{references}
\end{document}